%% file: paper.tex
\documentclass[11pt]{article} 
\usepackage{etoolbox}

\input{arxiv_style}

\input{ayush}

\input{project_macros}

\usepackage{amsfonts}

\usepackage{xcolor} 

\usepackage[suppress]{color-edits} 
\addauthor{as}{red} 
\addauthor{sk}{purple}  
\addauthor{ks}{red}

\usepackage{mathrsfs} 

\usepackage{longtable}
\usepackage{import} 
\usepackage{enumitem} 
\usepackage{algorithm}
\usepackage{graphicx}

\usepackage[noend]{algpseudocode} 
 
\algblockdefx[class]{Class}{EndClass}[1]{\textbf{object} \textsc{#1}:}{\textbf{end object}}
\makeatletter 
\ifthenelse{\equal{\ALG@noend}{t}}
  {\algtext*{EndClass}}
  {}
\makeatother
\usepackage{amsmath}
\usepackage{makecell}
\usepackage{enumitem}  
\usepackage{bbm} 
\usepackage{bbold} 
\usepackage{xspace} 
\usepackage[scaled=.9]{helvet}
\usepackage{booktabs} 
\usepackage[export]{adjustbox}
\title{From Gradient Flow on Population Loss to \\ Learning with Stochastic Gradient Descent} 

  \author{
  Satyen Kale\footnote{Authors sorted alphabetically by last name. Correspondence to: Ayush Sekhari $\tri{\text{as3663@cornell.edu}}$}~\thanks{Google Research, NY} \\
{\small\texttt{satyenkale@google.com}}
\and 
Jason D. Lee\thanks{Princeton University and Google Research, Princeton}\\
{\small\texttt{jasonlee@princeton.edu}} \\
\and
Chris De Sa\thanks{Cornell University}\\
{\small\texttt{cdesa@cs.cornell.edu}}
\and
Ayush Sekhari$^\ddag$\\
{\small\texttt{as3663@cornell.edu}}
\and
Karthik Sridharan$^\ddag$\\
{\small\texttt{ks999@cornell.edu}}
}

\date{}

\begin{document} 
\maketitle 

\begin{abstract}
\input{files/abstract.tex}
\end{abstract} 

\section{Introduction} 
\input{files/main_introduction.tex}

\section{Setup} 
\input{files/main_setup.tex}

\section{Gradient Flow, Potentials and Geometry} \label{sec:gf} 
\input{files/main_gf.tex}

\subsection{Geometric Interpretation} \label{sec:complete_proofs} 
\input{files/main_geometry.tex}

\section{Stochastic Gradient Descent and Gradient Descent} \label{sec:discrete_algs} 
\input{files/main_discrete.tex}

\section{Examples: From Gradient Flow to Gradient Descent}  \label{sec:applications} 

\input{files/main_examples.tex}

\section{Conclusion} 
\input{files/main_extensions.tex}

\subsubsection*{Acknowledgements}
AS thanks Robert D. Kleinberg for useful discussions. KS acknowledges support from NSF CAREER Award 1750575. JDL acknowledges support of the ARO under MURI Award W911NF-11-1-0304,  the Sloan Research Fellowship, NSF CCF 2002272, NSF IIS 2107304,  NSF CIF 2212262, ONR Young Investigator Award, and NSF CAREER Award 2144994. CD acknowledges support from NSF CAREER Award 2046760.

\setlength{\bibsep}{6pt} 
\bibliography{refs}

\clearpage 

\newpage
\appendix 
\renewcommand{\contentsname}{Contents of Appendix}
\tableofcontents 
\addtocontents{toc}{\protect\setcounter{tocdepth}{3}} 
\clearpage 

\setlength\parindent{0pt}
\setlength{\parskip}{0.25em} 

\section{Preliminaries}
\input{files/appx_basic.tex}

\section{Proofs from \pref{sec:gf}} 
\input{files/appx_gf.tex}

\section{Proofs from \pref{sec:discrete_algs}}  
\input{files/appx_ub1.tex} 
\input{files/appx_ub_gd.tex}

\input{files/appx_ub_sgd_new.tex}

\section{Proofs from \pref{sec:applications}} \label{app:applications} 
\input{files/KL_functions.tex}

\input{files/appx_pr.tex}

\input{files/appx_applications_pr_new.tex}
\input{files/appx_applications_utility.tex}

\input{files/appx_applications_matrix_completion.tex}
\input{files/appx_applications_chaterjee.tex}

\clearpage

\input{files/appx_todo.tex}
\end{document}

%% file: arxiv_style.tex
 \usepackage[letterpaper, left=1in, right=1in, top=1in, bottom=1in]{geometry}

\usepackage[colorlinks=true, linkcolor=blue!70!black, citecolor=blue!70!black]{hyperref}
\usepackage{microtype}

\usepackage{natbib}
\bibpunct{(}{)}{;}{a}{,}{,}

\usepackage{amsthm}
\usepackage{mathtools}
\usepackage{amsmath}
\usepackage{bbm}
\usepackage{amsfonts}
\usepackage{amssymb}

\usepackage{xpatch}
\usepackage{pifont}
\usepackage{array}
\usepackage{booktabs}
\usepackage{floatrow}
\newfloatcommand{capbtabbox}{table}[][\FBwidth]
\usepackage{blindtext}
\usepackage{caption}

%% file: ayush.tex
\usepackage{mathtools}
\usepackage{amsthm}
\usepackage{amsmath}
\usepackage{amsxtra}
\usepackage{enumitem}      
\usepackage{amsthm}
\usepackage{mathtools}
\usepackage{bbm}
\usepackage{amsfonts}
\usepackage{amssymb}

\usepackage{MnSymbol} 

\usepackage{xpatch}

\theoremstyle{definition}  
\theoremstyle{plain}
\newtheorem{assumption}{Assumption}

\newtheorem{proposition}{Proposition}

\newtheorem{corollary}{Corollary}
\newtheorem{lemma}{Lemma}
\newtheorem{remark}{Remark}

\newtheorem{theorem}{Theorem}
\newtheorem{definition}{Definition}

\newtheorem*{theorem*}{Theorem}

\xpatchcmd{\proof}{\itshape}{\normalfont\proofnameformat}{}{}
\newcommand{\proofnameformat}{\bfseries}

\usepackage{prettyref}
\newcommand{\pref}[1]{\prettyref{#1}}

\newcommand{\savehyperref}[2]{\texorpdfstring{\hyperref[#1]{#2}}{#2}}
\newrefformat{eq}{\savehyperref{#1}{\textup{(\ref*{#1})}}}
\newrefformat{eqn}{\savehyperref{#1}{Equation~\ref*{#1}}}
\newrefformat{con}{\savehyperref{#1}{Conjecture~\ref*{#1}}}
\newrefformat{lem}{\savehyperref{#1}{Lemma~\ref*{#1}}}
\newrefformat{def}{\savehyperref{#1}{Definition~\ref*{#1}}}
\newrefformat{line}{\savehyperref{#1}{line~\ref*{#1}}}
\newrefformat{thm}{\savehyperref{#1}{Theorem~\ref*{#1}}}
\newrefformat{corr}{\savehyperref{#1}{Corollary~\ref*{#1}}}
\newrefformat{sec}{\savehyperref{#1}{Section~\ref*{#1}}}
\newrefformat{app}{\savehyperref{#1}{Appendix~\ref*{#1}}}
\newrefformat{ass}{\savehyperref{#1}{Assumption~\ref*{#1}}}
\newrefformat{ex}{\savehyperref{#1}{Example~\ref*{#1}}}
\newrefformat{fig}{\savehyperref{#1}{Figure~\ref*{#1}}}
\newrefformat{alg}{\savehyperref{#1}{Algorithm~\ref*{#1}}}
\newrefformat{rem}{\savehyperref{#1}{Remark~\ref*{#1}}}
\newrefformat{conj}{\savehyperref{#1}{Conjecture~\ref*{#1}}}
\newrefformat{prop}{\savehyperref{#1}{Proposition~\ref*{#1}}}
\newrefformat{proto}{\savehyperref{#1}{Protocol~\ref*{#1}}}
\newrefformat{prob}{\savehyperref{#1}{Problem~\ref*{#1}}}
\newrefformat{claim}{\savehyperref{#1}{Claim~\ref*{#1}}}

\newcommand\numberthis{\addtocounter{equation}{1}\tag{\theequation}}
\allowdisplaybreaks

\DeclarePairedDelimiter{\abs}{\lvert}{\rvert} 
\DeclarePairedDelimiter{\brk}{[}{]}
\DeclarePairedDelimiter{\crl}{\{}{\}}
\DeclarePairedDelimiter{\prn}{(}{)}
\DeclarePairedDelimiter{\nrm}{\|}{\|}
\DeclarePairedDelimiter{\tri}{\langle}{\rangle}

\DeclarePairedDelimiter{\floor}{\lfloor}{\rfloor}

\let\Pr\undefined

\DeclareMathOperator{\En}{\mathbb{E}}

\DeclareMathOperator{\Pr}{Pr}

\DeclareMathOperator*{\argmin}{argmin} 

\newcommand{\ls}{\ell}

\newcommand{\ldef}{\vcentcolon=}
\newcommand{\rdef}{=\vcentcolon}

\newcommand{\wt}[1]{\widetilde{#1}}
\newcommand{\wh}[1]{\widehat{#1}}
\newcommand{\mb}[1]{\boldsymbol{#1}}

\def\ddefloop#1{\ifx\ddefloop#1\else\ddef{#1}\expandafter\ddefloop\fi}
\def\ddef#1{\expandafter\def\csname bb#1\endcsname{\ensuremath{\mathbb{#1}}}}
\ddefloop ABCDEFGHIJKLMNOPQRSTUVWXYZ\ddefloop
\def\ddefloop#1{\ifx\ddefloop#1\else\ddef{#1}\expandafter\ddefloop\fi}
\def\ddef#1{\expandafter\def\csname b#1\endcsname{\ensuremath{\mathbf{#1}}}}
\ddefloop ABCDEFGHIJKLMNOPQRSTUVWXYZ\ddefloop
\def\ddef#1{\expandafter\def\csname c#1\endcsname{\ensuremath{\mathcal{#1}}}}
\ddefloop ABCDEFGHIJKLMNOPQRSTUVWXYZ\ddefloop
\def\ddef#1{\expandafter\def\csname h#1\endcsname{\ensuremath{\widehat{#1}}}}
\ddefloop ABCDEFGHIJKLMNOPQRSTUVWXYZabcdefghijklmnopqrsuvwxyz\ddefloop    
\def\ddef#1{\expandafter\def\csname hc#1\endcsname{\ensuremath{\widehat{\mathcal{#1}}}}}
\ddefloop ABCDEFGHIJKLMNOPQRSTUVWXYZ\ddefloop
\def\ddef#1{\expandafter\def\csname t#1\endcsname{\ensuremath{\widetilde{#1}}}}
\ddefloop ABCDEFGHIJKLMNOPQRSTUVWXYZ\ddefloop
\def\ddef#1{\expandafter\def\csname tc#1\endcsname{\ensuremath{\widetilde{\mathcal{#1}}}}}
\ddefloop ABCDEFGHIJKLMNOPQRSTUVWXYZ\ddefloop

\usepackage{tikz}

\newcommand{\grad}{\nabla}

\newcommand{\proman}[1]{\prn*{\romannumeral #1}}

\newcommand{\overleq}[1]{\overset{ #1}{\leq{}}}
\newcommand{\overgeq}[1]{\overset{#1}{\geq{}}}
\newcommand{\overeq}[1]{\overset{#1}{=}}

%% file: project_macros.tex
\newcommand{\sign}[1]{\text{sign}\crl*{#1}}  

\newcommand{\indicator}[1]{\mb{1}\crl*{#1}}

\newcommand{\reals}{\bbR}

\newcommand{\trace}[1]{\text{tr}\prn{#1}}
\newcommand{\bw}{\bar{w}}

\newcommand*\dif{\mathop{}\!\mathrm{d}}

\newcommand{\smin}{\sigma_{d}}
\newcommand{\smax}{\sigma_{\max}}

\newcommand{\bridge}[1]{g\prn{#1}} 
\newcommand{\taylor}{\theta} 

\newcommand{\clo}{\mathrm{clo}}

\newcommand{\pr}{\mathrm{pr}} 
\newcommand{\ms}{\mathrm{ms}}

\renewcommand{\bR}{\bar{R}}
 
\newcommand{\w}{w}
\newcommand{\Ber}{\text{Ber}}
\newcommand{\Phig}{\Phi_g}

\newcommand{\subs}[1]{_{\text{\tiny $#1$}}}

\newcommand{\Fkl}{F_{\mathrm{kl}}}
\newcommand{\Rkl}{R_{\mathrm{kl}}}

%% file: files/abstract.tex
Stochastic Gradient Descent (SGD) has been the method of choice for learning large-scale non-convex models. While a general analysis of when SGD works has been elusive,  there has been a lot of recent progress in understanding the convergence of Gradient Flow (GF) on the population loss, partly due to the simplicity that a continuous-time analysis buys us.  An overarching theme of our paper is providing general conditions under which SGD converges, assuming that GF on the population loss converges. Our main tool to establish this connection is a general \textit{converse Lyapunov} like theorem, which implies the existence of a Lyapunov potential under mild assumptions on the rates of convergence of GF. In fact, using these potentials, we show a one-to-one correspondence between rates of convergence of GF and geometrical properties of the underlying objective. When these potentials further satisfy certain self-bounding properties, we show that they can be used to provide a convergence guarantee for Gradient Descent (GD) and SGD (even when the paths of GF and GD/SGD are quite far apart). It turns out that these self-bounding assumptions are in a sense also necessary for GD/SGD to work. Using our framework, we provide a unified analysis for GD/SGD not only for classical settings like convex losses, or objectives that satisfy P\L~/ K\L~ properties, but also for more complex problems including Phase Retrieval and Matrix sq-root, and extending the results in the recent work of  \citet{chatterjee2022convergence}. 

%% file: files/main_introduction.tex
Stochastic Gradient Descent (SGD) has been a method of choice to train complex,  large scale machine learning models. While understanding of SGD for convex objectives is comprehensive, a general understanding of when SGD works for non-convex models has been somewhat elusive.  A large slew of properties like, convexity \citep{nemirovskij1983problem}, one-point-convexity \citep{onepoint},  linearizability \citep{sgdnotimp}, K\L~\citep{kl1,kl2} and P\L~ \citep{pl1,pl2,pl3} properties, and more problem specific, tailored analysis of SGD and Gradient Descent (GD) for specific problem instances like matrix square-root problem, matrix completion \citep{mcomplete}, phase retrieval \citep{candes2015phase,chen2019gradient, tan2019online} and Dictionary learning \citep{AroraGMM15} have been proposed. Recent success of SGD in over-parameterized deep learning models have lead to the idea that SGD perhaps optimizes training objective with an implicit bias given by some implicit regularizer \citep{gunasekar2018characterizing,soudry2018implicit, ji2018risk,gunasekar2018implicitb,gunasekar2018implicit}. However, in \cite{sgdnotimp} it is argued that there are over-parameterized models for which SGD works but no method that minimizes an implicit regularized training objective can learn successfully, thus showing that in general, the success of SGD cannot be explained by implicit regularization.

The goal of our paper is to provide a unifying analysis for when SGD/GD works. More specifically, we do this via first showing that Gradient Flow (GF) works and then extending this analysis to SGD and GD. Gradient Flow (GF) can be seen as a continuous time analogue of GD. In an idealized world, if one had access to the population loss, it turns out that convergence analysis for running gradient flow on population loss is somewhat simpler due to tools from continuous time analysis and PDEs. There has been several recent works \citep{GF1,GF2, GF3} that have provided convergence analysis for GF even on non-convex objectives. The high level theme of this paper is to show that, under some mild/appropriate assumptions of population loss/objective and on the noise of gradient estimates, ``if, GF converges on population loss, then SGD that uses one fresh example per iteration is successful at learning’’.  Notice, that GF converging on population loss is a purely deterministic optimization problem. However, the fact that SGD works is a learning result that implies a sample complexity bound. 

There have been past works that have aimed at providing convergence analysis for Gradient Descent (GD) starting from Gradient Flow (GF). Typical route to obtain a convergence analysis of GD starting from GF tries to think of GD updates as approximating GF path. Even with more sophisticated discretization schemes like Euler discretization, obtaining convergence for GD, starting from GF can be quite complex. In this paper, to show that when GF converges, SGD/GD also converges, we take a different approach.  A key tool for proving convergence results for GF is by constructing so called Lyapunov potentials. In the literature of Ordinary Differential Equations (ODEs), when ODEs have regular enough convergence rates, one can show, so called converse Lyapunov theorems (see \cite{converse} for a nice survey of classic results) that state that when an ODE converges to stable solutions, there has to exist a corresponding Lyapunov potential. While convergence of GF in terms of sub-optimality is quite different from convergence in the ODE sense, in this paper, we first prove a converse Lyapunov style theorem for GF. Specifically, we show that when GF converges in terms of sub-optimality to a global minimum, then there has to exist a corresponding Lyapunov potential and using such potential, the rates can be recovered. This result becomes a starting point for our analysis. We show that if this Lyapunov potential (obtained from the converse Lyapunov style theorem) satisfies certain extra self-bounding regularity conditions, then one can show that GD and SGD algorithms converge in terms of sub-optimality when appropriate step sizes are used. Such convergence for SGD/GD happens even when the GF path and GD/SGD paths can be quite different. 
 
We summarize our main contributions below: 
\begin{itemize}
\item We prove a converse Lyapunov style theorem that shows that if gradient flow converges with rate specified by with an appropriate rate function, then there exists a corresponding Lyapunov potential that recovers this rate.
\item We provide a geometric characterization for a given rate of convergence of gradient flow (ie. GF converges at a particular rate if and only if a specific geometric condition on objective holds.)
\item There are problems for which GF converges at a specific rate but GD can be arbitrarily slow to converge.
\item This motivates the necessity of additional conditions to ensure GD/SGD converges even when GF converges. We provide certain self-bounding regularity conditions on the Lyapunov potential, under which we show that GD converges. We also provide conditions on gradient estimate noise under which we show that SGD using these gradient estimates also converges.
\item We instantiate our results for problems such K{\L} functions, matrix square-root and phase retrieval, amongst other applications.
\end{itemize} 

Informally speaking, our results suggest that the rate at which gradient flow convergences on the population loss can be used to get a learning guarantee for SGD (under mild additional regularity assumptions).

%% file: files/main_setup.tex
Given a continuously differentiable function and non-negative function  \(F: \bbR^d \mapsto \bbR\), our goal is to minimize \(F(w)\). Without any loss of generality, we assume that \(\min_{w} F(w) = 0\). First-order algorithms are popular for such optimization tasks. In the following,  we formally describe the Gradient Descent and Stochastic Gradient Descent algorithm, and their continuous time counterpart called gradient flow. 

\paragraph{Gradient Descent (GD).} Gradient descent is the most popular iterative algorithm to minimize differentiable functions. Starting from an initial point \(w_0\), GD on the function \(F(w)\) performs the following update on every iteration:
\begin{align}
w_{t+1} \leftarrow w_{t} - \eta \nabla F(w_t),  \label{eq:GD}
\end{align}
where $\eta$ denotes the step size. After $T$ rounds, GD algorithm returns the point $\widehat{\w}_T \ldef{} \argmin_{s \leq T} F(w_s)$. 

\paragraph{Stochastic Gradient Descent (SGD).} Stochastic gradient descent (SGD) has been the method of choice for optimizing complex convex and non-convex learning problems in practice. In the learning setting,  \(F(w)\) corresponds to the unknown population loss and can be written as \(F(w) = \En_{z \sim \cD}\brk*{f(w; z)}\) where the expectation is taken with respect to samples \(z\) drawn from an unknown distribution \(\cD\). SGD algorithm (mini-batch size 1) is an iterative algorithm that at every round \(t \geq 0\), draws a fresh sample \(z_t\) from \(\cD\) to compute a stochastic unbiased estimate \(\grad f(w_i; z_i)\) of the gradient \(\grad F(w_i)\), and performs the update \begin{align}\label{eq:SGD}
\w_{t+1} \leftarrow \w_{t} - \eta \nabla f(\w_t;z_t) 
\end{align} 
where \(\eta\) is the step size and \(w_0\) denotes the initial point. After \(T\) rounds, SGD algorithm returns $\wh w_T$ by sampling a point uniformly at random from the set \(\crl{w_1, \dots, w_T}\). 

\paragraph{Gradient Flow (GF).}  Gradient flow from a point \(w_0\) is continuous time process $(w(t))_{t \geq 0}$ that starts at \(w(0) = w_0\) and evolves as 
\begin{align}
\frac{\dif w(t)}{\dif t} = - \grad F(w(t)). \label{eq:GF}
\end{align} 
GF has been thought of as a continuous time analogue of GD and is popularly used to understand behavior of gradient based optimization algorithms in the limit, primarily due to its simplicity and lack of step size. 

\textbf{Additional notation.} For a vector \(w \in \bbR^d\), $\w[j]$ denotes its $j$-th coordinate and $\nrm{w}$ denotes its Euclidean norm. For any \(w_1, w_2 \in \bbR^d\),  $\tri{\w_1, \w_2}$ denotes their inner product. For a matrix \(W\), \(\smin(W)\) and \(\nrm{M}\) denotes its minimum singular value and spectral norm respectively. We define the set \(\bbR^+\) to contain all non-negative real numbers. We use $\mb{1}_{d}$ to denote a $d$-dimensional vector of all $1$s, and $\mathrm{I}_d$ to denote the identity matrix in $d$-dimensions. $\cN(0, \sigma^2 \mathrm{I}_d)$ denotes \(d\)-dimensional Gaussian distribution with variance $\sigma^2 \mathrm{I}_d$. $\Ber(p)$ denotes the Bernoulli distribution with mean $p$. 

For a function $f: \bbR^d \mapsto \bbR$, we denote the \(p\)-th derivative at the point $\w$ by $\grad^p f(\w) \in \bbR^d$. We say that a real valued \(f\) is monotonically increasing if \(f' \geq 0\), and monotonically decreasing if \(f' \leq 0\). The function $f$ is said to be $L$-Lipschitz if $f(\w_1) - f(\w_2) \leq L \nrm*{\w_1 - \w_2}$ for all $\w_1, \w_2$.  For a set of initial points \(\cW\), we denote \(\clo(\cW)\) as its \textit{closure} under GF, i.e.
 \(\clo(\cW)  = \crl{w' ~:~ w' \text{~is in GF path of some~} w_0 \in \cW}\). 

%% file: files/main_gf.tex
Lyapunov potentials are a popular tool for understanding convergence of GF  \citep{krichene2016lyapunov, wilson2018lyapunov, wilson2021lyapunov}. At an intuitive level, a Lyapunov potential is any non-negative function \(\Phi\) that satisfies  \(\tri{\grad {\Phi}(w), - \grad F(w)} \leq 0\), i.e. \(\Phi\) decreases along the GF paths of \(F\). This monotonicity property helps to show asymptotic convergence of GF to stable points of the underlying objective. In our work, we consider potential functions for which the rate of change (decrease) of the potential along the GF path is related to the suboptimality of the objective at that point.  

\begin{definition}[Admissible potentials]
\label{def:Phi} A differentiable potential function \(\Phi_g: \bbR^d \mapsto \bbR^+\) is admissible w.r.t.~\(F\) on a set \(\cW\) if there exists a monotonically increasing function \(g: \bbR^+ \mapsto \bbR^+\) with \(g(0) = 0\) such that for any \(w \in \cW\),\footnote{Whenever not specified, we assume that \(g(z) = z\). The function \(\Phi\) denote the potential function \(\Phi_g\) with \(g(z) = z\).} 
\begin{align*} 
\tri*{\grad \Phi_g(w), \grad F(w)} \geq \bridge{F(w)}. \numberthis \label{eq:linearity_property}
\end{align*} 
\end{definition} 
Existence of potential functions of the above form can be used to provide rates of convergence for GF as show in the following theorem. 
\begin{theorem}[From potentials to gradient flow] 
\label{thm:potential_to_gf}
Let \(\cW\) be a set of initial points that we want to consider, and let \(\Phi_g\) be an admissible potential w.r.t. \(F\) on the set \(\clo(\cW)\). Then, for any initialization \(w_0 \in \cW\), the point \(w(t)\) on the GF path with \(w(0) = w_0\) satisfies for any \(t \geq 0\), 
\begin{align*}
g(F(w(t))) \leq \frac{\Phi_g(w_0)}{t}. 
\end{align*} 
\end{theorem}
The idea that admissible potential functions imply convergence rates for GF has appeared in various forms in the prior literature \citep{bansal2017potential, krichene2016continuous, wilson2018lyapunov, wilson2021lyapunov}. As an example, consider the potential function \(\Phi(w) = \nrm{w - w^*}/2\). Notice that \(\Phi\) is an admissible potential for any \(F\) that is convex with \(g(z) = z\). This is because convexity implies that \pref{eq:linearity_property} is true for any \(w\). Hence, using \pref{thm:potential_to_gf} we get a \(\nrm{w - w^*}^2 / 2t\) rate of convergence for GF on any convex objective. 

Our main result in this section is to establish a converse Lyapunov style theorem\(-\)that given a rate, finds a potential function corresponding to that rate. We start by defining admissible rate functions.
\begin{definition}[Admissible rate functions] 
\label{def:rate} 
 A function \(R: \bbR^d \times \bbR \mapsto \bbR^+\) is an admissible rate function w.r.t.~\(F\) if for any \(w \in \bbR^d\), 
\begin{enumerate}[label=\((\alph*)\), leftmargin=8mm]  
\item \(R(w, t)\) is a non-increasing function of \(t\) such that \(\lim_{t \rightarrow \infty} R(w, t) = 0\). 
\item \(R\) satisfies the relation: \(\int_{t=0}^\infty \prn*{\tfrac{\partial R(w, t)}{\partial t} + \left\langle \grad  R(w, t), \nabla F(w) \right\rangle } \dif t \geq 0.\)
\end{enumerate} 
\end{definition}

\begin{remark}
\label{rem:rate_fn_relaxation}
 In order to simplify the task of checking whether a given rate is admissible, note that \pref{def:rate}-(b) is satisfied whenever the condition $$\frac{\partial R(w, t)}{\partial t} + \left\langle \grad  R(w, t), \nabla F(w) \right\rangle  \geq 0$$ holds for every \(w\) as \(t \rightarrow 0\). Many rate functions, e.g. \(R(w, t) = F(w) e^{-t}\) and \(F\) being K\L, in fact satisfy this condition for every \(w, t \geq 0\). 

Furthermore, also note that \pref{def:rate}-(b) is satisfied whenever the rate function is such that \(R(w(s), t) \leq R(w, s + t)\) for all \(s, t \geq 0\) and \(w \in \bbR^d\), which may be an easier to check condition, e.g. when \(R(w, t) = F(w) e^{-t}\). 
\end{remark}

We utilize admissible rate functions to characterize behavior of GF on \(F\). Before we proceed, let us motivate the two properties above. Property $(a)$ is natural for any rate function and captures the fact that running GF for more time leads to better guarantees. Property \((b)\), while seeming a bit mysterious, characterizes the compatibility of the rate function w.r.t. ~gradient flow dynamics. For interpretation consider the relaxed version given in \pref{rem:rate_fn_relaxation} which implies property-(b). Here, the condition that \(R(w(s), t) \leq R(w, s + t)\) for all \(s, t \geq 0\) and \(w \in \bbR^d\) simply captures the fact that having additional information about the GF path should only improve the rate. Note that \(R(w(0), s + t)\) corresponds to an upper bound on the sub-optimality at \(w(s + t)\) and \(R(w(s), t)\) corresponds to an upper bound on the same quantity but with the additional information that \(w(s)\) is a point on the GF path. We remark that for any rate function \(R\), it is easy to construct a new rate function \(\bar{R}\) that always satisfies this condition (hence, property \((b)\)) by defining \(\bar{R}(w, t) = \min_{s \geq 0} R(w', t + s)\) where \(w'\) is any point such that the point \(w\) lies on the GF path from \(w'\) at time \(s\). Furthermore, the function \(R(w, t) = F(w(t))\) is always an admissible rate function. All the  rate functions appearing in this paper satisfy both properties \((a)\) and \((b)\).  

Our next result shows that admissible rate functions for GF can be used to construction admissible potentials w.r.t.~\(F\). 

\begin{theorem}[From gradient flow to potentials]
\label{thm:gf_to_potential}   
Let \(\cW \subseteq \bbR^d\) be any set of initial points that we want to consider, and \(R\) be an admissible rate function w.r.t. all GF paths originating from any point in \(\cW\). Further, suppose that for any \(w_0 \in \cW\), the point \(w(t)\) on the GF path satisfies \(F(w(t)) \leq R(w_0, t)\), then the function  \(\Phi_g\) defined as  
\begin{align*}
\Phi_g(w) = \int_{t=0}^\infty g(R(w, t)) \dif t  \numberthis \label{eq:Phi_g_defn} 
\end{align*} 
is an admissible potential w.r.t. \(F\) on the set  \(\mathrm{clo}(\cW)\), for any differentiable and monotonically increasing function \(g: \bbR^+ \mapsto \bbR^+\) that satisfies \(\int_{t=0}^\infty g(R(w, t)) \dif t < \infty\) and \(\int_{t=0}^\infty g'(R(w, t)) \nrm{\grad R(w, t)} \dif t  < \infty\) for every \(w \in \mathrm{clo}(\cW)\). 
\end{theorem} 

As an illustration on how to apply \pref{thm:gf_to_potential}, assume that for \(F\) the rate for GF is  \(R(w, t) = F(w) e^{-t}\) For instance, we already know that such a rate holds when \(F\) is P{\L}. For this rate, by choosing \(g(z) = z\), we get that the function \(\Phi_g(w) = \int_{t=0}^\infty {F(w) e^{-t}} \dif t = F(w)\) is an admissible potential w.r.t.~\(F\). We provide more examples in \pref{sec:applications}. 

\pref{thm:potential_to_gf} and \pref{thm:gf_to_potential} are, in a sense, converse of each other. \pref{thm:potential_to_gf} shows that the existence of an admissible potential function implies a rate of convergence for GF. On the other hand, \pref{thm:gf_to_potential} shows how to construct admissible potentials starting from the fact that GF has a rate. One might wonder whether there always exist a Lyapunov function, more specifically a \(g\) function above, such that the rate implied by the constructed potential in \pref{thm:potential_to_gf} matches the rate that we started with for \pref{thm:gf_to_potential}, i.e. \(R(w, t) \approx g^{-1}\prn*{\Phi_g(w)/t}\). We answer this in the positive for rate functions that are of the product form. 

\begin{corollary} 
\label{corr:converse_gf_tight} 
Let \(\cW \subseteq \bbR^d\) be any set of initial points that we want to consider, and \(R\) be an admissible rate function w.r.t. all GF paths originating from points in \(\cW\). Additionally, suppose \(R\) has the product form \(R(w, t) = h(w) r(t)\) where \(h\) is differentiable and \(r\) is a non-increasing function that satisfies  \(r(t) \leq \lambda \abs{r'(t)} \max\crl{1, t}\) for any \(t \in \bbR\) (where \(\lambda\) is a universal constant). Furthermore, suppose that for any \(w_0 \in \cW\), the point \(w(t)\) on the GF path satisfies \(F(w(t)) \leq R(w_0, t)\). Then, there exists a monotonically increasing function \(g: \bbR^+ \mapsto \bbR^+\) such that the potential \(\Phi_g(w)\) constructed in \pref{thm:gf_to_potential} using $g$, when plugged in \pref{thm:potential_to_gf}, implies that GF has the rate 
\begin{align*} 
F(w(t)) \leq \max_{w \in \cW} R(w, t/\log^2(t))  
\end{align*} 
for any initialization \(w(0) \in \cW\). 
\end{corollary} 

%% file: files/main_geometry.tex
The definition of an admissible potential comes with a geometric condition on the function \(F\) given in \pref{eq:linearity_property}. Since \pref{thm:gf_to_potential} constructs admissible potentials,  when a rate \(R(w, t)\) holds for GF it suggests that the geometric property in \pref{eq:linearity_property} holds for the objective function \(F\). As an example, say GF on \(F\) satisfies the rate \(R(w, t) = F(w)e^{-t}\). From \pref{thm:gf_to_potential}, we note that \(\Phi_g(w) = F(w)\) is an admissible potential w.r.t~\(F\) with \(g(z)= z\). This implies the geometric property
\begin{align*}
\tri*{\grad F(w), \grad F(w)} \geq F(w) \numberthis \label{eq:linearizability}
\end{align*}
holds for \(F\) whenever GF has rate  \(R(w, t) = F(w) e^{-t}\). On the other, we know that whenever \pref{eq:linearizability} holds the function \(\Phi(w)=F(w)\) satisfies \pref{eq:linearity_property} and is thus an admissible potential for \(F\) (with \(g(z) = z\)), and hence \pref{thm:potential_to_gf} implies the rate of \(F(w)/t\), which is equivalent to the rate \(F(w) e^{-t}\) (c.f. \pref{lem:linear_to_exp}). This implies an equivalence between the rates \(R(w, t) = F(w) e^{-t}\) and the geometric property \pref{eq:linearizability}. We formalize this in the following. 
\begin{proposition} 
\label{prop:PL_complete}  
The following two properties are equivalent: 
\begin{enumerate}[label=(\alph*), leftmargin=8mm] 
\item For any \(w(0) \in \bbR^d\) and \(t \geq 0\), GF has the rate \(F(w(t)) \leq F(w(0)) e^{- \lambda t}\), 
\item  \(F(w)\) satisfies the Polyak-\L ojasiewicz (PL) property i.e. \(\lambda F(w) \leq \nrm*{\grad F(w)}^2\), 
\end{enumerate} 
for any \(\lambda \geq 0\). \pref{thm:gf_to_potential} implies $(b)$ and yields the potential function \(\Phi(w) = F(w)\).
\end{proposition} 

A similar equivalence also holds for the more general class of K\L~functions. We defer this result to \pref{prop:KL_complete} in \pref{sec:KL_functions}. In the following, we show a correspondence between the rate \(R(w, t) = \frac{\nrm{w(0) - w^*}^2 - \nrm{w(t) - w^*}^2}{2t}\), and linearizability---a condition that is weaker than convexity but is sufficient for the corresponding rate of convergence for GF. \begin{proposition}
\label{prop:convexity_complete} 
The following two properties are equivalent: 
\begin{enumerate}[label=(\alph*), leftmargin=8mm] 
\item For any \(w(0) \in \bbR^d\) and \(t \geq 0\), GF has the admissible rate \(F(w(t)) \leq \lambda \frac{\nrm{w(0) - w^*}^2 - \nrm{w(t) - w^*}^2}{2t}\), 
\item  \(F(w)\) is linearizable w.r.t. \(w^*\) i.e. \(F(w) \leq \lambda \tri*{\grad F(w), w - w^*}\), 
\end{enumerate} 
for any \(\lambda \geq 0\). 
\end{proposition}

More generally, the equivalence between GF rates and the corresponding geometry on \(F\) can be characterized as follows. 
\begin{remark} 
GF on \(F\) enjoys the admissible rate $R(w, t)  = g^{-1} \prn*{{\Phi_g(w)}/{t}}$ if and only if \(F\) has the geometric property $\tri*{\grad \Phi_g(w), \grad F(w)} \geq \bridge{F(w)}.$
\end{remark}

%% file: files/main_discrete.tex
GD can be thought of as an approximate discretization of gradient flow. Thus, for problems where GF converges with a given rate \(R\), one may try to get convergence guarantees for GD from an initial point \(w_0\) by bounding the distance between the GD and GF trajectories starting from \(w_0\). This is exactly the approach taken in prior works \citep{gunasekar2021mirrorless, NIPS2015_f60bb6bb, JMLR:v22:20-195, su2014differential, zhang2021revisiting, elkabetz2021continuous}. However, coming up with non-vacuous bounds on the distance between corresponding GF and GD iterated is often quite challenging and requires much stronger assumptions on the underlying objective. In fact there are cases where both GF and GD converge to the same global minimum but their paths can be quite far away from each other. We take a different approach for proving convergence of GD/SGD which directly relies on the properties of corresponding potential for \(F\). In the following theorem, we  note that further assumption on top of the premise that GF has a rate are required, to even hope that GD succeeds. 

\begin{theorem} 
\label{thm:f_Phi_regularity} For any integer $T_0 > 0$, there exists a continuously differentiable convex function $F$ for which \(\min_{w} F(w) = 0\) and \(w^* = 0\) is the unique minimizer, such that: 
\begin{enumerate}[label=(\alph*)] 
\item \(\Phi(w) = \nrm{w}^2/2\) is an admissible potential for \(F\). Thus, \pref{thm:potential_to_gf} implies that for any initial point \(w_0\), the point \(w(t)\) on its GF path satisfies \(F(w(t)) \leq \frac{\nrm{w_0}^2}{2t}\). 
\item There exists an initial point \(w_0\) with \(\nrm{w_0} \leq 1\) and \(F(w_0) \leq 2\) such that GD fails to find an \(1/10\)-suboptimal solution for any step size \(\eta\) within $t \le T_0$ steps. 
\end{enumerate}
\end{theorem} 

Before giving our exact assumptions and the convergence bounds, we provide the intuition behind how admissible potentials can be used for analyzing GD (or SGD). Let the sequence of iterates generated by GD algorithm be given by \(\crl*{w_t}_{t \geq 0}\), \(g(z) = z\) and \(\Phi\) be an admissible potential w.r.t.~$F$. For any time \(t\), the second-order Taylor's expansion of the potential \(\Phi\) implies that 
\begin{align*}
\Phi(w_{t+1}) &\leq \Phi(w_{t})  + \tri*{\grad \Phi(w_t), w_{t+1} - w_t} + \prn*{w_{t+1} - w_t}^T  \grad^2 \Phi(\wt{w_t})\prn*{w_{t+1} - w_t} \\
&\leq \Phi(w_{t})  - \eta \tri*{\grad \Phi(w_t), \grad F(w_t)} + \eta^2 \prn*{\grad F(w_t)}^T  \grad^2 \Phi(\wt{w_t})\prn*{\grad F(w_t)}, \end{align*}
where \(\wt{w_t} = \beta w_t + (1 - \beta) w_{t+1}\) for some \(\beta \in [0, 1]\), and the second line follows by plugging the GD update \(w_{t+1} = w_{t} - \eta \grad F(w_t)\). Rearranging the terms, we get that 
\begin{align*}
\tri*{\grad \Phi(w_t), \grad F(w_t)} &\leq  \frac{\Phi(w_{t})  - \Phi(w_{t+1})}{\eta} + \eta \prn*{\grad F(w_t)}^T  \grad^2 \Phi(\wt{w_t})\prn*{\grad F(w_t)}, \numberthis \label{eq:intuition_proof0} 
\end{align*}
The key idea that enables us to get performance guarantees for GD is that the linear term in the left hand side above upper bounds the suboptimality of \(F\) at the point \(w_t\) since \(\Phi\) is an admissible potential w.r.t.~$F$. In particular, the condition  \pref{eq:linearity_property} implies that 
\begin{align*}
\tri*{\grad \Phi(w_t), \grad F(w_t)} \leq F(w_t). 
\end{align*}
Using the above relation in \pref{eq:intuition_proof0}, telescoping \(t\) from \(0\) to \(T-1\), and dividing by \(T\), we get that 
\begin{align*}
\frac{1}{T} \sum_{t=0}^{T-1} g(F(w_t)) \leq \frac{\Phi(w_0) - \Phi(w_T)}{\eta T} + \frac{\eta}{T} \cdot \sum_{t=0}^{T-1} \prn*{\grad F(w_t)}^T  \grad^2 \Phi(\wt{w_t})\prn*{\grad F(w_t)}. \numberthis \label{eq:intuition_proof3}
\end{align*}
Thus, we can bound the expected suboptimality of the point \(\wh w \sim \text{Uniform}\prn*{\crl*{w_0, \dots, w_{T-1}}}\) returned by the GD algorithm after \(T\) steps, whenever the second order term in the bound \pref{eq:intuition_proof3} is well behaved. For example, if $ \prn*{\grad F(w_t)}^T  \grad^2 \Phi(\wt{w_t})\prn*{\grad F(w_t)} \leq K$ for any \(w_t, w_{t+1}\) and \(\wt w_t\), we immediately get that 
\begin{align*}
\frac{1}{T} \sum_{t=0}^{T-1} g(F(w_t)) \leq \frac{\Phi(w_0) - \Phi(w_T)}{\eta T} + \eta K = O \prn*{\frac{1}{\sqrt{T}}}, 
\end{align*} for \(\eta = O(1/\sqrt{T})\). While the above holds for a very simplified setup, the intuition can be extended to more general cases as well. Below we present two regularity conditions that are sufficient to show convergence of GD.   

\begin{assumption} 
\label{ass:F_regular} There exists a monotonically increasing function \(\psi: \bbR^+ \mapsto \bbR^+\) such that  \(\nrm{\grad F(w)}^2 \leq \psi(F(w))\) for any point \(w \in \cW\). 
\end{assumption}
\begin{assumption}
\label{ass:Phi_regular} 
The potential function \(\Phi\) is second-order differentiable, and there exists a monotonically increasing function \(\rho: \bbR^+ \mapsto \bbR^+\) such that \(\nrm{\grad^2 \Phi(w)} \leq \rho(\Phi(w))\) at any point \(w \in \cW\). 
\end{assumption}

We will refer to the above conditions on \(F\) and \(\Phi\) as self-bounding regularity conditions. The following theorem provides convergence guarantees for GD when an admissible potential exists and the above assumptions are satisfies. 

\begin{theorem}[GD convergence guarantee] 
\label{thm:GD_guarantee}
Let \(\Phi_g\) be an admissible potential w.r.t.~$F$. Assume that \(F\) satisfies  \pref{ass:F_regular} and \(\Phi_g\) satisfies \pref{ass:Phi_regular}. Then, for any \(T \geq 0\) and setting \(\eta\) appropriately, the point \(\wh w_T\) returned by GD algorithm has the convergence guarantee\footnote{\label{note1}The \(O(\cdot)\) notation here hides initialization and problem dependent constants fully specified in the Appendix.}
 \begin{align}
\bridge{F(\wh w_T)} &= O \prn[\big]{\tfrac{1}{\sqrt{T}}}. \label{eq:GD_guarantee1}
\end{align} Furthermore, if the function \(\lambda(z) \ldef{} \tfrac{\psi(z)}{g(z)}\) is monotonically increasing in \(z\), then for a different appropriate choice of \(\eta\), 
\begin{align} 
\bridge{F(\wh w_T)} &= O\prn[\big]{\tfrac{1}{T}}. \label{eq:GD_guarantee2} 
\end{align} 
\end{theorem} 

Let us consider an example. Suppose that gradient flow on \(F\) achieves the admissible rate \(R(w, t) = {\prn*{\nrm{w - w^*}^2 - \nrm{w(t) - w^*}^2}}/{2t}\). This implies that \(F\) is linearizable (\pref{prop:convexity_complete}), and thus \(\Phi_g(w) = \nrm{w - w^*}^2/2\) is an admissible potential for \(F\) with \(g(z)= z\) as it clearly satisfies \pref{eq:linearity_property}.  However,  as we saw in \pref{thm:f_Phi_regularity} just existence of such a rate function does not imply the GD will succeed and we need to make further assumptions. Notice that in this case \(\Phig(w)\) satisfies \pref{ass:Phi_regular} with \(\rho(z)=1\). If we further assume that \(F\) is \(L\)-Lipschitz, then \pref{ass:F_regular} is satisfied with \(\psi(z) = L^2\). Hence, applying \pref{thm:GD_guarantee} for this setting, we get that GD has convergence rate \(F(\wh w_T) = O({\nrm*{w - w^*} L}/{\sqrt{T}})\). Instead if \(F\) was \(H\)-smooth, \pref{ass:F_regular} is satisfied with \(\psi(z) = 4H z\) and \(\psi(z)/g(z) = 4H\) is a monotonically increasing function and thus using \pref{eq:GD_guarantee2}, we get that GD has the convergence rate \(F(\wh w_T) = O({H\nrm*{w - w^*}^2}/{T})\). Notice that both of these rates are optimal for GD under the Lipschitz/Smoothness assumptions on \(F\), and the fact that \(F\) is linearizable \citep{nemirovskij1983problem}. On similar lines, using the rates for GF convergence on PL/K\L~functions, we can also recover optimal convergence rates for GD under appropriate smoothness assumptions on \(F\).

We next consider the convergence of SGD algorithm. Recall that at the iterate \(w_t\), SGD  performs the updated using \(\grad f(w_t, z_t)\), a stochastic and unbiased estimate of \(\grad F(w_t)\). Of course, unless one has some form of control over the distribution of \(\grad f(w_t, z_t)\), one cannot hope to prove any convergence guarantees of SGD. To this end, we make the following regularity assumption on the noise in  \(\grad f(w, z_t)\) while estimating \(\grad F(w)\). 

\begin{assumption}[Noise regularity]  
\label{ass:gradient_noise} There exists a monotonically increasing function \(\chi: \bbR^+ \mapsto \bbR^+\) such that for any point \(w\), the gradient estimate \(\grad f(w, z)\) satisfies 
\begin{align*} 
\Pr\prn*{\nrm{\grad f(w; z)  - \grad F(w)}^2 \geq t \cdot \chi(F(w))} \leq e^{- t}. 
\end{align*}
\end{assumption} 
\pref{ass:gradient_noise} is quite general, and can be specialized by appropriately setting the function \(\chi\) to model various stochastic optimization problem settings observed in practice. For example,  the classical stochastic optimization setting in which \(\grad f(w; z) = \grad F(w) + \varepsilon_t\) where \(\varepsilon_t\) is a sub-Gaussian random variable with mean \(0\) and variance \(\sigma^2\) is captured by the above assumption when \(\chi(z) = \sigma^2\) \citep{nemirovskistochastic}. However, it turns out that for many interesting ML problems, the noise typically scales with the function value \citep{wojtowytsch2021discrete, wojtowytsch2021stochastic}. 

\begin{theorem}[SGD convergence guarantee] 
\label{thm:SGD_guarantee} 
Let \(\Phi_g\) be an admissible potential w.r.t.~$F$. Assume that \(F\) satisfies  \pref{ass:F_regular}, \(\Phi_g\) satisfies \pref{ass:Phi_regular} and the stochastic gradient estimates \(\grad f(w; z)\) satisfy \pref{ass:gradient_noise}. Then, for any \(T \geq 0\) and setting \(\eta\) appropriately, the point \(\wh w_T\) returned by SGD algorithm has the convergence guarantee$^{\textcolor{blue}{1}}$ 
\begin{align*} 
\bridge{F(\wh w_T)} &= \wt{O}\prn[\big]{\tfrac{1}{\sqrt{T}}}. 
\end{align*} 
with probability at least \(0.7\) over the randomization of the algorithm and stochastic gradients. 
\end{theorem} 

\begin{remark} In most classic settings, one expects a \(1/\sqrt{T}\) rate for SGD \cite{bubeck2015convex}. However, in cases where \(\Phi_g\) is an admissible potential and \(g(z) = o(z)\), \pref{thm:SGD_guarantee} seems to suggest a \(g^{-1}({1}/{\sqrt{T}})\) rate of convergence which is faster than  \(1/\sqrt{T}\). This is where the self-bounding regularity conditions play an important role. As an example for P\L~ style rates, one can show that \(F(w)^p\) is an admissible potential with \(g(z) = z^p\) for any \(p\). However, the self-regularity conditions are not satisfied unless \(p \geq 1\). Setting \(p = 1\)  recovers the \(1/\sqrt{T}\) rate of SGD for P\L~ functions  which is optimal  \cite{agarwal2009information}. 
\end{remark}

%% file: files/main_examples.tex
So far, we discussed classical examples like P\L~ functions, convex functions, etc. At a high level, in order to  show convergence of SGD for these problems, we first establish an admissible rate of convergence for gradient flow, which implies an admissible potential that is used to show convergence of SGD. In this section, we extend this approach for other more complex non-convex stochastic optimization problems. 

\subsection{Kurdyka-\L ojasiewicz (K\L) functions} \label{sec:KL_functions}
Kurdyka-\L ojasiewicz (K\L)~ functions appear in various non-convex learning settings, for instance,  generalized linear models \citep{mei2021leveraging}, low-rank matrix recovery \citep{bi2022local}, over parameterized neural networks \citep{zeng2018global, allen2019convergence}, reinforcement learning  \citep{agarwal2021theory, mei2020global, yuan2022general} and optimal control  \citep{bu2019lqr, fatkhullin2021optimizing}. We recall the following definition of K\L~ functions, where we assumed that \(\Fkl\) is non-negative and  \(\min_{w} \Fkl(w) = 0\).\footnote{Various other definitions K\L~functions appear in the literature. However all of them are equivalent under the appropriate change of variables.} 
\begin{definition}[K\L~ functions]
\label{def:KL_def_appx} 
The objective \(\Fkl\) satisfies Kurdyka-\L ojasiewicz (K\L) property with exponent \(\theta \in (0, 1)\) and coefficient \(\alpha \in \bbR^+\), if for any point \(w\), 
\begin{align*} 
\nrm{\grad \Fkl(w)}^2 \geq \alpha \Fkl(w)^{1 + \theta}.  
\end{align*} 
\end{definition} 

Note that the above K\L~property generalizes the P\L~property we considered in earlier sections; setting \(\theta = 0\) results in P\L~property. We note the following rate of convergence for gradient flow for  K\L~functions. 
\begin{lemma}
\label{lem:KL_rate_main}
For any initial point point \(w(0) = w_0\), the point \(w(t)\) on its gradient flow path satisfies  
\begin{align*}
\Fkl(w(t)) \leq \Rkl(w_0, t) \ldef{} \frac{\Fkl(w_0)}{\prn*{1 + \alpha \theta \Fkl(w_0)^\theta \cdot t}^{1/\theta}}. 
\end{align*} 
Furthermore, \(\Rkl\) is an admissible rate of convergence w.r.t.~\(F\). 
\end{lemma}
Plugging the above rate function in  \pref{thm:gf_to_potential} with \(g(z) = \alpha z^{1 + \theta}\) implies that the function \(\Phi_g(w) = \Fkl(w)\) is an admissible potential function w.r.t.~\(\Fkl\). We can thus use this potential function in \pref{thm:GD_guarantee} and \pref{thm:SGD_guarantee} to provide a convergence guarantee for GD and SGD. We note that the following additional assumption that  \(\Fkl\) is \(H\)-smooth, is sufficient to derive the required self-bounding regularity conditions on \(\Fkl\) and \(\Phi_g\). 
 \begin{assumption} 
\label{ass:F_sb_KL} There exists an \(H \in \bbR^+\) such that 
$\nrm{\grad^2 \Fkl(w)} \leq H$ for any \(w\). 
\end{assumption}

\noindent
We now state the convergence bound for GD and SGD algorithm. 

\begin{theorem} 
\label{thm:KL_main} 
Suppose \(\Fkl\) is \(K\L\)~ with exponent \(\theta\) and coefficient \(\alpha\), and satisfies \pref{ass:F_sb_KL}. Then, for any initial point \(w_0\) and \(T \geq 1\), setting \(\eta\) appropriately, 
\begin{enumerate}[label=(\alph*)] 
\item The point \(\wh {w}_T\)  returned by GD algorithm satisfies \(\Fkl(\wh w_T) \lesssim \prn*{\frac{H \Fkl(w_0)}{\alpha}}^{{1}/{1 + \theta}} \cdot \frac{1}{T^{1/(2 + 2 \theta)}}. \)
\item The point \(\wh {w}_T\)  returned by SGD starting from \(w_0\) and using stochastic gradient estimates for which \pref{ass:gradient_noise} holds with \(\chi(z) = \sigma^2\), satisfies  $\Fkl(\wh w_T) \lesssim \prn*{\frac{BH^3 \Fkl(w_0)}{\alpha^2 T} }^{1/2 + 2 \theta}$ with probability at least \(0.7\). 
\end{enumerate}
\end{theorem}  
We first observe that both GD and SGD converge at the rate of at least \(O\prn*{{1}/{T^{1/2 + 2\theta}}}\). Furthermore, \(\theta = 0\) corresponds to the function being P\L~, in which case, we can improve the rate for GD (by extending \pref{lem:linear_to_exp}) to be of the form \(\Fkl(w(t)) \leq \Fkl(w_0) e^{- O(t)}\) which recovers the bound in \pref{prop:PL_complete}. We also note that the classical stochastic optimization setting in which \(\grad f_{\text{kl}}(w; z) = \grad \Fkl(w) + \varepsilon_t\) where \(\varepsilon_t\) is a sub-Gaussian random variable with mean \(0\) and variance \(\sigma^2\) satisfies \pref{ass:gradient_noise}. As a result we have convergence guarantees for SGD algorithm for this case. Finally, we note that similar to the results in \pref{sec:complete_proofs}, we have the following geometric equivalence between K\L~ functions and rates for GF. 

\begin{proposition} 
\label{prop:KL_complete} 
The following two properties are equivalent for any function \(F\): 
\begin{enumerate}[label=(\alph*)]  
\item For any \(w(0) \in \bbR^d\) and \(t \geq 0\), GF has the admissible rate \( F(w(t)) \leq \frac{F(w_0)}{\prn*{1 + \alpha \theta F(w_0)^\theta \cdot t}^{1/\theta}}\), 
\item  \(F(w)\) satisfies the Kurdyka-\L ojasiewicz (PL) property i.e. \(\alpha \Fkl(w)^{1 + \theta} \leq \nrm{\grad \Fkl(w)}^2  
 \), 
\end{enumerate} 
for any \(\alpha \geq 0\) and \(\theta \in (0, 1)\). \end{proposition} 

\subsection{Phase retrieval} \label{sec:PR}
In the phase retrieval problem \citep{candes2015phase, chen2019gradient, tan2019online}, we wish to reconstruct a hidden vector \(w^* \in \bbR^d\) with \(\nrm{w^*} = 1\) using phaseless observations \(\cS = \crl*{(a_j, y_j)}_{j \leq T}\) of the form \(y_j = \tri{a_j, w^*}^2\) where \(a_j \sim \cN(0, \mathrm{I}_d)\). The classical approach to recover \(w^*\) is by using the per-sample loss function \(f_\pr(w; (a_j, y_j)) = \prn{(a_j^\top w)^2 - y_j}^2\) for which the corresponding population loss is given by
\begin{align*} 
F_\pr(w) = \En \brk*{f_\pr(w; (a, y))} = \En_{a \sim \cN(0, \mathrm{I}_d)} \brk*{\prn*{(a^\top w)^2 - (a^\top w^*)^2}^2}.   \numberthis \label{eq:pr_loss} 
\end{align*} $F_\pr$ is non-convex, and has stationary points (and local minima) that do not correspond to the global minima. In the following, we  provide convergence guarantees for GD algorithm on \(F_\pr\), and SGD algorithms that computes stochastic gradient estimates using \(\cS\). We first note that \(F_\pr\) satisfies self-bounding regularity conditions, and GF on \(F_\pr\) converges to the global minimizer for any initial point \(w_0\).

\begin{lemma} 
\label{lem:F_gf_pr} 
\(F_\pr\) satisfies \pref{ass:F_regular}. Furthermore, for any initial point \(w_0\), the point \(w(t)\) on its gradient flow path satisfies  
\begin{align*}
F_\pr(w(t)) \leq \min \crl*{F_\pr(w_0), F_\pr(w_0) e^{-t + \frac{1}{\tri{w_0, w^*}^2}}}  \rdef{} R_{\pr}(w(0), t). 
\end{align*} 
Furthermore, the function \(R_\pr\) above is an admissible rate of convergence w.r.t.~\(F_{\pr}\). 
\end{lemma} 

The above rate follows from independently analyzing the parallel and perpendicular components \(\tri*{w, w^*}\) and \(\nrm{w}^2 - \tri{w, w^*}\) respectively. Our main tool for getting the convergence guarantee for GD / SGD is to utilize   \pref{thm:gf_to_potential} to get an admissible potential w.r.t.~\(F_\pr\), which can be plugged in  \pref{thm:GD_guarantee} and \ref{thm:SGD_guarantee} to get the corresponding rates. 
\begin{theorem} 
\label{thm:pr_main} 
Consider the phase retrieval objective \(F_\pr\) given in \pref{eq:pr_loss}. For any initial point \(w_0\) and \(T \geq 1\), setting \(\eta\) appropriately, 
\begin{enumerate}[label=(\alph*), leftmargin=8mm] 
\item The point \(\wh {w}_T\)  returned by GD starting from \(w_0\) satisfies  $F_\pr(\wh w_T) = O\prn*{\min\crl*{\frac{1}{T}, e^{- O(T- t_0)}}}$ for all \(T \geq t_0\),  where \(t_0\) is a \(w_0\) dependent constant. 
\item The point \(\wh {w}_T\)  returned by SGD starting from \(w_0\) and using stochastic gradient estimates for which \pref{ass:gradient_noise} holds, satisfies  $F_\pr(\wh w_T) = \wt O\prn*{\tfrac{1}{\sqrt{T}}}$ with probability at least \(0.7\).  
\end{enumerate} 
\end{theorem} 
The \(O(\cdot)\) notation above hides \(w_0\) dependent constants which we specify in the Appendix. Our rate for GD above matches the best known result in the literature in terms of the dependence on \(T\) \citep{chen2019gradient}. To the best of our knowledge, ours is also the first convergence analysis of SGD under arbitrary noise conditions satisfying \pref{ass:gradient_noise}. While this rate is optimal under certain noise conditions, e.g. when \(\chi(z) = \sigma^2\),  further improvements are possible when \(\chi\) is favorable. For example, suppose the stochastic gradient estimates were computed using samples from \(\cS\) by taking a fresh sample for each estimate, i.e. \(\grad f(w; (a, y)) = 4\prn{(a^\top w)^2 - y} (a^\top w)w\). In this case, the stochastic gradient satisfy  \pref{ass:gradient_noise} with \(\chi(z) = \min\crl{\sqrt{z}, c}\) where $c$ is a universal constant (c.f. \citet[Lemma 7.4, 7.7]{candes2015phase}). While, our framework implies that this SGD algorithm (computing estimates using samples)  converges at the rate of \(1/\sqrt{T}\), this rate can be  improved further \citep{chen2019gradient}, and we defer the refined analysis for future research. 

\subsection{Initialization specific rates} 
In many applications, GF is only known to converge from nice enough initial points that satisfy certain properties. In this section, we extend show how to use our tools for establishing convergence of GD/SGD for such problems, and consider matrix square root as an example. We first provide the following general utility lemma that shows how to construct admissible potentials when the rate for GF from \(w_0\) holds only when \(w_0\) satisfies a certain property characterized by \(h(w_0) \geq 0\). 

\begin{lemma}\label{lem:indicator_utility_lemma}
Let \(h: \bbR^d \mapsto [0, 1]\) be a continuously differentiable function, and suppose that for any point \(w\) for which \(h(w) > 0\), GF with \(w(0) = w\) has rate \(F(w(t)) \leq R(w, t)\) where \(R(w, \cdot)\) is a monotonically decreasing function in \(t\). Furthermore, suppose that \(F(w) \leq R(w, 0)\), \(F\) satisfies \pref{ass:F_regular}, $R(w, h(w) t)$ is an admissible rate function w.r.t.~\(F\), and for any w, 
 \begin{enumerate}[label=$(\alph*)$, leftmargin=8mm] 
 \item the function \(\Gamma(w) \ldef{} \int_{t=0}^\infty R(w, t) \dif t\) is continuously differentiable, and \(\max\crl{\nrm{\grad \Gamma(w)}, \nrm{\grad^2 \Gamma(w)}} \leq \lambda(\Gamma(w))\) where \(\lambda\) is a positive, monotonically increasing function.  
\item  \(\max\crl{\nrm{\grad h(w)}, \nrm{\grad^2 h(w)}} \leq \pi(\Gamma(w))\) where \(\pi\) is a positive, monotonically increasing function.  
\item  \(\prn{h(w) - h(w^*)}^2 \leq \mu(\Gamma(w))\) where \(\mu\) is a positive, monotonically increasing function with the property that \(k \mu(z) \leq \mu(k z)\) for any \(k \geq 1\). 
 \end{enumerate} 
Then, the function \(\Phi_g(w) = {\Gamma(w)}/{h(w)}\) is an admissible potential  w.r.t.~\(F\) with \(g(z) = z\), and satisfies the self-bounding regularity condition in \pref{ass:Phi_regular}. 
\end{lemma} 

While the conditions (a), (b) and (c) above are technical, we note that they are easily satisfied for many problems of interest, e.g. Matrix square root. At an intuitive level, these conditions ensure that the function \(\Phi_g\) is an admissible potential and satisfies the desired prerequisites for \pref{thm:GD_guarantee} and \ref{thm:SGD_guarantee}. The proof details are deferred to the Appendix. 

\subsubsection{Matrix square root} \label{sec:MS} 
In the matrix square root problem \citep{de2015global, jain2017global}, we are given a positive definite and symmetric matrix \(M \in \bbR^{d \times d}\) with \(\smin(M) > 0\), and wish to find a symmetric \(W \in \bbR^{d \times d}\) that minimizes the objective 
\begin{align*} 
F_\ms(W) = \nrm{M - W^2}_F^2. \numberthis \label{eq:matrix_sqrt_loss} 
\end{align*}  

$F_\ms$ is non-convex in \(W\), and has spurious stationary points. In the following, we  provide convergence guarantees for GD/SGD algorithm on \(F_\ms\). We first note that \(F_\ms\) satisfies self-bounding regularity conditions, and GF on \(F_\ms\) converges to the global minimizer when the initial point \(w_0\) satisfies additional assumptions. We capture these initial conditions using the function \(h_\ms\) defined as 
\begin{align*}
h_\ms(W) &=  \sigma\prn*{\phi(W^2) - \alpha},   \numberthis \label{eq:h_defn_msqrt_main}
\end{align*} where the function \( \phi(Z) \ldef{} \frac{-1}{\gamma} \log(\trace{e^{- \gamma Z}} + e^{- 16 \alpha \gamma })\), \(\alpha = \sigma_{\min}(M)/1600\), \(\gamma = \log(d + 1)/\alpha\), and 
\(\sigma\) denotes a smoothened version of the indicator function given by  \(\sigma(z) = \crl{0~\text{if}~ z \leq 0, ~\tfrac{2}{\alpha^2} z^2~\text{if}~ 0 \leq z \leq \tfrac{\alpha}{2},~ - \tfrac{2}{\alpha^2} z^2 + \tfrac{4}{\alpha} z - 1~ \text{if}~  \tfrac{\alpha}{2} \leq z  \leq \alpha,~\text{and}~	1~ \text{if}~ \alpha \leq z }\). 

\begin{lemma}
\label{lem:F_gf_matrix_completion}  \(F_\ms\) satisfies \pref{ass:F_regular}. Furthermore, for any initial point \(W_0\) for which \(h_\ms(W_0) > 0\), the point \(W(t)\) on its GF path satisfies 
\begin{align*}
F_\ms(W(t)) \leq F_\ms(W_0) \exp\prn*{- 16 \alpha t} 
\rdef{} R_\ms(W(0), t), 
\end{align*}  
where  \(\alpha = \sigma_{\min}(M)/1600\), \(\gamma = \log(d + 1)/\alpha\) and the function \(h_\ms\) is defined in \pref{eq:h_defn_msqrt_main}. 
\end{lemma} 

The above rate follows from directly solving the  PDE  associated with the gradient flow on the underlying objective.  \pref{lem:F_gf_matrix_completion}  provides conditions on \(W_0\) under which the GF path converges with the rate function \(R_\ms\). Our main tool for showing the convergence of GD / SGD is by using \pref{lem:indicator_utility_lemma} to get admissible potentials.  Note that the function \(h_\ms\) takes values in \([0, 1]\), is continuously differentiable, and as we show in the appendix satisfies all the required self-bounding regularity conditions in \pref{lem:indicator_utility_lemma}. Thus, \pref{lem:indicator_utility_lemma} provides an admissible potential w.r.t.~ \(F_\pr\) which can be used to get the following rates. 

\begin{theorem}
\label{thm:ms_main} Consider the matrix square root objective \(F_\ms\) given in \pref{eq:matrix_sqrt_loss}. For any \(\kappa > 0\), initial point \(W_0\) for which \(h_\ms(W_0) > 0\) and setting \(\eta\) appropriately, 
\begin{enumerate}[label=(\alph*), leftmargin=8mm] 
\item The point \(\wh {W}_T\)  returned by GD starting from \(W_0\) satisfies  $F_\ms(\wh W_T) = O\prn*{\min\crl*{\frac{1}{T}, e^{- O(T- t_0)}}}$ for all \(T \geq t_0\),  where \(t_0\) is a \(w_0\) dependent constant. 
\item The point \(\wh {W}_T\)  returned by SGD starting from \(W_0\) and using stochastic gradient estimates for which \pref{ass:gradient_noise} holds, satisfies   $F_\ms(\wh W_T) = \wt{O}\prn*{ \frac{1}{\sqrt{T}}}$ with probability at least \(0.7\). 
\end{enumerate} 
\end{theorem} 

The \(O(\cdot)\) notation above hides \(W_0\) dependent constants which we specify in the Appendix. Our rate for GD above matches the best known result in the literature in terms of the dependence on \(T\)  \citep{jain2017global}. Ours is also the  first convergence analysis of SGD under arbitrary noise conditions satisfying \pref{ass:gradient_noise}. Note that the classical stochastic optimization setting in which \(\grad f_\ms(w; z) = 2(W^2 - M) W  + 2W (W^2 - M) + \varepsilon_t\) where \(\varepsilon_t\) is a sub-Gaussian random variable with mean \(0\) and variance \(\sigma^2\) satisfies \pref{ass:gradient_noise} with \(\chi(z) = \sigma^2\), and as a result of \pref{thm:ms_main}, we get that SGD converges at the rate of \(1/\sqrt{T}\). To the best of our knowledge, convergence of SGD in the stochastic optimization setting for matrix square root problem was not known before. 

\subsection{Extending \cite{chatterjee2022convergence}} 
If the objective $F$ is such that some potential $\Phi_g$ satisfies the geometric condition in \pref{eq:linearity_property} for every $w$, then we have a rate of convergence for GF (\pref{thm:potential_to_gf}). As we saw earlier, for instance, using this machinery one can obtain rates for GF/GD/SGD when \(F\) has P\L~ property everywhere. However, such global properties, that \pref{eq:linearity_property} holds for every $w$ are often too stringent to hold in practice. In order to go beyond global assumption, in \pref{lem:indicator_utility_lemma}  we showed how to extend our tools (by defining corresponding admissible potentials) when such properties (and thus rates for GF) only hold in some region. Convergence under such local properties has also been considered before in other works \cite{chen2019gradient, du2018gradient, jin2016provable, mohammadi2019global, vardi2021implicit, jain2017global, ma2018implicit}. However, all of these results usually rely on being able to choose an initialization \(w_0\) in the good region, where the corresponding local property holds, and is close enough to the global minima that we wish to converge to. This is not always practical, and to circumvent this issue in a recent work of \citet{chatterjee2022convergence}, an assumption that is ``local'' w.r.t. initial point is provided under which one can show that GF and GD starting from this initialization is guaranteed to converge (at an exponential rate).  The interesting property of this condition is that it is local to initial point \(w_0\) considered and does not make any global assumption on the objective. 

Using the tools in this paper, this type of local property can be easily extended to more general properties than what was considered in \citet{chatterjee2022convergence}. For ease of presentation, we present below the result for $H$-smooth objective $F$ and for GF convergence, the corresponding techniques can be easily extended show GD/SGD convergence when \pref{ass:F_regular} holds. 
Given a function $r:\reals^d \mapsto \reals^+$ and a monotonically increasing positive function $g$, define 
\begin{align}\label{eq:sauravalpha}
    \alpha_{r,g}(w_0,\kappa) = \inf_{w : \|w - w_0\|_2 \le \kappa, F(w) \ne 0} \frac{\nabla r(w)^\top \nabla F(w)}{g(F(w))}
\end{align}
Our main assumption on the initial point $w_0$ is that for some $\kappa >0$ and some functions $R$ and $g$, 
\begin{align}\label{eq:saurav}
\int_{0}^\infty \sqrt{g^{-1}\left(\frac{r(w_0)}{t \alpha_{r,g}(w_0,\kappa)}\right) dt}\le \frac{\kappa}{H}
\end{align}

The next lemma shows that for any initial point $w_0$ that satisfies the local condition above, one has a rate of convergence for GF starting from \(w_0\). 
\begin{lemma}\label{lem:saurav} 
Suppose $w_0$ satisfies \pref{eq:saurav} for some functions $R$ and $g$, and radius $\kappa = \kappa_0> 0$. Then, gradient flow starting from $w(0) = w_0$ satisfies for any \(t \geq 0\), 
$$
F(w(t)) \le g^{-1}\left(\frac{r(w_0)}{T \alpha(w_0,\kappa_0)} \right). 
$$ 
\end{lemma}

To obtain nearly matching rates for the type of condition in \citet{chatterjee2022convergence}, one can choose $r(w) = p \cdot F(w)^{1/p}$ and $g(z) = z^{1/p}$. Since $p$ is arbitrary, setting $p = T \alpha(w_0,\kappa_0)/e $ we obtain nearly the same rate and the local condition as \citet{chatterjee2022convergence} (upto constants). The interesting part though, is that this is for only one choice of $g$ and $r$, whereas we can get the convergence for GF when the condition holds for any $g,r$. In \citet{chatterjee2022convergence}, examples of overparmeterized deep neural nets are shown to satisfy the assumption (for the specific $r$ and $g$ above). With a wider choice of $g$ and $r$ we can extend these to more general models (eg. neural networks with milder assumptions on the activation function).

%% file: files/main_extensions.tex
In this paper, we provide a new framework for establishing performance guarantees for SGD in stochastic non-convex optimization. We introduce admissible potentials, and use them to get finite-time convergence guarantees for SGD. We also provide a method for constructing such  admissible potentials using the rate function with which  gradient flow converges on the underlying non-convex objective, provided that this rate function satisfies additional admissibility conditions. Thus, informally speaking, our results suggest that whenever gradient flow has an admissible rate of convergence and additional regularity conditions hold,  SGD succeeds in minimizing the underlying non-convex objective (with the rate given in \pref{thm:SGD_guarantee}). In the following, we discuss some extensions and open problems:  
\begin{itemize}
\item Contrary to the prior approaches \citep{gunasekar2021mirrorless, NIPS2015_f60bb6bb, JMLR:v22:20-195, su2014differential, zhang2021revisiting, elkabetz2021continuous}, our convergence proof for SGD does not proceed by showing that the corresponding paths of  SGD and gradient flow dynamics are point-wise close to each other. In fact, the example in \pref{thm:f_Phi_regularity}  suggests that this may not be true even for convex functions, since for that example, gradient flow converges to minimizers but SGD diverges away from good solution.  Our key technique is to use admissible potentials, that satisfy \pref{eq:linearity_property} w.r.t.~gradient flow dynamics, to analyze discrete time algorithms like SGD. 

\item Our framework is motivated by Lyapunov analysis of dynamical systems \citep{cencini2013lyapunov, chellaboina2008nonlinear, clarke2004lyapunov, wilson2021lyapunov}. The property \pref{eq:linearity_property} in fact implies that any admissible potential is a Lyapunov potential w.r.t.~the gradient flow dynamics on the underlying non-convex loss. It would be interesting to explore if techniques from the Lyapunov analysis of dynamical systems can be used to improve our rates further, or to relax various regularity and admissibility assumptions that we assume for our results. In particular, it would be interesting to explore how to extend our framework for non-smooth non-convex stochastic optimization.   

\item While we restricted ourselves to GD in the paper, our framework can be easily extended to analyze mirror descent algorithms (to get improved dependence on the problem geometry), by modifying the admissibility condition \pref{eq:linearity_property} to hold w.r.t.~ gradient flow dynamics in the dual space (mirror space). Furthermore, we can also extend our framework to other first-order algorithms like acceleration, momentum, etc., by changing \pref{eq:linearity_property} to hold w.r.t.~the corresponding continuous time dynamics for these algorithms \citep{kovachki2021continuous, su2014differential, orvieto2019continuous}. 

\item \pref{thm:gf_to_potential} gives a construction of admissible potentials using the rate function \(R\) for gradient flow on the underlying objective. However, the convergence bound for SGD in \pref{thm:SGD_guarantee} holds only when this constructed potential satisfies additional self-bounded regularity conditions in \pref{ass:Phi_regular}. In order to get an end-to-end result, it would be interesting to explore what structural conditions on the rate function \(R\) implies that the obtained potential satisfies \pref{ass:Phi_regular}. 
\end{itemize}

In the paper, we demonstrate the generality of our framework by considering various non-convex stochastic optimization problems including P{\L}/K{\L} functions, phase retrieval and matrix square root, and show that admissible rate functions and the corresponding admissible potentials can be easily obtained by explicitly solving the partial differential equation associated with gradient flow; hence getting rates of convergence for SGD for these problems. Looking forward, it would be interesting to apply our framework for other non-convex stochastic optimization problems appearing in machine learning, and in particular deep learning.

%% file: files/appx_basic.tex
In  the following, we provide some basic definitions, probabilistic inequalities, and technical results. 
\begin{definition}[\(L\)-Lipschitz function] A function \(F:\bbR^d \mapsto \bbR\) is said to be \(L\)-Lipschitz if for any \(w_1, w_2\), \(\abs{F(w_1) - F(w_2)} \leq L \nrm{w_1 - w_2}.\)
\end{definition}

\begin{definition}[\(H\)-smooth functions] A differentiable function \(F:\bbR^d \mapsto \bbR\) is said to be \(H\)-Lipschitz if for any \(w_1, w_2\),
\begin{align*}
F(w_2) \leq F(w_1) + \tri{\grad F(w_1), w_2 - w_1} + \frac{H}{2} \nrm{w_2 - w_1}^2. 
\end{align*}
\end{definition} 

\begin{definition}[$\lambda$-Linearizable] A function $F(w)$ is  $\lambda$-Linearizable if there exists a $w^* \in \argmin F(w)$ such that for every point $w \in \bbR^d$, 
\begin{equation*}
		F(w) - F(w^*) \leq \lambda \tri*{\grad F(w), w - w^*}. 
\end{equation*}
\end{definition} 

\begin{lemma}[Azuma's inequality] 
\label{lem:Azuma2} 
Let \(\crl*{X_t}_{t \geq 0}\) be a super-martingale sequence such that for any \(t \geq 0\), \(A_t \leq X_{t+1} - X_{t} \leq B_t\) where \(A_t\) and \(B_t\) are \(\cF_t\)-measurable, and satisfy \(\abs*{B_t - A_t} \leq c_t\). Then, for any \(\gamma > 0\), 
\begin{align*}
\Pr\prn*{X_t - X_0 \geq \gamma} &\leq \exp\prn*{- \frac{\gamma^2}{2 \sum_{t=1}^n c^2_t}}. 
\end{align*}
\end{lemma}

The next technical lemma shows that \(F(w(t))\) monotonically decreases along any GF path. 
\begin{lemma} 
\label{lem:gf_decrease} 
Let \(w_0\) be any initial point. Then, for any \(t \geq 0\), the point \(w(t)\) on the GF path with \(w(0) = w_0\) satisfies \(F(w(t)) \leq F(w(0))\). 
\end{lemma}
\begin{proof} Fix \(w(0) = w_0\) and define the function \(\ls(t) = F(w(t))\), where \(w(t)\) is on the GF path from \(w_0\) at time \(t\). Using Chain rule, we note that  
\begin{align*}
\frac{\dif g(t)}{\dif t} &=  \tri{\grad F(w(t)), \frac{\dif w(t)}{\dif t}} = - \nrm*{\grad F(w(t))}^2, 
\end{align*} where the last equality holds from the definition of GF in \pref{eq:GF}. The above implies that \(g(t) = F(w(t))\) is monotonically increasing with \(t\). 
\end{proof}

\begin{lemma}
\label{lem:linear_to_exp} 
Suppose starting from any initial point \(w(0)\) and for any \(t \geq 0\), the point \(w(t)\) on the GF path satisfies 
\begin{align*}
F(w(t)) \leq \frac{F(w(0))}{\lambda t}.
\end{align*} 
Then, we have that for any \(w(0)\) and \(t \geq 1\), 
\begin{align*}
F(w(t)) \leq F(w(0)) e^{- \floor*{{\lambda t}/{e}}}. 
\end{align*}. 
\end{lemma}
\begin{proof} Fix any \(t \geq e\) and divide \([0, t]\) into \(k =  \floor*{\lambda t/e}\) many chunks of size $e/\lambda$ each. Let this partition be \([0, t_1, \dots, t_{k} = t]\). Clearly, we have that for any \(j \leq k\), the point \(w(t_j)\) corresponds to the point at time \(e/\lambda\) on the GF path starting from  \(w(t_{j-1})\). The given rate assumption thus implies that 
\begin{align*}  
F(w(t_j)) \leq \frac{F(w(t_{j-1}))}{e}. 
\end{align*} 
Recursing the above for \(j\) from \(1\) to \(k\), we get that 
\begin{align*}
F(w(t)) = F(w(t_k)) \leq e^{-k} F(w(0)) =  F(w(0)) e^{- \floor*{{\lambda t}/{e}}}
\end{align*}. 
\end{proof}

\begin{lemma}[Lemma 2.1, \cite{NIPS2010_76cf99d3}]  
	\label{lem:smoothness_inequality} 
For any \(H\) smooth function \(F: \bbR^d \mapsto \bbR\), for any \(x \in \bbR^d\), 
\begin{align*}
\nrm*{\grad F(x)} &\leq \sqrt{4 H \prn*{F(x) - F^*}},
\end{align*} 
where \(F^* \ldef{} \min_x F(x)\), 
\end{lemma}

%% file: files/appx_gf.tex
\begin{proof}[Proof of \pref{thm:potential_to_gf}] 
Let \(w(s)\) be the point on the GF path after time \(s\) when starting from the point \(w(0)\). An application of chain rule implies that  
\begin{align*}
\frac{\dif \Phi(w(s))}{\dif s} &= \tri*{\grad \Phi(w(s)), \frac{\dif w(s)}{\dif t}} \\
&= \tri*{\grad \Phi(w(s)), - \grad F(w(s))} \\ 
&\leq - g(F(w(s))), 
\end{align*} where the equality in the second line holds by the update rule of GF, i.e. \(\frac{\dif w(s)}{\dif s} = - \grad F(w(s))\) and the last line follows by using \pref{def:Phi} where \(g\) is a monotonically increasing function that satisfies \pref{eq:linearity_property}. Rearranging the terms and integrating both the sides for \(s\) from \(0\) to \(t\), we get 
\begin{align*}
\int_{s=0}^t g(F(w(s))) \dif s \leq - \int_{s=0}^t \frac{\dif \Phi(w(s))}{\dif s}  \dif s = \Phi(w(0))  - \Phi(w(s)) \leq \Phi(w(0)),  \numberthis \label{eq:potential_to_gf}
\end{align*} where the last inequality in the above holds because \(\Phi(\cdot) \geq 0\) by definition. 

We finally conclude by noting that \(F(w(t))\) is a decreasing function of \(t\) since
\begin{align*}
\frac{\dif F(w(t))}{\dif t} = \tri*{\grad F(w(t)), \frac{\dif t(w)}{\dif t}} &= - \tri*{\grad F(w(t)),\grad F(w(t))} \leq 0, 
\end{align*} where the second equality above follows from GF update rule. Since \(g\) is a monotonically increasing function, the above implies that \(g(F(w(t))) \leq g(F(w(s)))\) for all \(s \leq t\). Using this relation in \pref{eq:potential_to_gf} implies that 
\begin{align*}
g(F(w(t))) \cdot t \leq \int_{s=0}^t g(F(w(s))) \dif s \leq \Phi(w(0)). 
\end{align*}
Rearranging the terms gives the desired relation. 
\end{proof}

\begin{proof}[Proof of \pref{thm:gf_to_potential}] The following proof uses the most general conditions for admissibility of \(R\) stated in \pref{def:rate}. Let \(w \in \clo(W)\) be any initial point.  Since \(\int_{t=0}^\infty g(R(w, t)) \dif t < \infty\) and \(\int_{t=0}^\infty g'(R(w, t)) \nrm{\grad R(w, t)} \dif t  < \infty\) for every \(w \in \mathrm{clo}(\cW)\), the function \(\Phi_g\) is well defined and is differentiable along the gradient flow path at the point \(w\). Additionally, in the following \(w(t)\) denotes the point at time \(t\) on the GF path starting from \(w\). 

First, note that because \(F(w(t)) \leq R(w, t)\), and \(g\) is positive and monotonically increasing, we have 
\begin{align*}
g(F(w)) = g(F(w(0))) &\leq g(R(w, 0)) \\ 
&= - \int_{t=0}^\infty  \frac{\partial g(R(w, t))}{\partial t} \dif t \\
&= - \int_{t=0}^\infty  g'(R(w, t)) \frac{\partial R(w, t)}{\partial t} \dif t \\
&\leq \int_{t=0}^\infty g'(R(w, t)) \tri{\grad R(w, t), \grad F(w)} \dif t 
\end{align*} where the first equality is a tautology since \(\lim_{t \rightarrow \infty} g(R(w, t)) = 0\), and the second equality follows from Chain rule. The inequality in the last line uses the property \pref{def:rate}-(b). Next, note that 
\begin{align*}
\int_{t=0}^\infty g'(R(w, t)) \tri{\grad R(w, t), \grad F(w)} \dif t &= \lim_{s \rightarrow 0^+} \int_{t=0}^\infty g'(R(w(s), t)) \tri{\grad R(w(s), t), \grad F(w(s))} \dif t \\
&= \lim_{s \rightarrow 0^+} \int_{t=0}^\infty \frac{\partial R(w(s), t)}{\partial s} \dif t \\
&= \lim_{s \rightarrow 0^+} \frac{\partial }{\partial s} \int_{t=0}^\infty g(R(w(s), t))  \dif t, 
\end{align*} where the equality in the second line above holds due to Chain rule and the last line follows from interchanging the integral and the derivative, which is permissible since we have that  \(\int_{t=0}^\infty g(R(w(s), t))  \dif t < \infty\) for \(w(s) \in \clo(W)\). Finally, note that 
\begin{align*}
\lim_{s \rightarrow 0^+} \frac{\partial }{\partial s} \int_{t=0}^\infty g(R(w(s), t))  \dif t = \lim_{s \rightarrow 0^+} \frac{\partial }{\partial s} \Phi_g(w(s)) &= \tri{\grad \Phi_g(w(0)), \grad F(w(0))} 
\end{align*} where the first equality uses the definition of \(\Phi_g\)  and the second equality is due to Chain rule. 

Combining the above chain of inequalities and plugging in \(w(0) = w\) implies the desired condition,  
\begin{align*}
\tri{\grad \Phi_g(w), \grad F(w)}  &\geq g(F(w)). 
\end{align*}
\end{proof}

\begin{proof}[Proof of \pref{corr:converse_gf_tight}] Define \(H = \max_{w \in \cW} h(w)\), and the function \(g\) as 
\begin{align*}
g(z) = \frac{1}{\sigma(z/H) \log^2(\sigma(z/H))}, 
\end{align*} where the function \(\sigma\) is defined as \(\sigma(x) = e + r^{-1}(x)\). Using the above \(g\) in \pref{thm:gf_to_potential}, we get the potential 
\begin{align*}
\Phi(w) &= \int_{t=0}^\infty \frac{1}{\prn*{\sigma\prn*{\frac{h(w)}{H}r(t)}} \log^2 \prn*{\sigma\prn*{\frac{h(w)}{H}r(t)}}} \dif t. 
\end{align*}

The potential satisfies  
\begin{align*} 
\Phi(w) &\leq  \int_{t=0}^\infty \frac{1}{\prn*{\sigma\prn*{r(t)}} \log^2 \prn*{\sigma\prn*{r(t)}}} \dif t \\
&= \int_{t=0}^\infty \frac{1}{\prn*{e + t} \log^2 \prn*{e + t}} \dif t = 1, \numberthis \label{eq:corr_bound1} 
\end{align*} 
where the first inequality holds because \(h(w)/H \leq 1\) and since \(\sigma\) is inverse of \(r\), it has to be monotonically decreasing. 

In addition to the above, we also note that 
\begin{align*}
\int_{t=0}^\infty g'(R(w, t)) \nrm{\grad R(w, t)} \dif t &\leq \int_{t=0}^\infty \frac{3}{\sigma\prn*{\frac{h(w)}{H} r(t)}^2 \log^2\prn*{\sigma\prn*{\frac{h(w)}{H} r(t)}}} \sigma'\prn*{\frac{h(w)}{H} r(t)} \frac{\nrm{\grad h(w)}}{H} r(t) \dif t  \\ 
&= \int_{t=0}^\infty \frac{3}{\sigma\prn*{\frac{h(w)}{H} r(t)}^2 \log^2\prn*{\sigma\prn*{\frac{h(w)}{H} r(t)}}} \frac{1}{r'\prn*{\sigma\prn*{\frac{h(w)}{H} r(t)}}} \frac{\nrm{\grad h(w)}}{H} r(t) \dif t \\
&\leq \int_{t=0}^\infty \frac{3}{\sigma\prn*{\frac{h(w)}{H} r(t)} \log^2\prn*{\sigma\prn*{\frac{h(w)}{H} r(t)}}} \frac{c}{r\prn*{\sigma\prn*{\frac{h(w)}{H} r(t)}}} \frac{\nrm{\grad h(w)}}{H} r(t) \dif t \\ 
&\leq \frac{3c \nrm{\grad h(w)}}{h(w)} \int_{t=0}^\infty \frac{1}{\sigma\prn*{\frac{h(w)}{H} r(t)} \log^2\prn*{\sigma\prn*{\frac{h(w)}{h(w)} r(t)}}} \dif t \\ 
&\leq  \frac{3c \nrm{\grad h(w)}}{H}  < \infty, 
\end{align*} where the first inequality is from Chain rule and a trivial algebraic upper bound. The second inequality uses the relation that \(r(t) \leq c \abs{r'(t)} t\) for any \(t \geq 0\). The third inequality uses the fact that \(r\) is monotonically decreasing and that \(\sigma\) is the inverse of \(r\), and the last line follows similar to the bound in \pref{eq:corr_bound1}. Thus, \(g\) is a valid function and \(\Phi\) defined above is an admissible potential. Using \pref{thm:potential_to_gf}, we get that 
\begin{align*}
g(F(w(t))) &\leq \frac{\Phi(w(0))}{t} \leq \frac{1}{t}. 
\end{align*}
Rearranging the terms, we get 
\begin{align*}
\sigma\prn*{\frac{F(w)}{H}} \geq \frac{t}{\log^2(t)}.  
\end{align*} 
Using the fact that \(\sigma(x) = r^{-1}(x)\) in the above, we get that 
\begin{align*} 
F(w) \leq H r \prn*{{t}/{\log^2(t)}}.  
\end{align*} 
\end{proof}  

\begin{proof}[Proof of \pref{prop:PL_complete}] ~ We prove the forward and reverse direction as follows: 
\begin{enumerate}[label=(\alph*), leftmargin=8mm] 
\item \textit{Proof of \((a) \Rightarrow (b)\).}  First note that  \(R(w, t) = F(w) 2^{\lambda t}\) is an admissible rate function for \(F\). Clearly, it is a decreasing function of \(t\) and \(\lim_{t \rightarrow \infty} R(w, t) = 0\) for any \(w\). Furthermore, note that for \(w(t)\) on the GF path of \(w(0)\), we have 
\begin{align*}
R(w(t), 0) = F(w(t)) \leq F(w(0)) e^{- \lambda t} = R(w(0), t), 
\end{align*}
where the inequality follows from the rate assumption. Thus, $R$ satisfies all the conditions in  \pref{def:rate}. Thus,  invoking \pref{thm:gf_to_potential} with \(g(z) = z\), we get that 
\begin{align*}
\Phi(w) &= \int_{t=0}^\infty R(w, t)  \dif t =  \int_{t=0}^\infty F(w) e^{-\lambda t} \dif t = \frac{F(w)}{\lambda} 
\end{align*} 
is an admissible potential for \(F\). Thus, from \pref{eq:linearity_property}, we get that 
\begin{align*}
\frac{\nrm*{\grad F(w)}^2}{\lambda} = \tri*{\grad \Phi(w), \grad F(w)} \geq F(w), 
\end{align*} 
which implies the desired P\L~ property. 

\item \textit{Proof of \((b) \Rightarrow (a)\).} This follows by directly solving the corresponding differential equation along the GF path. Consider the potential function \(\Phi(w) = \frac{F(w)}{\lambda}\). Note that \(\Phi\) is positive, and due to the P\L~ property, satisfies \pref{eq:linearity_property}. Thus, \(\Phi\) is an admissible potential w.r.t.~\(F\). Let \(w(0)\) be the initial point for GF, we note that at the point \(w(t)\) on its GF path, 
\begin{align*}
\frac{\dif \Phi(w(t))}{\dif (t)} &= \tri*{\grad \Phi(w(t)), \frac{\dif w(t)}{\dif t}} \\ 
&= - \tri*{\grad \Phi(w(t)), \grad F(w(t))} \\ 
&= - \frac{1}{\lambda} \nrm{\grad F(w(t))}^2 \\ 
&\leq - F(w(t)), 
\end{align*} where the last line follows from the P\L~ property. Plugging in the definition of \(\Phi\) in the above, we get 
\begin{align*}
\frac{\dif F(w(t))}{F(w(t))} \leq - \lambda. 
\end{align*}
The above differential equation in \(F\) has the following solution
\begin{align*}
F(w(t)) \leq F(w(0))e^{- \lambda t}. 
\end{align*}  Since the above holds for any \(w(0)\), \((a)\) immediately follows. 
\end{enumerate}
\end{proof} 

\begin{proof}[Proof of \pref{prop:convexity_complete}] We prove the forward and reverse direction as follows: 
\begin{enumerate}
\item \textit{Proof of \((a) \Rightarrow (b)\)} Since the rate is admissible, we must have that 
\begin{align*}
F(w) &\leq \lim_{t \rightarrow 0} R(w, t) \\ 
&\leq \lambda \lim_{t \rightarrow 0} \frac{\nrm{w - w^*}^2 - \nrm{w(t) - w^*}^2}{t} \\
&= \lambda \tri*{w - w^*, \grad F(w)}.
\end{align*}

\item \textit{Proof of \((b) \Rightarrow (a)\).} Clearly, 
\(\Phi(w) = \lambda \nrm*{w - w^*} / 2 \) is an admissible potential w.r.t.~F since \(\Phi(w) \geq 0\) and 
\begin{align*} 
\tri*{\grad \Phi(w), \grad F(w) } = \lambda \tri*{\grad F(w), w - w^*} \geq F(w), 
\end{align*}
where the last inequality holds because \(F\) is Linearizable. Thus, from \pref{thm:potential_to_gf} we get that for any initialization \(w(0)\), the point \(w(t)\) on its GF path satisfies 
\begin{align*}
F(w(t)) \leq \frac{\Phi(w(0)) - \Phi(w(t))}{t} = \lambda \frac{ \nrm{w(0) - w^*}^2 - \nrm{w(t) - w^*}^2}{2t}. 
\end{align*}  
\end{enumerate} 
\end{proof}

%% file: files/appx_ub1.tex
\begin{proof}[Proof of \pref{thm:f_Phi_regularity}] Fix any \(T_0 > 0\) and set \(d = (3T_0/2)^3\). Denote the variable \(u = w[1:d-1]\) and \(v = w[d]\), i.e. \(w = (u, v)\) and consider the function 
\begin{align*}
F(w) = \frac{1}{2}\nrm{u}^2_{3/2} + g(v), 
\end{align*} 
where  
\begin{align*}
\nrm{u}_{3/2} = \prn*{\sum_{i=1}^{d-1} u[i]^{3/2}}^{2/3} \qquad \text{and} \qquad \qquad g(v) &= \begin{cases}
	v^2 & \text{if} \quad  \abs{v} \leq 1/2 \\
\abs{v} - \frac{1}{4} & \text{if} \quad \abs{v} \geq 1/2
\end{cases}. 
\end{align*} 
Note that the \(\min F(w)\) is attained at the point \(w = 0\) and  
\begin{align*} 
\grad F(w)[i] &= \begin{cases}
	\sqrt{\nrm{u}_{3/2} \cdot u[i]} \cdot \sign{u[i]} & \text{for \(1 \leq i \leq d - 1\)} \\ 
	\sign{v[i]} & \text{for \(i = 1\) and \(\abs{v} \geq \frac{1}{2}\)} \\
	2v[i] & \text{for \(i = 1\) and \(\abs{v} \leq \frac{1}{2}\)}. 
\end{cases}. 
\end{align*}

We first argue that gradient flow converges at a rate of \(O(1/t)\) for any initial point \(w_0\). This follows from the fact that \(f(w)\) is convex in \(w\) and thus \(\Phi(w) = \nrm{w}^2/2\) is a valid potential function that satisfies for any time \(t\),  
\begin{align*}
\frac{\dif \Phi(w(t))}{\dif t} &= \tri*{w(t), - \grad F(w(t))}  \\ 
&\leq - (F(w(t)) - F^*). && \text{(since \(F\) is convex)} 
\end{align*}

Integrating on both the sides for \(t\) from \(0\) to \(T\) implies that: 
\begin{align*}
\Phi(w(T)) - \Phi(w(0)) \leq - \int_{t=0}^\infty (F(w(t)) - F^*) \dif t \leq - T  (F(w(T)) - F^*), 
\end{align*}
where the inequality in the second line holds because the function value is non-increasing along any gradient flow path. Rearranging the terms and ignoring negative terms, implies the following rate of convergence for gradient flow:  
\begin{align*}
F(w(T)) - F^* \leq \frac{\Phi(w_0)}{T} \leq \frac{\nrm{w_0}^2}{2T}. 
\end{align*}
	
Next, we argue that gradient descent algorithm given by the recursive process \(w_{k+1} \leftarrow w_{k} - \eta \grad F(w_{k})\) fails to find a \(1/10\) suboptimal solution when  starting from the initial point \(w_0 = \prn*{\frac{1}{d^{2/3}}, \dots, \frac{1}{d^{2/3}}, 1}\). We consider two cases of step size \(\eta\) below: 
\begin{enumerate} 
\item Case 1: \(\eta \leq \frac{3}{d^{1/3}}\). Note that any \(w\) for which \(F(w) \leq 1/10\) must satisfy that \(\abs{V} \leq 1\). However, recall that at initialization, \(v = 1\). Furthermore, $\frac{\partial F(w)}{\partial v} = v$ whenever \(v \in \brk{1/2, 1}\) and thus gradient descent needs to take at least \(\floor*{2 d^{1/3}/3}\) many steps to ensure that \(v \leq 1/2\). 
\item  Case 2: \(\eta > \frac{3}{d^{1/3}}\). We argue that  gradient descent diverges to infinity in this case. In particular, after \(k\) iterations of GD, the iterate \(w_k = (u_k , v_k)\) satisfies
\begin{align*}
u_k[i] = \frac{(1 - \eta d^{1/3})^k}{d^{2/3}}  \numberthis \label{eq:GD_trajectory_lower_bound} 
\end{align*} 
We prove the above via induction. The base case for \(k = 0\) follows by initialization. For the induction step, note that: 
\begin{align*}
u_{k+1}[i] &= u_{k}[i] - \eta \grad F(u_k[i]) \\
&= \frac{(1 - \eta d^{1/3})^k}{d^{2/3}} - \eta \sign{u_k[i]} \cdot \frac{\abs{1 - \eta d^{1/3}}^k}{d^{1/3}} \\ 
&=  \frac{(1 - \eta d^{1/3})^{k+1}}{d^{2/3}}.
\end{align*}
Thus the above implies that after \(T\) iterations, we have that \(F(w) \geq (\eta d^{1/3} - 1)^T\), and thus GD fails to find a \(1/10\) suboptimal solution for any \(T \geq 1\). 
\end{enumerate}  

Combining the two cases above, we get that in order to find a \(1/10\) suboptimal solution, we need \(T \geq \floor*{2 d^{1/3}/3} \geq T_0\) implying the desired lower bound. Since \(T_0\) is arbitrary, the above construction can be extended to hold for any \(T > 0\) (by setting \(d = \infty\)). Thus, there exists a function for which GF succeeds at the rate of \(1/T\) but GD fails to converge. 

We finally conclude by noting that for the function \(F(w)\) and the potential \(\Phi(w) = \nrm{w^2}/2\), we have that for any point \(w\) and \(w'\), 
\begin{align*}
\grad^2 \Phi(w')[\grad F(w), \grad F(w)] &= \nrm{\grad F(w)}^2 \geq \nrm{u}_{3/2} \nrm{u}_1. 
\end{align*} 
On the gradient descent trajectory (given in \pref{eq:GD_trajectory_lower_bound}), the point \(u_k\) satisfies \(\nrm{u_k}_1 = d^{1/3} \nrm{u_k}_{3/2}\) for any \(k \geq 0\). Thus, we have that on the points of GD trajectory, 
\begin{align*}
\grad^2 \Phi(w'_k)[\grad F(w_k), \grad F(w_k)] &= \nrm{\grad F(w_k)}^2 \geq d^{1/3}\nrm{u}^2_{3/2} = 2 d^{1/3} \prn*{F(w_k) - g(v)}. 
\end{align*} 	

Note that the above proof holds for any arbitrarily large \(T_0\). 
\end{proof} 

%% file: files/appx_ub_gd.tex
\subsection{Supporting technical results for proofs of \pref{thm:GD_guarantee}  and \ref{thm:SGD_guarantee}}
Before delving into the proof, we first establish the following structural lemma that relates the function \(F\) and a corresponding potential \(\Phi\). 

\begin{lemma} 
\label{lem:jont_sb_assumption} 
let F(w) be any function that satisfies \pref{ass:F_regular}, and \(\Phi\) be an admissible potential for \(F\) (see \pref{def:Phi}). Then, there exists a monotonically increasing function \(\zeta: \bbR^+ \mapsto \bbR^+\) such that for any \(w\), 
\begin{align*} 
F(w) \leq \zeta \prn*{\Phi(w)}. 
\end{align*} 
\end{lemma} 
\begin{proof}[Proof of \pref{lem:jont_sb_assumption}] \pref{ass:F_regular} implies that for any \(w\), 
\begin{align*} 
\nrm{\grad F(w)}^2 \leq \psi(F(w)) 
\end{align*}
for some monotonically increasing function \(\psi\). Note that without loss of generality, we can assume that \(\psi(F(w)) > 1\) as one can substitute \(\psi(F(w))\) by \(\psi(F(w)) + 1\) while still satisfying the above condition. The above implies that
\begin{align*} 
\bridge{F(w)} \cdot \frac{\nrm{\grad F(w)}^2}{\psi \prn{F(w)}} \leq \bridge{F(w)}. 
\end{align*} 
	
Using the relation in \pref{def:Phi}, we get that the potential \(\Phi\) satisfies 
\begin{align*} 
\frac{\bridge{F(w)}}{\psi \prn{F(w)}} \nrm{\grad F(w)}^2 \leq \bridge{F(w)} \leq \tri{\grad \Phi(w), \grad F(w)}. \numberthis \label{eq:F_Phi_rel0}
\end{align*} 

We first set up additional notation. Define a function \(\sigma(z): \bbR^+ \mapsto \bbR^+\) such that \(\sigma(0) \ldef{} 0\) and for any \(z\), $\sigma'(z) = \bridge{z} / \psi(z)$, 
and note that \(\sigma\) is non-negative and monotonically increasing. We are now ready to delve into the proof. Consider any point $w$. Integrating along the gradient flow path starting from the point \(w\), we get that 
\begin{align*}
\sigma(F(w)) = \sigma(F(w(0))) &= \sigma(F(w(\infty))) - \int_{t=0}^{\infty} \frac{\dif \sigma(F(w(t)))}{\dif t} \dif t \\
&\overeq{\proman{1}} \sigma(F(w(\infty))) + \int_{t=0}^{\infty} \sigma'(F(w(t))) \nrm{\grad F(w(t))}^2 \dif t \\ 
&\overeq{\proman{2}} \int_{t=0}^{\infty} \sigma'(F(w(t))) \nrm{\grad F(w(t))}^2 \dif t \\
&\overeq{\proman{3}} \int_{t=0}^{\infty} \frac{\bridge{F(w(t))}}{\psi(F(w(t)))} \nrm{\grad F(w(t))}^2 \dif t,  \numberthis \label{eq:F_Phi_rel1}
\end{align*} 
where the equality in \(\proman{1}\) follows from Chain rule and because \(\frac{\dif F(w(t))}{\dif t} = - \nrm{\grad F(w(t))}^2\), \(\proman{2}\) holds because of our assumption that \(F(w(\infty)) = 0\) since gradient flow converges to the global minimizer and because \(\sigma(0) = 0\). Finally, \(\proman{3}\) follows from the definition of \(\sigma'(z)\). 

Similarly, integrating along the gradient flow path, we also have that  
\begin{align*}
\Phi(w) = \Phi(w(0)) &= \Phi(w(\infty)) -  \int_{t=0}^{\infty} \frac{\dif \Phi(w(t))}{\dif t} \dif t \\
&\overeq{\proman{1}} \Phi(w(\infty)) +  \int_{t=0}^{\infty} \tri{\grad \Phi(w(t)), \grad F(w(t))}\dif t \\
&\overeq{\proman{2}} \int_{t=0}^{\infty} \tri{\grad \Phi(w(t)), \grad F(w(t))}\dif t, \numberthis \label{eq:F_Phi_rel2}
\end{align*} where in \(\proman{1}\) we used Chain rule and the fact that \(\grad w(t) = - \grad F(w(t))\) and   \(\proman{2}\) holds because \(\Phi(w(\infty)) = \Phi(w^*) = 0\) since \(g(0) = 0\). 

Finally, integrating \pref{eq:F_Phi_rel0} along the gradient flow path, we get the relation
\begin{align*} 
\int_{t=0}^\infty \tri{\grad \Phi(w(t)), \grad F(w(t))} \dif t &\geq \int_{t=0}^\infty \frac{\bridge{F(w)}}{\psi \prn{F(w)}}\nrm{\grad F(w)}^2\dif t. 
\end{align*} 

Plugging the relations \pref{eq:F_Phi_rel1} and \pref{eq:F_Phi_rel2} in the above, we get 
\begin{align*}
\Phi(w) &\geq \sigma(F(w)), 
\end{align*}
which implies that 
\begin{align*} 
F(w) \leq \zeta \prn*{\Phi(w)}, 
\end{align*} where the \(\zeta(z) = \sigma^{-1}(z)\) can be uniquely defined, is positive and monotonically increasing. 
\end{proof}

We next establish the following utility lemma which is an alternative to second-order Taylor's expansion and will be useful in developing convergence bounds for GD and SGD. 
\begin{lemma} 
\label{lem:g_one_sided_smooth} Let \(\Phi\) be any function that satisfies \pref{ass:Phi_regular}. Define the function \(\taylor: \bbR^+ \mapsto \bbR^+\) such that \(\taylor(0) = 0\) and \(\taylor'(z) = 1/\rho(z) \) for any \(z \geq 0\). 
Then, for any \(u \in \bbR^d\), we have 
\begin{align*}
\taylor(\Phi(w + u)) \leq \taylor(\Phi(w)) + \frac{1}{\rho(\Phi(w))}\tri*{\grad \Phi(w), u} + \frac{1}{2}\nrm*{u}^2. 
\end{align*} 

Furthermore, at any point \(w\), \begin{align*}
\nrm*{\grad \Phi(w)} &\leq \rho(\Phi(w)) \sqrt{ 2 \taylor(\Phi(w))}. 
\end{align*}
\end{lemma} 

\begin{proof}[Proof of \pref{lem:g_one_sided_smooth}] Define the function
\begin{align*}
\ls(\alpha) \ldef{} \taylor(\Phi(w + \alpha u)),    \numberthis \label{eq:new_loss}
\end{align*}
and note that 
\begin{align*}
\ls'(\alpha) = \frac{\dif \ls(\alpha)}{\dif \alpha} &=  \taylor'(\Phi(w + \alpha u)) \tri*{\grad \Phi(w + \alpha u), u}, 
\intertext{and}
\ls''(\alpha) = \frac{\dif^2 \ls(\alpha)}{\dif \alpha^2} &=  \taylor''(\Phi(w + \alpha u)) \tri*{u,  \grad \Phi(w + \alpha u)}^2  + \taylor'(\Phi(w + \alpha u)) \tri*{\grad^2 \Phi(w + \alpha u)u, u} \\ 
&\overleq{\proman{1}} \taylor'(\Phi(w + \alpha u)) \tri*{\grad^2 \Phi(w + \alpha u)u, u} \\
&\overleq{\proman{2}} \taylor'(\Phi(w + \alpha u)) \nrm{\grad^2 \Phi(w + \alpha u)} \nrm{u}^2 \\
&\overleq{\proman{3}} \taylor'(\Phi(w + \alpha u)) \rho( \Phi(w + \alpha u)) \nrm{u}^2 \\
&\overleq{\proman{3}} \nrm{u}^2, 
\end{align*} 
where \(\proman{1}\) holds because \(\theta''(z) = \frac{- \rho'(z)}{\rho'(z)^2} \leq 0\) as \(\rho'(z) \geq 0\) since \(\rho\) is a monotonically increasing function, \(\proman{2}\) follows from H\"older's inequality, \(\proman{3}\) is due to \pref{ass:Phi_regular} and finally \(\proman{4}\) is from the definition of the function \(\theta\). 

Using Taylor expansion of \(\ls(1)\) at the point \(\alpha = 0\), we get that 
\begin{align*}
\ls(1) &\leq \ls(0) + \ls'(0) + \frac{1}{2} \ls''(\alpha'), 
\end{align*}
where \(\alpha' \in [0, 1]\). Plugging in the values of \(\ls(0)\), \(\ls(1)\), \(\ls'(0)\) and \(\ls''(\alpha')\) from the above, we get 
\begin{align*}
 \taylor(\Phi(w + u)) &\leq  \taylor(\Phi(w)) + \taylor'(\Phi(w)) \tri{\grad \Phi(w), u} + \frac{1}{2} \nrm{u}^2 \\
 &=  \taylor(\Phi(w)) + \frac{1}{\rho(\Phi(w))} \tri{\grad \Phi(w), u} + \frac{1}{2} \nrm{u}^2,  \numberthis \label{eq:g_one_sided_smooth_5}
\end{align*} where the last line follows by using that fact that \(\taylor'(z) = {1}/{\rho(z)}\). This proves the first relation. 

We next prove the bound on \(\nrm{\grad \Phi(w)}\). Starting from \pref{eq:g_one_sided_smooth_5}, we have that for any \(u \in \bbR^d\), 
\begin{align*}
\taylor(\Phi(w + u)) \leq \taylor(\Phi(w)) + \frac{1}{\rho(\Phi(w))}\tri*{\grad \Phi(w), u} + \frac{1}{2}\nrm*{u}^2. 
\end{align*}
Plugging in \(u = - \frac{\grad \Phi(w)}{\rho(\Phi(w))}\), we get 
\begin{align*}
\taylor(\Phi(w - \frac{\grad \Phi(w)}{\rho(w)})) \leq \taylor(\Phi(w)) - \frac{1}{2\rho(\Phi(w))^2}\nrm*{\grad \Phi(w)}^2. 
\end{align*}

Rearranging the terms, we get
\begin{align*}
\nrm*{\grad \Phi(w)}^2 &\leq 2 \rho(\Phi(w))^2 \prn[\Big]{\taylor(\Phi(w))  - \taylor(\Phi(w - \frac{\grad \Phi(w)}{\rho(w)}))} \\ 
&\leq 2\rho(\Phi(w))^2 \taylor(\Phi(w)),  
\end{align*} where the inequality in the second line holds because \(\taylor(z) > 0\). This proves the second relation. 
\end{proof}

\subsection{Proof of \pref{thm:GD_guarantee}}
We are now ready to prove the convergence guarantee for GD. We first state the full version of  \pref{thm:GD_guarantee} that shows all the problem dependent constants hidden in the main body. While the following bound for GD looks complex at the first sight, this is the price we pay for the generality of our framework. Various invocations of this result are presented in \pref{sec:applications}. 
\begin{theorem*}[\pref{thm:GD_guarantee} restated with problem dependent constants] 
Let \(\Phi_g\) be an admissible potential w.r.t.~$F$. Assume that \(F\) satisfies \pref{ass:F_regular} with the bound given by the function \(\psi\), and \(\Phi_g\) satisfies \pref{ass:Phi_regular} with the bound given by the function \(\rho\). Then, for any initial point \(w_0\), 
\begin{itemize}[leftmargin=5mm] 
\item For any \(T \geq 1\) and \(\eta > 0\), the point \(\wh w_T\) returned by GD algorithm has the convergence guarantee
 \begin{align}
\bridge{F(\wh w_T)} &\leq \frac{2 \taylor(\Phi_g(w_0)) \rho(\Phi_g(w_0))}{\eta T} + 2 \eta \rho(\Phi_g(w_0)) \psi(\zeta(\Phi_g(w_0))), \label{eq:GD_guarantee1_appx}
\end{align} 
Setting \(\eta = \sqrt{\frac{\theta(\Phi_g(w_0))}{\psi(\zeta(\Phi_g(w_0)))} \cdot \frac{1}{T}} \) in the above implies the rate 

\begin{align*}
\bridge{F(\wh w_T)} &\leq 4 \rho(\Phi_g(w_0)) \sqrt{\theta(\Phi_g(w_0)) \psi(\zeta(\Phi_g(w_0)))} \cdot \frac{1}{\sqrt{T}}. \numberthis \label{eq:GD_guarantee2_appx}
\end{align*} 

\item Furthermore, if the function \(\tfrac{\psi}{g}\) is monotonically increasing, then for any \(T \geq 1\) and \(\eta \leq \tfrac{\bridge{\zeta(\Phi_g(w_0))} }{\psi(\zeta(\Phi_g(w_0))) \cdot \rho(\Phi_g(w_0))}\), the point \(\wh w_T\) has the convergence guarantee
\begin{align} 
\bridge{F(\wh w_T)} &\leq \frac{2 \taylor(\Phi_g(w_0)) \rho(\Phi_g(w_0))}{\eta T}.  \numberthis \label{eq:GD_guarantee3_appx}
\end{align} 
Setting  \(\eta = \tfrac{\bridge{\zeta(\Phi_g(w_0))} }{\psi(\zeta(\Phi_g(w_0))) \cdot \rho(\Phi_g(w_0))}\) in the above implies the rate 
\begin{align*}
\bridge{F(\wh w_T)} &\leq\frac{2 \taylor(\Phi_g(w_0)) \psi(\zeta(\Phi_g(w_0))) \rho^2(\Phi_g(w_0))}{\bridge{\zeta(\Phi_g(w_0))}} \cdot \frac{1}{T}. \numberthis \label{eq:GD_guarantee4_appx}
\end{align*}
\item Finally, if \(\Phi_g = F\), then the above bounds hold with all 
occurrence of the term \(\zeta(\Phi_g(w_0))\) replaced with \(F(w_0)\). 
\end{itemize} 

In the above, the function \(\theta(z) \ldef{}  \int_{y=0}^{z} \frac{1}{\rho(y)} \dif y\) and the function \(\zeta\) is defined such that  \(\zeta^{-1}(z) = \int_{y=0}^{z} \frac{\bridge{y}}{\psi(y)} \dif y\). 
\end{theorem*} 

\begin{proof}[Proof of  \pref{thm:GD_guarantee}] For the ease of notation, we remove the subscript \(g\) from the potential \(\Phi_g\) throughout the proof. 
 Fix any \(T > 0\) and let \(\crl{w_t}_{t =  0}^T\) be the sequence of iterates generated by GD on \(F(w)\) when starting from the point \(w_0\) at \(t = 0\). First note that for any \(t \geq 0\), invoking \pref{lem:g_one_sided_smooth}  with \(w = w_t\) and \(u = - \eta \grad F(w_t)\), and using  \pref{def:Phi}, we get 
\begin{align*} 
\taylor(\Phi(w_{t +1})) &\leq \taylor(\Phi(w_t)) - \frac{\eta}{\rho(\Phi(w_t))} \bridge{F(w_t)} + \frac{\eta^2}{2}  \nrm{\grad F(w_t)}^2 \\
&\leq \taylor(\Phi(w_t)) - \frac{\eta}{\rho(\Phi(w_t))} \bridge{F(w_t)} + \frac{\eta^2}{2}  \psi(F(w_t)), \numberthis \label{eq:GD_proof_1}
\end{align*} where \(\theta\) is a monotonically increasing function and the second last line follows from \pref{ass:F_regular}. 

We now proceed with the proof of convergence for GD. Assume that for every \(t \leq T\) 
\begin{align*}
	\bridge{F(w_t)} \geq \eta \rho(\Phi(w_0)) \psi(\zeta(\Phi(w_0))). \numberthis \label{eq:GDsb_1} 
\end{align*}
If the case above condition is violated, we immediately have that 
\begin{align*}
\min_{t \leq T} g(F(w_t)) \leq \eta \rho(\Phi(w_0)) \psi(\zeta(\Phi(w_0))). \numberthis \label{eq:GDsb_5}
\end{align*}
Thus, moving forward we assume that \pref{eq:GDsb_1} holds. Fix any \(t \leq T\). Starting from \pref{eq:GD_proof_1}, we get 
\begin{align*} 
\taylor(\Phi(w_{t +1})) &\leq \taylor(\Phi(w_t))  - \frac{\eta}{\rho(\Phi(w_t))} \bridge{F(w_t)} + \frac{\eta^2}{2}  \psi(F(w_t)) \\
&\leq \taylor(\Phi(w_t)) - \frac{\eta}{\rho(\Phi(w_t))} \bridge{F(w_t)} + \frac{\eta^2}{2}  \psi(\zeta(\Phi(w_t))), \numberthis \label{eq:GDsb_6} 
\end{align*} where the last inequality is due to \pref{lem:jont_sb_assumption} and because \(\Psi\) is a monotonically increasing function. Before we delve into the proof of convergence of GD, we will first establish a useful property that \(\Phi(w_t) \leq \Phi(w_0)\) for all \(t \leq T\). We prove this via induction. For the base case $(t = 0)$, starting from \pref{eq:GDsb_6}, we have 
\begin{align*}
\taylor(\Phi(w_{1})) &\leq \taylor(\Phi(w_0)) - \frac{\eta}{\rho(\Phi(w_0))}  \bridge{F(w_0)} + \frac{\eta^2}{2} \psi(\zeta(\Phi(w_0))) \\ 
&\leq \taylor(\Phi(w_0)) - \frac{\eta}{ 2\rho(\Phi(w_0))} \bridge{F(w_0)} \\
&\leq \taylor(\Phi(w_0)) , 
\end{align*} where the inequality in the second line above holds due to \pref{eq:GDsb_1}. Since \(\theta\) is a monotonically increasing function, the above implies that \(\Phi(w_{1}) \leq \Phi(w_0)\). We next prove the induction step. Assume that  \(\Phi(w_{\tau}) \leq \Phi(w_{0})\) for any \(\tau \leq t\). Again, using \pref{eq:GDsb_6}, we have 
\begin{align*} 
\taylor(\Phi(w_{t+1})) &\leq \taylor(\Phi(w_t)) - \frac{\eta}{\rho(\Phi(w_t))} \bridge{F(w_t)} + \frac{\eta^2}{2}  \psi(\zeta(\Phi(w_t))) \\  
&\overleq{\proman{1}} \taylor(\Phi(w_t)) - \frac{\eta}{\rho(\Phi(w_0))} \bridge{F(w_t)} + \frac{\eta^2}{2}  \psi(\zeta(\Phi(w_0))) \\ 
&\overleq{\proman{2}} \taylor(\Phi(w_t)) - \frac{\eta}{2 \rho(\Phi(w_0))}  \bridge{F(w_t)} \numberthis \label{eq:GDsb_7} \\
&\leq \taylor(\Phi(w_t)), 
\end{align*} where \(\proman{1}\) holds because \(\Phi(w_{t}) \leq \Phi(w_{0})\) via the induction hypothesis and because \(\rho\), \(\zeta\) and \(\psi\) are monotonically increasing and non-negative functions and \(F(w_t) \geq 0\). \(\proman{2}\) is due to the relation in \pref{eq:GDsb_1}. Since \(\theta\) is monotonic, this implies that \(\Phi(w_{t+1}) \leq \Phi(w_{t})\),  completing the induction step and proving that \(\Phi(w_{t}) \leq \Phi(w_{0})\) for all \(t \leq T\). 

Since \(\Phi(w_t) \leq \Phi(w)\) for all \(t \leq T\), starting from \pref{eq:GDsb_6} and replicating the steps till \pref{eq:GDsb_7}, we get  that for any \(t \leq T\), 
\begin{align*} 
\taylor(\Phi(w_{t+1})) &\leq \taylor(\Phi(w_t)) - \frac{\eta}{2 \rho(\Phi(w_0))}  \bridge{F(w_t)}.
\end{align*}
Telescoping the above for \(t\) from \(0\) to \(T-1\) and rearranging the terms, we get that 
\begin{align*} 
\frac{\eta}{2T \rho(\Phi(w_0))} \sum_{t=1}^T \bridge{F(w_t)} &\leq \frac{\taylor(\Phi(w_0)) - \taylor(\Phi(w_{T+1}))}{T}. 
\end{align*}

Ignoring negative terms on the right hand side, we get 
\begin{align*} 
\frac{1}{T} \sum_{t=1}^T \bridge{F(w_t)} &\leq \frac{2 \taylor(\Phi(w_0)) \rho(\Phi(w_0))}{\eta T}, 
\intertext{and thus} 
\min_{t \leq T} \bridge{F(w_t)} &\leq \frac{2 \taylor(\Phi(w_0)) \rho(\Phi(w_0))}{\eta T}. \numberthis \label{eq:GDsb_3} 
\end{align*}

The above analysis shows that at least one of the bound in \pref{eq:GDsb_5} or \pref{eq:GDsb_3} holds. Thus, taking both of them together, we get that 
\begin{align*}
\min_{t \leq T} \bridge{F(w_t)} &\leq \frac{2 \taylor(\Phi(w_0)) \rho(\Phi(w_0))}{\eta T} + \eta \rho(\Phi(w_0)) \psi(\zeta(\Phi(w_0))). 
\end{align*}

\textbf{Improved bound when \(\tfrac{\psi(z)}{\bridge{z}}\) is a monotonically increasing function of \(z\).} In this case, \pref{eq:GD_proof_1}  implies that for any \(t \geq 0\), 
\begin{align*} 
\taylor(\Phi(w_{t +1})) &\leq \taylor(\Phi(w_t)) - \eta \bridge{F(w_t)} \prn*{\frac{1}{\rho(\Phi(w_t))} -  \frac{\eta}{2}  \cdot \frac{\psi(F(w_t))}{\bridge{F(w_t)}}} \\
&\leq  \taylor(\Phi(w_t)) - \eta \bridge{F(w_t)} \prn*{\frac{1}{\rho(\Phi(w_t))} -  \frac{\eta}{2}  \cdot \frac{\psi(\zeta(\Phi(w_t)))}{\bridge{\zeta(\Phi(w_t)})}} \numberthis \label{eq:GD_proof_2}, 
\end{align*} where the last inequality follows from the fact that \(\psi(z)/\bridge{z}\) is an increasing function of \(z\) and from \pref{lem:jont_sb_assumption}. In the following, we will provide a convergence guarantee for GD whenever 
\begin{align*}
\eta \leq \frac{\bridge{\zeta(\Phi(w_0))} }{\psi(\zeta(\Phi(w_0))) \cdot \rho(\Phi(w_0))}. \numberthis \label{eq:GD_proof_1}
\end{align*} We first show that for such an \(\eta\), the iterates produced by GD satisfy \(\Phi(w_t) \leq \Phi(w_0)\) for all \(t \leq T\). The proof follows by induction. For the base case $(t = 0)$, starting from relation \pref{eq:GD_proof_2}, we have 
\begin{align*} 
\taylor(\Phi(w_{1})) &\leq \taylor(\Phi(w_0))  - \eta \bridge{F(w_0)} \prn*{\frac{1}{\rho(\Phi(w_0))} -  \frac{\eta}{2}   \cdot \frac{\psi(\zeta(\Phi(w_0)))}{\bridge{\zeta(\Phi(w_0))}}} \\  
&\leq \taylor(\Phi(w_0)) - \frac{\eta}{ 2\rho(\Phi(w_0))}  \bridge{F(w_0)} \\
&\leq  \taylor(\Phi(w_0)) ,  \numberthis \label{eq:GD_proof_2}
\end{align*} where the inequality in the second line follows by plugging the bound on \(\eta\) from  \pref{eq:GD_proof_1}, and the last inequality holds since \(\bridge{F(w_0)} \geq 0\). Since \(\taylor\) is a monotonically increasing function, the above implies that \(\Phi(w_{1}) \leq \Phi(w_0)\) thus proving the base case. 

We next prove the induction step. Assume that \(\Phi(w_{\tau}) \leq \Phi(w_{0})\) for any \(\tau \leq t\). Again, starting from  relation \pref{eq:GD_proof_2}, we have  
\begin{align*} 
\taylor(\Phi(w_{t+1})) &\leq \taylor(\Phi(w_t)) - \eta \bridge{F(w_t)} \prn*{\frac{1}{\rho(\Phi(w_t))} -  \frac{\eta}{2}  \cdot \frac{\psi(F(w_t))}{\bridge{F(w_t)}}} \\ 
&\leq \taylor(\Phi(w_t)) - \eta \bridge{F(w_t)} \prn*{\frac{1}{\rho(\Phi(w_0))} -  \frac{\eta}{2}  \cdot \frac{\psi(\zeta(\Phi(w_0)))}{\bridge{\zeta(\Phi((w_0))}}} \\ 
&\leq \taylor(\Phi(w_t)) - \frac{\eta}{2 \rho(\Phi(w_0))}  \bridge{F(w_t)} \numberthis \label{eq:GDsb_7} \\
&\leq \taylor(\Phi(w_t)), 
\end{align*} where the second line holds because \(F(w_{t}) \leq \zeta(\Phi(w_t))  \leq \zeta(\Phi(w_{0}))\) and \(\psi(z)/\bridge{z}\) is a monotonically increasing function of \(z\), the third line holds by plugging the bound on $\eta$ from \pref{eq:GD_proof_1}, and the last inequality holds since \(F(w_0) \geq 0\). Since \(\taylor\) is monotonically increasing, this implies that \(\taylor(w_{t+1}) \leq \taylor(w_{t})\),  completing the induction step and proving that \(\Phi(w_{t}) \leq \Phi(w_{0})\) for all \(t \leq T\). 
 
We are now ready to complete the proof of convergence of GD. Since \(\Phi(w_t) \leq \Phi(w)\) for all \(t \leq T\), starting from \pref{eq:GD_proof_2} and replicating the steps till \pref{eq:GDsb_7}, we get that for any \(t \leq T\), 
\begin{align*}
\taylor(\Phi(w_{t+1})) &\leq \taylor(\Phi(w_t)) - \frac{\eta}{2 \rho(\Phi(w_0))}  \bridge{F(w_t)}.  \numberthis \label{eq:GDsb_11} 
\end{align*}
Telescoping the above for \(t\) from \(0\) to \(T\) and rearranging the terms, we get that 
\begin{align*} 
\frac{\eta}{2T \rho(\Phi(w_0))} \sum_{t=1}^T \bridge{F(w_t)} &\leq \frac{\taylor(\Phi(w_0)) - \taylor(\Phi(w_{T+1}))}{T}. 
\end{align*} 

Ignoring negative items on the right hand side, we get 
\begin{align*} 
\frac{1}{T} \sum_{t=1}^T \bridge{F(w_t)} &\leq \frac{2 \taylor(\Phi(w_0)) \rho(\Phi(w_0))}{\eta T}, 
\intertext{and thus} 
\min_{t \leq T} \bridge{F(w_t)} &\leq \frac{2 \taylor(\Phi(w_0)) \rho(\Phi(w_0))}{\eta T}. 
\end{align*} 

\textbf{Improved analysis when \(\Phi_g = F\).} The proof follows identically, with the only major change being that \pref{lem:jont_sb_assumption} now holds with the function \(\zeta(z) = z\) since \(F(w) = \Phi_g(w)\). 
\end{proof}

%% file: files/appx_ub_sgd_new.tex
\subsection{Proof of \pref{thm:SGD_guarantee}} 
We first note the following high probability and in-expectation bounds on the norm of the stochastic gradient estimate. 
\begin{lemma} 
\label{lem:SGD_bound_hp}	 
Let \(\crl*{w_t}_{t \leq T}\) be the sequence of iterates generated by SGD algorithm on \(F\) using stochastic estimates based on \(\crl{z_t}_{t \leq T}\). Then, with probability at least \(1 - \delta\), for any time \(t \leq T\), 
\begin{align*}
\nrm*{\grad f(w;z)}^2 &\leq  \Lambda(F(w)) \log\prn*{T/\delta} 
\end{align*}
and for any \(w > 0\), 
\begin{align*}
\En \brk*{\nrm{\grad f(w; z)}^2} &\leq \Lambda(F(w)), 
\end{align*}
where the function $\Lambda(z) \ldef{}  2 \psi(z) + 2\chi(z)$, and the functions \(\psi\) and \(\chi\) given in \pref{ass:F_regular} and \ref{ass:gradient_noise} respectively. 
\end{lemma} 
\begin{proof}[Proof of \pref{lem:SGD_bound_hp}] Note that for any \(0 \leq t \leq T\), with probability at least \(1 - \frac{\delta}{T}\), 
\begin{align*}
\nrm*{\grad f(w; z)  - \grad F(w)}^2 &\leq \chi(F(w)) \cdot \log\prn*{\frac{T}{\delta}}, 
\intertext{which implies that}
\nrm*{\grad f(w;z)}^2 &\leq 2 \nrm*{\grad F(w)}^2  + 2 \nrm*{\grad f(w; z)  - \grad F(w)}^2 \\ 
&\leq 2 \psi(F(w)) + 2 \nrm*{\grad f(w; z)  - \grad F(w)}^2 \\ 
&\leq \prn*{2\psi(F(w))  +  2 \chi(F(w))} \cdot \log\prn*{\frac{T}{\delta}} \\  
&= \Lambda(F(w)) \log\prn*{\frac{T}{\delta}}, 
\end{align*} where the inequality in the second to last line follows from \pref{ass:gradient_noise} and the last line is from the definition of \(\Lambda\). The desired bounds follows with probability at least \(1 - \delta\) by taking a union bound w.r.t.~\(t\).

For the in-expectation bound, since for any random variable \(X\), \(\En \brk*{X} = \int_{t = 0}^\infty \Pr(X \geq t) \dif t\), we have  
\begin{align*} 
\frac{\En \brk*{\nrm{\grad f(w; z) - \grad F(w)}^2}}{\chi(F(w))} &= \int_{t=0}^\infty  \Pr \prn*{\frac{\En \brk*{\nrm{\grad f(w; z) - \grad F(w)}^2}}{\chi(F(w))}  \geq t} \dif t \\ 
&\leq \int_{t=0}^\infty  e^{- t} \dif t = 1.  \numberthis \label{eq:basic_expect1}
\end{align*}
Thus, 
\begin{align*}
\En \brk*{\nrm{\grad f(w; z)}^2} &\leq 2\nrm*{\grad F(w)}^2 + 2 \En \brk*{\nrm{\grad f(w; z) - \grad F(w)}^2} \\ &\leq  2\nrm*{\grad F(w)}^2 + 2 \chi(F(w)) \\
&\leq 2 \psi(F(w)) + 2\chi(F(w)) \rdef{} \Lambda(F(w)), 
\end{align*} where the inequality in the second line above follows \pref{eq:basic_expect1} and the last line is due to \pref{ass:gradient_noise}. 
\end{proof} 

We are now ready to prove the convergence guarantee for SGD. We first state the full version of \pref{thm:SGD_guarantee} that shows all the problem dependent constants hidden in the main body, but keeps \(\kappa\) as a free variable. Then, we provide an easier to understand result in  \pref{rem:SGD_bound}  by setting  \(\kappa\) appropriately. Various invocations of this result are presented in \pref{sec:applications}. 
 
\begin{theorem*}[\pref{thm:SGD_guarantee} restated with problem dependent constants] 
Let \(\Phi_g\) be an admissible potential w.r.t.~$F$. Assume that \(F\) satisfies  \pref{ass:F_regular} with the bound given by the function \(\psi\), \(\Phi_g\) satisfies \pref{ass:Phi_regular} with the bound given by the function \(\rho\), and the stochastic gradient estimates \(\grad f(w; z)\) satisfy \pref{ass:gradient_noise} with the bound given by the function \(\chi\). Then, for any \(T \geq 1\), \(\kappa > 1\), initial point \(w_0\),  setting $$\eta \leq  \frac{M - \theta(\Phi_g(w_0))}{20\log^2(20 T) \sqrt{MBT}},$$ we get that with probability at least \(0.7\), the point \(\wh w_T\) returned by SGD algorithm satisfies 
\begin{align*} 
\bridge{F(\wh w_T)} &\leq \kappa \rho(\Phi(w_0)) \prn*{\frac{100 M}{\eta T} + 50 \eta B \log^2\prn*{20T}}.  
\end{align*} 
where the function \(\theta(z) \ldef{} \int_{y=0}^z \frac{1}{\rho(y)} \dif y\), the function \(\zeta\) is defined such that  \(\zeta^{-1}(z) = \int_{y=0}^{z} \frac{\bridge{y}}{\psi(y)} \dif y\) and the function \(\Lambda(z) \ldef{} 2 \psi(z) + 2 \chi(z)\). Furthermore, the constant  \(B = \Lambda(\zeta(\rho^{-1}(\kappa \rho(\Phi_g(w_0)))))\) and \(\)
\(M = \taylor(\rho^{-1}(\kappa \rho(\Phi_g(w_0))))\). 
\end{theorem*} 
\begin{remark}  
\label{rem:SGD_bound} Fix any initial point \(w_0\) and let \(\bw\) be any point such that \(\Phi_g(\bw) > \Phi_g(w_0)\). Then, setting \(\kappa = \tfrac{\rho(\Phi_g(\bw))}{\rho(\Phi_g(w_0))}\) in \pref{thm:SGD_guarantee} (above) implies that \(B = \Lambda(\zeta(\Phi_g(\bar{w})))\) and \(M = \theta(\Phi_g(\bar{w}))\). Thus, for any \(T \geq 1\), setting $$\eta \leq  \frac{\theta(\Phi_g(\bw)) - \theta(\Phi_g(w_0))}{20\log^2(20 T) \sqrt{\Lambda(\zeta(\Phi_g(\bw))) \theta(\Phi_g(\bw)) \cdot T}},$$ we get that with probability at least \(0.7\), the point \(\wh w_T\) returned by SGD algorithm satisfies   
\begin{align*} 
\bridge{F(\wh w_T)} &\leq \widetilde{O} \prn*{\rho(\Phi_g(\bw)) \cdot \frac{\theta(\Phi_g(\bw))}{\theta(\Phi_g(\bw)) - \theta(\Phi_g(w_0))} \cdot \sqrt{\Lambda(\zeta(\Phi_g(\bw)) \theta(\Phi_g(\bw)))} \cdot \frac{1}{\sqrt{T}}}.  
\end{align*} 
\end{remark}

\begin{proof}[Proof of \pref{thm:SGD_guarantee}]
Let \(\crl*{w_t}_{t \leq T}\) be the sequence of iterates generated by SGD algorithm in the first \(T\) times steps using the random samples \(\crl*{z_t}_{t \leq T}\) sampled i.i.d.~ from an unknown distribution. Let \(\cF_t\) be the natural filtration at time \(t\) such that \(\crl{w_{j}, z_{j}}_{j \leq t}\) are \(\cF_t\)-measurable, and let \(\bar{\eta} = \frac{M - \theta(\Phi(w_0))}{20\log^2(20 T) \sqrt{MBT}}\). 

\vspace{2mm} 

\textbf{Part 1: Setup.} For any \(0 \leq t \leq T\), an application of \pref{lem:g_one_sided_smooth} with \(w = w_t\) and \(u = - \eta \grad f(w_t; z_t)\) implies that   
\begin{align*} 
\taylor(\Phi(w_{t +1})) &= \taylor(\Phi( w_t - \eta \grad f(w_t; z_t))) \\
&\leq \taylor(\Phi(w_t)) - \frac{\eta}{\rho(\Phi(w_t))} \tri*{\grad f(w_t; z_t),  \grad \Phi(w_t)} + \frac{\eta^2}{2}  \nrm{\grad f(w_t; z_t)}^2. \numberthis \label{eq:SGD_proof_1} 
\end{align*} 
 
Taking expectation on both the sides with respect to \(z_t\), we get 
\begin{align*} 
\En_t \brk*{\taylor(\Phi(w_{t +1}))} &\leq \taylor(\Phi(w_t)) - \frac{\eta}{\rho(\Phi(w_t))} \En_t \brk*{ \tri*{\grad f(w_t; z_t),  \grad \Phi(w_t)}} + \frac{\eta^2}{2} \En_t \brk*{ \nrm{\grad f(w_t; z_t)}^2 } \\
&\leq \taylor(\Phi(w_t)) - \frac{\eta}{\rho(\Phi(w_t))} g(F(w_t)) + \frac{\eta^2}{2} \En_t \brk*{ \nrm{\grad f(w_t; z_t)}^2 } \\ 
&\leq \taylor(\Phi(w_t)) - \frac{\eta}{\rho(\Phi(w_t))} g(F(w_t)) + \frac{\eta^2}{2} \Lambda(F(w_t)),  \numberthis \label{eq:SGD_proof_14}
\end{align*} where the inequality in the second line holds because \( \En \brk*{ \tri*{\grad f(w_t; z_t),  \grad \Phi(w_t)}} = \tri{\grad F(w_t), \grad \Phi(w_t)} \geq g(F(w_t))\) since \(w_t\) is independent of \(z_t\), and the last line follows from \pref{lem:SGD_bound_hp}. Rearranging the terms and summing for \(t\) from \(0\) to \(T - 1\), we get that  
\begin{align*} 
\sum_{t=0}^{T-1} \frac{\eta}{\rho(\Phi(w_t))} g(F(w_t)) &\leq  \sum_{t=0}^{T-1} \prn*{\taylor(\Phi(w_t)) - \En_t \brk*{\taylor(\Phi(w_{t +1}))} + \frac{\eta^2}{2} \Lambda(F(w_t))}.  \numberthis \label{eq:SGD_proof_24} 
\end{align*} 

Our focus in Part-2 below will be to control the term on the left hand size above. 

\vspace{2mm} 

\textbf{Part 2: Lower bound on \(\mb{\rho(\Phi(w_t)})\)}.  We first set up additional notation and derive some supporting results. Consider the stochastic process \(\crl{Y_t}_{t \leq T}\) defined as 
\begin{align*} 
Y_t = \begin{cases} 
\taylor(\Phi(w_{t})) + \sum_{j=0}^{t-1} \prn*{\frac{\eta}{\rho(\Phi(w_j))} \bridge{F(w_j)} - \frac{\eta^2 \Lambda(F(w_j))}{2}}  & \text{if \(t \leq \tau\)} \\
	Y_{\tau} & \text{if \(t > \tau\)}
\end{cases}, \numberthis \label{eq:SGD_sb_process_definition} 
\end{align*}
where \(\tau\) is defined as the first time smaller than or equal to \(T\) at which \(\rho(\Phi(w_t)) > \kappa \rho(\Phi(w_0))\) i.e., 
\begin{align*} 
\tau \ldef{} \inf \crl*{ t \leq T \mid \rho(\Phi(w_t)) > \kappa \rho(\Phi(w_0))}, \numberthis \label{eq:stopping_time_SGD_sb} 
\end{align*}
where \(\kappa > 1\) and will be set later. If there is no such \(\tau\) for which  \pref{eq:stopping_time_SGD_sb} holds, we set \(\tau = T\). Essentially, \(\crl{Y_t}_{t \leq T}\) is a stochastic process where \(Y_t\) depends on the random variable  \(w_t\), and is stopped as soon as \(\rho(\Phi(w_t)) > \kappa \rho(\Phi(w_0))\). To keep the current proof concise, we show in \pref{lem:Y_t_supmartingale} (below) that the process \(\crl*{Y_t}_{t \geq 0}\) is a super-martingale with respect to the filtration \(\cF_t\), and that with probability at least \(0.95\), for all \(t \leq T\), 
\begin{align}
Y_t - Y_0 \leq \sqrt{\frac{1}{2} \sum_{j=0}^{t-1} \prn*{  5 \eta \sqrt{M} \cdot \nrm*{\grad f(w_{j}; z_j)} + 4 \eta^2 \nrm*{\grad f(w_{j}; z_j)}^2}^2 \log(20T)}. \label{eq:SGD_proof_15}   \tag{\(\mathscr{E}_1\)}
\end{align}
where  \(M = \taylor(\rho^{-1}(\kappa \rho(\Phi(w_0)))) \) .
We additionally also note that from \pref{lem:SGD_bound_hp}, with probability at least \(0.95\), for all \(t \leq T\), 
\begin{align}
\nrm*{\grad f(w_t;z_t)}^2 &\leq  \Lambda(F(w_t)) \log\prn*{20T} \label{eq:SGD_proof_16}   \tag{\(\mathscr{E}_2\)}
\end{align} 

Taking a union bound over the events \(\mathscr{E}_1\) and \(\mathscr{E}_2\) above, we get that for any \(t \leq T\), 
\begin{align}
Y_t - Y_0 &\leq \sqrt{\frac{1}{2} \sum_{j=0}^{t-1} \prn*{  5 \eta \sqrt{M \cdot \Lambda(F(w_j))} + 4 \eta^2  \Lambda(F(w_j))}^2 \log^3(20T)}. \label{eq:SGD_proof_17}   \tag{\(\mathscr{E}_3\)}
\end{align}
In the following, we show that under the event \(\mathscr{E}_3\), the condition in \pref{eq:stopping_time_SGD_sb} never occurs. Suppose the contrary is true and that \pref{eq:stopping_time_SGD_sb} occurs for some \(\tau \leq T\). Then, we have that 
\begin{align*} 
Y_{\tau} - Y_0 &\leq \sqrt{\frac{1}{2}  \sum_{j=0}^{\tau -1} \prn*{  5 \eta \sqrt{M \cdot \Lambda(F(w_j))} + 4 \eta^2  \Lambda(F(w_j))}\log^3(20T)}\\
&\leq \sqrt{\frac{\tau}{2} \prn*{  5 \eta \sqrt{M B} + 4 \eta^2 B}^2 \log^3(20T)} \\
&\leq 9 \eta \sqrt{M B}  \log^2 (20T) \cdot \sqrt{\tau}  \numberthis \label{eq:SGD_proof_18}
\end{align*} where the last line holds because \(\eta \leq \bar{\eta} \leq \sqrt{M/B}\) and in the second to last line, we used the fact that 
\begin{align*}
\Lambda(F(w_j)) \overleq{\proman{1}} \Lambda(\zeta(\Phi(w_j))) \overleq{\proman{2}} \Lambda(\zeta(\rho^{-1}(\kappa \rho(\Phi(w_0))))) \rdef{} B, \numberthis \label{eq:SGD_proof_19}
\end{align*} where \(\proman{1}\) holds due to \pref{lem:jont_sb_assumption} and \(\proman{2}\) follows from the fact that \(\rho(\Phi(w_j)) \leq \kappa \rho(\Phi(w_0))\) for all \(j < \tau\). However, from the definition of \(Y_t\), we also have that 
\begin{align*}
Y_\tau - Y_0 &= \taylor(\Phi(w_\tau)) - \taylor(\Phi(w_0)) + \sum_{j=0}^{\tau -1} \prn*{\frac{\eta}{\rho(\Phi(w_j))} \bridge{F(w_j)} - \frac{\eta^2 \Lambda(F(w_j))}{2}} \\
&\geq \taylor(\Phi(w_\tau)) - \taylor(\Phi(w_0)) - \sum_{j=0}^{\tau -1} \frac{\eta^2 \Lambda(F(w_j))}{2} \\
&\overgeq{\proman{1}} M - \taylor(\Phi(w_0)) - \sum_{j=0}^{\tau -1} \frac{\eta^2 \Lambda(F(w_j))}{2} \\
&\overgeq{\proman{2}} M - \taylor(\Phi(w_0)) - \frac{\eta^2 \tau B}{2}  \\
&\geq \frac{M - \taylor(\Phi(w_0))}{2} \numberthis \label{eq:SGD_proof_20} 
\end{align*} where in \(\proman{1}\), we used the fact that \(\Phi(w_\tau) > \rho^{-1}(\kappa \rho(\Phi(w_0))) = M\),  \(\proman{2}\) follows by noting the bound in \pref{eq:SGD_proof_19} for any \(j < \tau\). The last line follows from the fact that \(\eta \leq \bar{\eta} \leq \sqrt{(M - \theta(\Phi(w_0))/B T}\). However, note that this leads to a contradiction as both \pref{eq:SGD_proof_18} and \pref{eq:SGD_proof_20} can not be simultaneously true when when \(\eta \leq \bar{\eta} = \frac{M - \theta(\Phi(w_0))}{20\log^2(20 T) \sqrt{MBT}}\). Thus, we must have that with probability at least \(0.9\), for any \(t \leq T\), 
\begin{align*}
\rho(\Phi(w_t)) \leq  \kappa \rho(\Phi(w_0)) \numberthis \label{eq:SGD_proof_sb5}
\end{align*}

In the following, we condition on the fact that \pref{eq:SGD_proof_sb5} holds. 

\vspace{2mm}

\textbf{Part 3: Convergence guarantee.} The following proof conditions on the events \(\mathscr{E}_1\), \(\mathscr{E}_2\) , \(\mathscr{E}_3\). First note that, telescoping \pref{eq:SGD_proof_1} from \(t = 0\) to \(T-1\) and ignoring negative terms in the right hand side, we get that 
\begin{align*} 
\eta \sum_{t=0}^{T-1} \frac{\tri*{\grad f(w_t; z_t),  \grad \Phi(w_t)}}{\rho(\Phi(w_t))} &\leq \taylor(\Phi(w_0)) + \frac{\eta^2}{2} \sum_{t=0}^{T-1}   \nrm{\grad f(w_t; z_t)}^2.  \numberthis \label{eq:SGD_proof_27} 
\end{align*} 
The left hand side above can be controlled using Azuma-Hoeffding's inequality (\pref{lem:Azuma2}), which implies that with probability at least \(0.95\), 
\begin{align*}
 \sum_{t=0}^{T-1} \frac{\tri*{\grad f(w_t; z_t),  \grad \Phi(w_t)}}{\rho(\Phi(w_t))} &\geq  \sum_{t=0}^{T-1} \En \brk*{\frac{\tri*{\grad f(w_t; z_t),  \grad \Phi(w_t)}}{\rho(\Phi(w_t))}} - 2 \max_{t < T}  \frac{\tri*{\grad f(w_t; z_t),  \grad \Phi(w_t)}}{\rho(\Phi(w_t))} \sqrt{T \log(20)} \\
 &\overeq{\proman{1}} \sum_{t=0}^{T-1} \En \brk*{\frac{\tri*{\grad F(w_t),  \grad \Phi(w_t)}}{\rho(\Phi(w_t))}} - 2 \max_{t < T}  \frac{\nrm*{\grad f(w_t; z_t)}\nrm*{\grad \Phi(w_t)}}{\rho(\Phi(w_t))} \sqrt{T \log(20)} \\
 &\overgeq{\proman{2}} \sum_{t=0}^{T-1} \En \brk*{\frac{\bridge{F(w_t)}}{\rho(\Phi(w_t))}} - 2 \max_{t < T}  \sqrt{2 \taylor(\Phi(w_t))} \nrm*{\grad f(w_t; z_t)} \sqrt{T \log(20)}
\end{align*}
 where \(\proman{1}\) above holds due to linearity of expectation w.r.t.~\(z_t\) and the inner product, and using Cauchy-Schwarz inequality. The inequality in \(\proman{2}\) holds because of the relation \pref{eq:linearity_property} and \pref{lem:g_one_sided_smooth}. 
 
Plugging the above bound in \pref{eq:SGD_proof_27} and rearranging the terms, we get 
\begin{align*}
\eta \sum_{t=0}^{T-1} \En \brk*{\frac{\bridge{F(w_t)}}{\rho(\Phi(w_t))}} &\leq \taylor(\Phi(w_0)) + \frac{\eta^2}{2} \sum_{t=0}^{T-1}   \nrm{\grad f(w_t; z_t)}^2 + 2 \eta \max_{t < T}  \sqrt{2 \taylor(\Phi(w_t))} \nrm*{\grad f(w_t; z_t)} \sqrt{T \log(20)}. 
\end{align*}

An application of Markov's inequality in the above implies that with probability at least \(0.9\), 
\begin{align*}
\eta \sum_{t=0}^{T-1} \frac{\bridge{F(w_t)}}{\rho(\Phi(w_t))} &\leq 10 \eta \sum_{t=0}^{T-1} \En \brk*{\frac{\bridge{F(w_t)}}{\rho(\Phi(w_t))}}  \\
&\leq 10 \taylor(\Phi(w_0)) + 5\eta^2 \sum_{t=0}^{T-1}   \nrm{\grad f(w_t; z_t)}^2 + 20 \eta \max_{t < T}  \sqrt{2 \taylor(\Phi(w_t))} \nrm*{\grad f(w_t; z_t)} \sqrt{T \log(20)}.  \tag{\(\mathscr{E}_4\)}
\end{align*}
Conditioning on the event \(\mathscr{E}_2\) and plugging in the corresponding bound on \(\nrm*{\grad f(w_t;z_t)}^2\), and dividing both the sides by \(\eta\), we get that 
\begin{align*}
\sum_{t=0}^{T-1} \frac{\bridge{F(w_t)}}{\rho(\Phi(w_t))} &\leq \frac{10 \taylor(\Phi(w_0))}{\eta} + 5\eta \sum_{t=0}^{T-1} \Lambda(F(w_t)) \log\prn*{20T} + 20 \max_{t < T}  \sqrt{2 \taylor(\Phi(w_t)) \Lambda(F(w_t))T  \log^2\prn*{20T}} \\
&\leq  \frac{10 \taylor(\Phi(w_0))}{\eta} + 5\eta \sum_{t=0}^{T-1} \Lambda(\zeta(\Phi(w_t))) \log\prn*{20T} \\
& \qquad \qquad \qquad \qquad + 20 \max_{t < T}  \sqrt{2 \taylor(\Phi(w_t)) \Lambda(\zeta(\Phi((w_t)))T  \log^2\prn*{20T}} \\
&\leq \frac{10 M}{\eta} + 5\eta B T \log\prn*{20T} + 20  \sqrt{2 M B T  \log^2\prn*{20T}}, 
\end{align*}
where the second line above holds because of \pref{lem:jont_sb_assumption} and because \(\Lambda\) is monotonically increasing. The inequality in the last line follows from plugging in the bound \pref{eq:SGD_proof_sb5} which implies that \( \Lambda(\zeta(\Phi(w_t))) \leq \Lambda(\zeta(\rho^{-1}(\kappa \rho(\Phi(w_0))))) = B\), and \(\taylor(\Phi(w_t)) \leq \taylor(\rho^{-1}(\kappa \rho(\Phi(w_0))) ) = M\) since both \(\Lambda\) and \(\zeta\) are monotonically increasing functions. Using  \pref{eq:SGD_proof_sb5} in the LHS above, rearranging the terms and dividing both the sides by \(T\), we get that
\begin{align*}
\frac{1}{T}  \sum_{t=0}^{T-1} \bridge{F(w_t)} &\leq \kappa \rho(\Phi(w_0)) \prn*{\frac{10 M}{\eta T} + 5\eta B \log\prn*{20T} + 20 \sqrt{\frac{2 M B  \log^2\prn*{20T}}{T}}} \\
&\leq \kappa \rho(\Phi(w_0)) \prn*{\frac{100 M}{\eta T} + 50 \eta B \log^2\prn*{20T}}, 
\end{align*} where the last line is by applying AM-GM inequality on the last term. 

Accounting for the union bounds for events \(\mathscr{E}_1\), \(\mathscr{E}_2\) , \(\mathscr{E}_3\)  and \(\mathscr{E}_4\), we get that the above bound on the rate of convergence of GD holds with probability at least \(0.7\). 
\end{proof}

The following technical result is used in the proof of \pref{thm:SGD_guarantee}. 
\begin{lemma}
\label{lem:Y_t_supmartingale} 
Suppose the premise of \pref{thm:SGD_guarantee} holds, and let \(\crl*{w_t}_{t \leq T}\) be the sequence of iterates generated by SGD algorithm on \(F\) using stochastic estimates based on \(\crl{z_t}_{t \leq T}\). Let the process \(\crl{Y_t}_{t \geq 0}\) be  defined as 
\begin{align*} 
Y_t = \begin{cases} 
\taylor(\Phi(w_{t})) + \sum_{j=0}^{t-1} \prn*{\frac{\eta}{\rho(\Phi(w_j))} \bridge{F(w_j)} - \frac{\eta^2 \Lambda(F(w_j))}{2}}  & \text{if \(t \leq \tau\)}\\
	Y_{\tau} & \text{if \(t > \tau\)}
\end{cases}, \numberthis \label{eq:SGD_sb_process_definition} 
\end{align*}
where \(\tau = \min \crl{T,  \inf \crl*{ t \mid \rho(\Phi(w_t)) > \kappa \rho(\Phi(w_0))}}\) and $\Lambda(z) =  2 \psi(F(w)) + 2\chi(F(w))$ where the function \(\psi\) and \(\chi\) given in \pref{ass:F_regular} and \ref{ass:gradient_noise} respectively. Then, \(\crl{Y_t}_{t \geq 0}\) is a super-martingale. Furthermore, with probability at-least \(0.95\), for all \(t \leq T\),  
\begin{align*}
Y_t - Y_0 \leq \sqrt{\frac{1}{2} \sum_{j=0}^{t-1} \prn*{  5 \eta \sqrt{M} \cdot \nrm*{\grad f(w_{j}; z_j)} + 4 \eta^2 \nrm*{\grad f(w_{j}; z_j)}^2} \log(20T)}, 
\end{align*}
where \(M = \taylor(\rho^{-1}(\kappa \rho(\Phi(w_0))) ) \).  
\end{lemma} 
\begin{proof}[Proof of \pref{lem:Y_t_supmartingale}] Let \(\cF_t\) be the natural filtration at time \(t\) such that \(\crl{w_{j}, z_{j}}_{j \leq t}\) are \(\cF_t\)-measurable.
For any \(t \geq 0\), repeating the steps till \pref{eq:SGD_proof_14} in the proof of \pref{thm:SGD_guarantee} above we get that 
\begin{align*} 
\En_t \brk*{\taylor(\Phi(w_{t +1}))} & \leq \taylor(\Phi(w_t)) - \frac{\eta}{\rho(\Phi(w_t))} \bridge{F(w_t)} + \frac{\eta^2 \Lambda(F(w_t))}{2},   \numberthis \label{eq:SGD_martingale1} 
\end{align*}
where \(\En_t\) denotes expectation w.r.t.~the random variable \(z_t\), and conditioning on \(\cF_{t-1}\). 
We first show that the process \(\crl{Y_t}_{t \geq 0}\) is a super-martingale. Note that for any time \(t \leq \tau\), 
\begin{align*} 
\En_t \brk*{Y_{t + 1}} &= \En_t \brk*{ \taylor(\Phi(w_{t +1}))} + \sum_{j=0}^{t} \prn*{\frac{\eta}{\rho(\Phi(w_j))} \bridge{F(w_j)} - \frac{\eta^2 \Lambda(F(w_j))}{2}}  \\ 
&\leq \taylor(\Phi(w_t)) + \sum_{j=0}^{t-1} \prn*{\frac{\eta}{\rho(\Phi(w_j))} \bridge{F(w_j)} - \frac{\eta^2 \Lambda(F(w_j))}{2}} = Y_t, 
\end{align*}  
where the inequality in the second line above follows from \pref{eq:SGD_martingale1}. When \(t > \tau\), by definition we have that \(\En_t \brk*{Y_{t + 1}} = Y_t\). Hence, the process \(\crl{Y_t}_{t \geq 0}\) is a super-martingale. 

\paragraph{Bound on the difference sequence.} There are two cases, either (a) \(t > \tau\), or (b) \(t \leq \tau\). In the first case, \(\abs*{Y_{t + 1} - Y_t} = 0\). In the following, we  provide a bound on the difference sequence for \(t \leq \tau\). First note that 
\begin{align*}
Y_{t+1} - Y_t &= \taylor(\Phi(w_{t + 1})) - \taylor(\Phi(w_{t})) + \underbrace{\frac{\eta}{\rho(\Phi(w_t))} \bridge{F(w_t)} - \frac{\eta^2 \Lambda(F(w_t))}{2}}_{ \ldef{} C_t}. \numberthis \label{eq:SGD_martingale2} 
\end{align*} 

Note that the term \(C_t\) is \(\cF_t\)-predictable. Thus, we just need to find \(\cF_{t}\)-measurable processes \(A'_t\) and \(B'_t\) such that 
\begin{align*}
A'_t \leq \taylor(\Phi(w_{t + 1})) - \taylor(\Phi(w_{t})) \leq B'_t. 
\end{align*} Recall that an application of \pref{lem:g_one_sided_smooth} with \(w = w_t\) and \(u = - \eta \grad f(w_t; z_t)\) implies that 
\begin{align*}
\taylor(\Phi(w_{t + 1})) - \taylor(\Phi(w_{t})) &\leq - \frac{\eta}{\rho(\Phi(w_t))} \tri*{\grad f(w_t; z_t),  \grad \Phi(w_t)} + \frac{\eta^2}{2}  \nrm{\grad f(w_t; z_t)}^2\\
&\overleq{\proman{1}}  \frac{\eta}{\rho(\Phi(w_t))} \nrm*{\grad f(w_t; z_t)} \nrm*{\grad \Phi(w_t)} + \frac{\eta^2}{2}  \nrm{\grad f(w_t; z_t)}^2  \\ 
&\overleq{\proman{2}} \underbrace{\eta \sqrt{2 \taylor(\Phi(w_t))} \nrm*{\grad f(w_t; z_t)} + \frac{\eta^2}{2}  \nrm{\grad f(w_t; z_t)}^2}_{\rdef{} B'_t}\numberthis \label{eq:SGD_martingale3}, 
\end{align*} where \(\proman{1}\) above follows from Cauchy-Schwarz inequality, and \(\proman{2}\) holds due to \pref{lem:g_one_sided_smooth}. Note that  \(B'_t\) defined to be the terms on the RHS above is \(\cF_t\)-measurable. 

We next consider the lower bound on \(\taylor(\Phi(w_{t + 1})) - \taylor(\Phi(w_{t}))\). Plugging in \(w = w_{t+1}\) and  \(u = \eta \grad f(w_t; z_t)\) in \pref{lem:g_one_sided_smooth}, we get that 
\begin{align*} 
\taylor(\Phi(w_t)) &= \taylor(\Phi(w_{t+1} + \eta \grad f(w_t; z_t))) \\ &\leq \taylor(\Phi(w_{t+1})) + \frac{\eta}{\rho(\phi(w_{t+1}))} \tri*{\grad \Phi(w_{t+1}), \grad f(w_{t}; z_t)} + \frac{\eta^2}{2} \nrm*{\grad f(w_{t}; z_t)}^2, 
\intertext{rearranging the terms gives us} 
 \taylor(\Phi(w_{t+1}))  - \taylor(\Phi(w_{t})) &\geq - \frac{\eta}{\rho(\phi(w_{t+1}))} \tri*{\grad \Phi(w_{t+1}), \grad f(w_{t}; z_t)} -  \frac{\eta^2}{2} \nrm*{\grad f(w_{t}; z_t)}^2  \\ 
 &\overgeq{\proman{1}} - \frac{\eta}{\rho(\phi(w_{t+1}))} \nrm*{\grad \Phi(w_{t+1})} \nrm*{\grad f(w_{t}; z_t)} -  \frac{\eta^2}{2} \nrm*{\grad f(w_{t}; z_t)}^2 \\ 
 &\overgeq{\proman{2}} - \eta \sqrt{2 \taylor(\Phi(w_{t+1}))} \cdot \nrm*{\grad f(w_{t}; z_t)}  -  \frac{\eta^2}{2} \nrm*{\grad f(w_{t}; z_t)}^2 \\ 
  &= - \eta \sqrt{2 \taylor(\Phi(w_{t+1}) - \taylor(\Phi(w_t)) + \taylor(\Phi(w_t)))} \cdot \nrm*{\grad f(w_{t}; z_t)}  -  \frac{\eta^2}{2} \nrm*{\grad f(w_{t}; z_t)}^2 \\ 
  &\overgeq{\proman{3}} - \eta \sqrt{2 \abs{\taylor(\Phi(w_{t+1}) - \taylor(\Phi(w_t)))}} \cdot \nrm*{\grad f(w_{t}; z_t)}  - \eta \sqrt{2 \taylor(\Phi(w_t)))} \cdot \nrm*{\grad f(w_{t}; z_t)}  \\
  & \qquad \qquad \qquad \qquad \qquad -  \frac{\eta^2}{2} \nrm*{\grad f(w_{t}; z_t)}^2 \\ 
    &\overgeq{\proman{4}} - \frac{\abs*{\taylor(\Phi(w_{t+1}) - \taylor(\Phi(w_t))}}{2} - \eta \sqrt{2 \taylor(\Phi(w_t)))} \cdot \nrm*{\grad f(w_{t}; z_t)} -  \frac{3 \eta^2}{2} \nrm*{\grad f(w_{t}; z_t)}^2 
   \end{align*} where \(\proman{1}\) follows from Cauchy-Schwarz inequality, and \(\proman{2}\) holds due to \pref{lem:g_one_sided_smooth}. Inequality \(\proman{3}\) follows from subadditivity of sq-root. Finally, \(\proman{4}\) follows from an application of AM-GM inequality. Rearranging the terms, we get 
\begin{align*} 
 \taylor(\Phi(w_{t+1}))  - \taylor(\Phi(w_{t})) &\geq \underbrace{- 2 \eta \sqrt{2 \taylor(\Phi(w_t)))} \cdot \nrm*{\grad f(w_{t}; z_t)} - 3 \eta^2 \nrm*{\grad f(w_{t}; z_t)}^2}_{\rdef{} A'_t}.  \numberthis \label{eq:SGD_martingale4}      
\end{align*}   
Note that  \(A'_t\), defined to be the terms on the RHS above, is \(\cF_t\)-measurable. 

The bounds in \pref{eq:SGD_martingale3} and \pref{eq:SGD_martingale4} imply that the processes \(\crl{A'_t}_{t \geq 0}\) and \(\crl{B'_t}_{t \geq 0}\) are \(\cF_t\)-measurable and satisfy 
\begin{align*} 
A'_t  \leq	 \taylor(\Phi(w_{t+1}))  - \taylor(\Phi(w_{t})) \leq B'_t 
\end{align*} for any \(t \geq 0\). Plugging this in \pref{eq:SGD_martingale2}, we get 
\begin{align*} 
A_t \ldef{} A'_t + C_t \leq Y_{t+1} - Y_{t} \leq B'_t + C_t \rdef{} B_t. 
\end{align*}

Clearly both \(A_t\) and \(B_t\) are \(\cF_t\)-measurable and satisfy 
\begin{align*}
\abs*{B_t  - A_t} &\leq 5 \eta \sqrt{\taylor(\Phi(w_t))} \cdot \nrm*{\grad f(w_{t}; z_t)} + 4 \eta^2 \nrm*{\grad f(w_{t}; z_t)}^2 \\ 
&\leq 5 \eta \sqrt{M} \cdot \nrm*{\grad f(w_{t}; z_t)} + 4 \eta^2 \nrm*{\grad f(w_{t}; z_t)}^2,
\end{align*}  where the last line follow from the fact that \(t \leq \tau\) and thus \(\Phi(w_t) \leq \rho^{-1}(\kappa \rho(\Phi(w_0)))\) which implies that  \(\taylor(\Phi(w_t)) \leq \taylor(\rho^{-1}(\kappa \rho(\Phi(w_0))) ) \rdef{} M\) since \(\theta\) is a monotonically increasing function. 

\textbf{High probability bound.} An application of Azuma's inequality (\pref{lem:Azuma2}) implies that for any \(t \geq 0\), with probability at least \(1 - 1/20T\), 
\begin{align*}
Y_t - Y_0 \leq \sqrt{\frac{1}{2} \sum_{j=0}^{t-1} \prn*{  5 \eta \sqrt{M} \cdot \nrm*{\grad f(w_{j}; z_j)} + 4 \eta^2 \nrm*{\grad f(w_{j}; z_j)}^2}^2 \log(20T)}. 
\end{align*}
The desired statement follows by taking a union bound in the above for \(t\) from \(0\) to \(T-1\). 
\end{proof}

%% file: files/KL_functions.tex
\subsection{Kurdyka-\L ojasiewicz (K\L) functions}

We recall the following definition of K\L~ functions. Recall that we assumed that \(F\) is non-negative with \(\min_{w} F(w) = 0\).
\begin{definition}[K\L~ functions] 
\label{def:KL_def_appx} 
The objective \(F\) satisfies Kurdyka-\L ojasiewicz (K\L) property with exponent \(\theta \in (0, 1)\) and coefficient \(\alpha \in \bbR^+\), if for any point \(w\), 
\begin{align*} 
\nrm{\grad F(w)}^2 \geq \alpha F(w)^{1 + \theta}.  
\end{align*}
\end{definition} 

In the following, we will provide convergence guarantees for K\L~ functions that are \(H\)-smooth (c.f. \pref{ass:F_sb_KL}). 

\subsubsection{Rate of convergence for gradient flow}
The next lemma provides an admissible rate of convergence for K\L~function. 
\begin{lemma}
\label{lem:KL_rate}
 Suppose \(F\) is K\L~ with exponent \(\gamma\)  $\in (0, 1/2)$ (\pref{def:KL_def_appx}). Then, for any initial point point \(w(0) = w_0\), the point \(w(t)\) on its gradient flow path satisfies  
\begin{align*}
F(w(t)) \leq R(w_0, t) \ldef{} \frac{F(w_0)}{\prn*{1 + \alpha \theta F(w_0)^\theta \cdot t}^{1/\theta}}. 
\end{align*} 
Furthermore, \(R\) is an admissible rate of convergence w.r.t.~\(F\). 
\end{lemma}
\begin{proof}[Proof of \pref{lem:KL_rate}] Note that 
\begin{align*} 
\frac{\dif F(w(t))}{\dif t} & = \tri{\grad F(w(t)), \frac{\dif w(t)}{\dif t}} \\ 
&= - \nrm{\grad F(w(t))}^2 \\
&\leq - \alpha F(w(t))^{1 + \theta}.
\end{align*}
Rearranging the terms above implies the differential equation
\begin{align*}
\frac{\dif F(w(t))}{F(w(t))^{1 + \theta}} \leq - \alpha \dif t, 
\end{align*}
solving which for \(\theta \in (0, 1)\) gives the bound 
\begin{align*}
F(w(t)) \leq \frac{F(w(0))}{\prn*{1 + \alpha \theta t \cdot F(w(0))^\theta}^{1/\theta}}
\end{align*}

The desired statement following by plugging in \(w(0) = w_0\) and defining 
\begin{align*}
R(w, t) \ldef \frac{F(w)}{\prn*{1 + \alpha \theta t \cdot F(w)^\theta}^{1/\theta}}
\end{align*}
We next show that the above function \(R\) is an admissible rate of convergence w.r.t.~\(F\). Recall that a sufficient conditions for admissibility of \(R\) is that for any point \(w\), 
\begin{align*} 
\int_{t=0}^\infty \prn[\Big]{\frac{\partial R(w, t)}{\partial t} + \left\langle \grad R(w, t), \nabla F(w) \right\rangle } \dif t \geq 0. \numberthis \label{eq:kl_1} 
\end{align*} 

Note that 
\begin{align*} 
\int_{t=0}^\infty \frac{\partial R(w, t)}{\partial t} \dif t &= - F(w), 
\intertext{and}
\int_{t=0}^\infty \left\langle \grad R(w, t), \nabla F(w) \right\rangle &= \nrm{\grad F(w)}^2 \int_{t=0}^\infty \frac{1}{(1 + \alpha \theta t \cdot F(w)^\theta)^{\frac{1}{\theta}}} \dif t \\
&\qquad \qquad \qquad \qquad - \alpha \theta F(w)^\theta \nrm{\grad F(w)}^2\int_{t=0}^\infty \frac{t}{(1 + \alpha \theta t \cdot F(w)^\theta)^{1 + \frac{1}{\theta}}} \dif t\\
& =\frac{\nrm{\grad F(w)}^2}{(1 - \theta) \alpha F(w)^\theta} - \frac{\theta \nrm{\grad F(w)}^2}{(1 - \theta) \alpha F(w)^\theta} \\
&= \frac{\nrm{\grad F(w)}^2}{\alpha F(w)^\theta} 
\end{align*}

Combining the two bounds together implies that a sufficient condition for \(R\) to be admissible is that 
\begin{align*}
\frac{\nrm{\grad F(w)}^2}{\alpha F(w)^\theta} \geq F(w). 
\end{align*} 
Since \(F\) is K\L~with exponent \(\theta\) and coefficient \(\alpha\), the above holds true for any \(w\), thus implying that \(R\) is an admissible rate function. 
\end{proof}

\subsubsection{Potential function and self-bounding regularity conditions}
Consider the function 
\begin{align*}
R(w, t) \ldef \frac{F(w)}{\prn*{1 + \alpha \theta t \cdot F(w)^\theta}^{1/\theta}}
\end{align*}

\pref{lem:KL_rate} implies that \(R\) is an admissible rate of convergence for any K\L~ objective function \(F\). Thus, using \pref{thm:gf_to_potential} with \(g(z) = \alpha z^{1 + \theta}\), we get that the function \(\Phi_g\)  constructed in the following is an admissible potential function for \(F\), 
\begin{align*} 
\Phi_g(w) &= \int_{t=0}^\infty g(R(w, t)) \dif t \\
&= \alpha \int_{t=0}^\infty \frac{F(w)^{1 + \theta}}{\prn*{1 + \alpha \theta t \cdot F(w)^\theta}^{\frac{1}{\theta} + 1}} \dif t \\
&= F(w). \numberthis \label{eq:kl_potential} 
\end{align*} 

Note that we already assumed self-bounding regularity conditions on \(F\) in \pref{ass:F_sb_KL}. In the following, we derive self-bounding regularity conditions for the potential \(\Phi_g\). 
\begin{lemma}
\label{lem:phi_sb_reg} 
Suppose that \(F\) satisfies \pref{ass:F_sb_KL}. Then, for any point \(w\), the potential function \(\Phi_g\) in \pref{eq:kl_potential} satisfies that 
 \begin{align*}
\nrm{\grad^2 \Phi_g(w)} &\leq \psi(\Phi_g(w)), 
\end{align*} 
where \(\psi\) is the positive, monotonically increasing function given in \pref{ass:F_sb_KL}. 
\end{lemma}
\begin{proof} From the definition of \(\Phi_g\), we have that \(\nrm{\grad^2 \Phi_g(w)} = \nrm{\grad^2 \Phi_g(w)} \). The desired self-bounding regularity conditions on \(\Phi_g\) thus follows from \pref{ass:F_sb_KL}. 
\end{proof}

We next prove \pref{prop:KL_complete}. 
\begin{proof}[Proof of \pref{prop:KL_complete}] The proof of \((b) \Rightarrow (a)\) follows from \pref{lem:KL_rate}. For the proof of \((a) \Rightarrow (b)\), we note that plugging the given rate in \pref{thm:gf_to_potential}, we get that the function \(\Phi_g(w) = F(w)\) is an admissible potential function w.r.t.~ \(F(w)\) with \(g(z) = \alpha z^{1 + \theta}\). Thus, from \pref{eq:linearity_property}, we get that 
\begin{align*}
\nrm{\grad F(w)}^2 = \tri{\grad \Phi_g(w), F(w)} \geq g(F(w)) = \alpha F(w)^{1 + \theta}, 
\end{align*}
which implies the desired P\L~property for \(F\). 
\end{proof}
\subsubsection{GD for K\L~functions} \label{app:GD_kl}
In the following, we provide the respective problem dependent quantities and instantiate \pref{thm:GD_guarantee}  to provide a convergence bound for GD for K\L~ functions. 
\begin{itemize}
\item We set 
\begin{align*}
g(z) = \alpha z^{1 + \theta}. 
\end{align*}
\item \pref{ass:F_regular} follows from  \pref{lem:smoothness_inequality} and \pref{ass:F_sb_KL} which implies that
\begin{align*}
\psi(z) = 4 H z. 
\end{align*}
\item \pref{ass:Phi_regular} follows from  \pref{ass:F_sb_KL} which implies that
\begin{align*}
\rho(z) = H.  
\end{align*}
\item The function \(\theta\) is given by
\begin{align*}
\theta(z) =  \int_{y=0}^{z} \frac{1}{\rho(y)} \dif y = \frac{z}{H}. 
\end{align*}
\item The function \(\zeta\) is defined such that  
\begin{align*}
\zeta^{-1}(z) = \int_{y=0}^{z} \frac{\bridge{y}}{\psi(y)} \dif y = \int_{y=0}^z \frac{\alpha y^{\theta}}{4 H} \dif y = \frac{\alpha}{4 H (1 + \theta)} z^{1 + \theta}, 
\end{align*} 
which implies that 
\begin{align*}
\zeta(z) = \prn*{\frac{4 H (1 + \theta) z}{\alpha}}^{{1}/{1 + \theta}}. 
\end{align*} 
\end{itemize} 

  Plugging the above problem-dependent constants in \pref{thm:GD_guarantee} (under the case that \(\Phi_g = F\))  implies that setting 
\begin{align*}
\eta &= \sqrt{\frac{\theta(F(w_0))}{\psi(F(w_0))} \cdot \frac{1}{T}} \leq \frac{1}{2 H \sqrt{T}}, 
\end{align*} 
GD has the rate 

\begin{align*}	
\bridge{F(\wh w_T)} &\leq 4 \rho(\Phi_g(w_0)) \sqrt{\theta(\Phi_g(w_0)) \psi(\zeta(\Phi_g(w_0)))} \cdot \frac{1}{\sqrt{T}}. \\
&\leq \frac{8 H F(w)}{\sqrt{T}}, 
\end{align*} Plugging $g(z) = \alpha z^{1 + \theta}$ in the above implies that 
\begin{align*}
F(\wh w_T) \leq \prn*{\frac{4H F(w)}{\alpha}}^{\frac{1}{1 + \theta}} \cdot \frac{1}{T^{1/(2 + 2 \theta)}}. 
\end{align*}

Clearly, the function \(\tfrac{\psi(z)}{g(z)} = \tfrac{4H}{\alpha z^\theta}\) is not a monotonically increasing function of \(z\), and thus the improved analysis for GD does not extend to this case. 

\subsubsection{SGD for K\L~functions} \label{app:SGD_kl}
Suppose \pref{ass:gradient_noise} is satisfies with \(\chi(z) = \sigma^2\). In addition to the problem dependent quantities in \pref{app:GD_kl}, we define the function $\Lambda$ used in \pref{thm:SGD_guarantee} as 
\begin{align*}
\Lambda(z) = 4 H z + 2 \sigma^2. 
\end{align*} 
Fix any \(\bw\) such that \(2F(w_0) \leq F(\bw) \leq 4F(w_0)\) and define \(B = 16  \prn*{H^{2 + \theta} \cdot \frac{F(\bw)}{\alpha}}^{{1}/{1 + \theta}} + 2\sigma^2\). Following \pref{thm:SGD_guarantee} (in particular the bound in \pref{rem:SGD_bound}), we note that for any 
\begin{align*}
\eta \leq \frac{1}{20  \log^2(20T) } \cdot \frac{F(\bw) - F(w_0)}{ \sqrt{B H  F(\bw) T}}, 
\end{align*}
the point returned by SGD algorithm after \(T\) iterations satisfies with probability at least \(0.7\), 
\begin{align*} 
\bridge{F(\wh w_T)} &\lesssim H \sqrt{BH F(w_0)} \cdot \frac{1}{\sqrt{T}}. 
\end{align*} 
which implies that 
\begin{align*} 
F(\wh w_T) &\lesssim \prn*{\frac{BH^3 F(w_0)}{\alpha^2 T} }^{1/2 + 2 \theta}. 
\end{align*} 

\clearpage

%% file: files/appx_pr.tex
\subsection{Phase retrieval}
For any \(w \in \bbR^d\), the population loss for phase retrieval is given by 
\begin{align*}
F(w) = \En_{a \sim \cN(0, \mathrm{I}_d)} \brk*{\prn*{(a^\top w)^2 - (a^\top w^*)^2}^2}.  \numberthis \label{eq:phase_retrieval_objective} 
\end{align*}

Throughout this section, we will assume that the optimal parameter \(w^*\) satisfies \(\nrm{w^*} = 1\). 
The following technical lemma establishes some useful properties of \(F\). 

\begin{lemma} 
\label{lem:phase_ret_F_dist_relation} 
Suppose \(\nrm{w^*}=1\). Then, the function \(F\) given in \pref{eq:phase_retrieval_objective} satisfies for any \(w \in \bbR^d\), 
\begin{enumerate}[label=\((\alph*)\)] 
\item \(F(w) = w^\top (\mathrm{I} - (w^*)(w^*)^\top) w + \frac{3}{4} \prn*{\nrm{w}^2 - 1}^2\). 
\item \(\tri*{w^*, \grad F(w)} = 3 \prn{\nrm{w}^2 - 1} \tri{w, w^*}\). 
\item \(\nrm{\grad F(w)}^2 = 12 \nrm{w}^2 F(w)  - 8 \prn*{\nrm{w}^2 - \tri{w, w^*}^2}\). 
\item \(F(w) \geq \prn{\nrm{w}^2 - 1}^2\). 
\item if  \(F(w) \leq 1/4\), then \(w\) must satisfy \(\tri*{w, w^*}^2 \geq 1/4\).
\end{enumerate}
\end{lemma}
\begin{proof}[Proof of \pref{lem:phase_ret_F_dist_relation}] We prove each part separately below: 
\begin{enumerate}[label=\((\alph*)\)] 
	\item The proof is straightforward. We refer the reader to Section 2.3 of \citet{candes2015phase} for the proof.   

	\item Note that 
\begin{align*}
	\grad F(w) &= 2 w - 2 \tri{w, w^*} w^* + 3\prn{\nrm{w}^2 - 1} w. 
\end{align*}
Thus, 
\begin{align*}
\tri{w^*, \grad F(w)} &= 2 \tri{w, w^*} - 2 \tri{w, w^*} \nrm{w^*}^2 + 3\prn{\nrm{w}^2 - 1} \tri{w , w^*} \\ 
&= 2 \tri{w, w^*} - 2 \tri{w, w^*} +  3\prn{\nrm{w}^2 - 1} \tri{w , w^*}\\
&=  3\prn{\nrm{w}^2 - 1} \tri{w , w^*}, 
\end{align*} where the second line above holds because $\nrm{w^*}^2 = 1$. 

\item We have
\begin{align*}
\nrm{\grad F(w)}^2 &= \nrm{2 w - 2 \tri{w, w^*} w^* + 3\prn{\nrm{w}^2 - 1} w}^2 \\ 
&= 4 \nrm{w}^2 + 4 \tri{w, w^*}^2 \nrm{w^*}^2 + 9\prn{\nrm{w}^2 - 1}^2 \nrm{w}^2 - 8 \tri{w, w^*}^2 \\
&\qquad \qquad \qquad - 12 \prn{\nrm{w}^2 - 1} \tri{w, w^*}^2  + 12 \prn{\nrm{w}^2 - 1} \nrm{w}^2 \\ 
&= - 12 \nrm{w}^2 \tri{w, w^*}^2  + 12 \nrm{w}^4 +  9\prn{\nrm{w}^2 - 1}^2 \nrm{w}^2 + 8 \tri{w, w^*}^2 - 8 \nrm{w}^2 \\ 
&= 12 \nrm{w}^2 \prn[\big]{ \nrm{w}^2 - \tri{w, w^*}^2 + \frac{3}{4} \prn{\nrm{w}^2 -1}^2} - 8 \prn*{\nrm{w}^2 - \tri{w, w^*}^2} \\ 
&= 12 \nrm{w}^2 F(w)  - 8 \prn*{\nrm{w}^2 - \tri{w, w^*}^2}, 
\end{align*} where the equality in the third line holds because \(\nrm{w^*}^2 = 1\) and the last line follows from the definition of the function \(F(w)\) in part-(a) of this lemma. 

\item An application of Jensen's inequality implies that 
\begin{align*}
F(w) &= \En_{a} \brk*{\prn{(a^\top w)^2 - (a^\top w^*)^2}^2} \\
&\geq \prn*{ \En_{a} \brk{ (a^\top w)^2 - (a^\top w^*)^2}}^2 \\
&= \prn{\nrm{w}^2 - \nrm{w^*}^2}^2,  
\end{align*} where the last line follow from the fact that  for any \(w\), we have \(\En_{a \sim \cN(0, \mathrm{I})} \brk*{(a^\top w)^2} = \nrm*{w}^2\). The desired statement follows since \(\nrm{w^*} = 1\). 

\item An application of \pref{lem:phase_ret_F_dist_relation}-(d) implies that 
\begin{align*}
\prn*{\nrm{w}^2 - 1}^2 \leq F(w) \leq \frac{1}{4}, 
\end{align*}
which implies that \(1/2 \leq \nrm{w}^2 \leq 3/2\). Next, using \pref{lem:phase_ret_F_dist_relation}-(a), we note that 
\begin{align*}
F(w) &= w^\top (\mathrm{I} - (w^*)(w^*)^\top) w + \frac{3}{4} \prn*{\nrm{w}^2 - 1}^2  \geq \nrm{w}^2 - \tri{w, w^*}^2 \geq \frac{1}{2} - \tri{w, w^*}^2, 
\end{align*} where the last line uses the above derived bound on \(\nrm{w}^2\). Rearranging the terms and using the fact that \(F(w) \leq 1/4\) implies that \(\tri{w, w^*}^2 \geq 1/4\). 
\end{enumerate}
\end{proof}

\subsubsection{Rate of convergence for gradient flow} 

The next lemma provides a rate of convergence for the phase retrieval population objective. 
\begin{lemma}
\label{lem:pr_gf_rate}
 Consider the objective function \(F\) given in  \pref{eq:phase_retrieval_objective}. Then, for any initial point point \(w(0) = w_0\), the point \(w(t)\) on its gradient flow path satisfies  
\begin{align*}
F(w(t)) \leq \min \crl{F(w_0), F(w_0) e^{-t + \frac{1}{\tri{w_0, w^*}^2}}}, 
\end{align*}
\end{lemma}
\begin{proof}[Proof of \pref{lem:pr_gf_rate}] Let \(w(t)\) be the point on the GF path with starting point \(w(0) = w_0\). For the ease of notation, define \(\alpha(t) = \tri{w(t), w^*}^2 \) and \(\beta(t) = \nrm{w}^2 - \alpha(t)\). A closer look at the gradient flow dynamics \(w'(t) = - \grad F(w(t))\) reveals that:
\begin{align*}
\alpha'(t) = 6 \prn*{\alpha(t) - \alpha(t)^2 - \alpha(t) \beta(t)}, \\
\beta'(t) = 2 \prn*{\beta(t) - 3 \alpha(t) \beta(t) - 3 \beta(t)^2}.  \numberthis \label{eq:pr_gf1} 
\end{align*}
Define the variable \(\gamma(t) = \alpha(t) / \beta(t)\) and note that 
\begin{align*}
	\gamma'(t) &= \frac{1}{\beta(t)^2} \prn*{\beta(t) \alpha'(t) - \alpha(t) \beta'(t)}  \\ 
	&=  \frac{2}{\beta(t)^2} \prn*{\beta(t) \alpha'(t) - \alpha(t) \beta'(t)} \\
	&\overleq{\proman{1}} \frac{4 \alpha(t)}{\beta(t)} = 4 \gamma(t), 
\end{align*} where \(\proman{1}\) follows from plugging in the relations in \pref{eq:pr_gf1}. Solving the above differential equation implies that \(\gamma(t) = \gamma(0) e^{4t}\), which on plugging in the form of \(\gamma\) implies that 
\begin{align*}
\beta(t) = \alpha(t) \frac{\beta(0)}{\alpha(0)} e^{-4t}.   \numberthis \label{eq:pr_gf2}
\end{align*}
Plugging the above relation in \pref{eq:pr_gf1} gives us the differential equation 
\begin{align*}
\alpha'(t) = 6 \prn*{\alpha(t) - \alpha(t)^2 - 3 \alpha(t)^2 \frac{\beta(0)}{\alpha(0)} e^{-4t}},   \numberthis \label{eq:pr_gf3}
\end{align*}
solving which implies that 
\begin{align*}
\alpha(t) = \frac{\alpha(0) e^{6t}}{1 + 3 \beta(0) (e^{2t} - 1) + \alpha(0)(e^{6t} - 1)}. \numberthis \label{eq:pr_gf4}
\end{align*} Plugging the above form of \(\alpha(t)\) in \pref{eq:pr_gf2} further implies that \begin{align*}
\beta(t) &= \frac{\beta(0) e^{2t}}{1 + 3 \beta(0) (e^{2t} - 1) + \alpha(0)(e^{6t} - 1)}. \numberthis \label{eq:pr_gf5}
\end{align*}

In the rest of the proof, we will show that 
\begin{align*}
F(w(t)) \leq \min \crl{F(w(0)), F(w(0)) e^{-t + \frac{1}{\alpha(0)}}}.  \numberthis \label{eq:pr_gf6}
\end{align*}
For the ease of notation, we will use \(\alpha\) and \(\beta\) to denote \(\alpha(0)\) and \(\beta(0)\) respectively. There are two natural cases for the above, (a) when \(t \leq {1}/{\alpha(0)}\) and (b) when \(t > 1/\alpha(0)\). In the former case, recalling that the function value is non-increasing along any gradient flow path (\pref{lem:gf_decrease}) we get that 
\begin{align*}
F(w(t)) \leq F(w) \leq \min \crl{F(w(0)), F(w(0)) e^{-t + \frac{1}{\alpha(0)}}}. 
\end{align*}
We next show that \pref{eq:pr_gf6} continues to holds when \(t > 1/\alpha(0)\). Note that, from the form of \(F\) in \pref{lem:phase_ret_F_dist_relation}-(a), we have 
\begin{align*}
F(w(t)) &= \nrm{w(t)}^2 - \tri{w(t), w^*}^2 + \frac{3}{4} \prn*{\nrm{w}^2 - 1}^2 \\
&= \beta(t) + \frac{3}{4} \prn*{\alpha(t) + \beta(t) - 1}^2 \\ 
&\overeq{\proman{1}} \frac{\beta e^{2t}}{1 + 3 \beta (e^{2t} - 1) + \alpha(e^{6t} - 1)} + \frac{3}{4} \prn*{\frac{\alpha e^{6t} + \beta e^{2t}}{1 + 3 \beta (e^{2t} - 1) + \alpha(e^{6t} - 1)} - 1}^2 \\
&= \frac{\beta e^{2t}}{1 + 3 \beta (e^{2t} - 1) + \alpha(e^{6t} - 1)} + \frac{3}{4} \prn*{\frac{ - 2 \beta(e^{2t} - 1) - (1 - \alpha - \beta) }{1 + 3 \beta (e^{2t} - 1) + \alpha(e^{6t} - 1)}}^2 \\ 
&\overleq{\proman{2}} \frac{\beta e^{2t}}{1 + 3 \beta (e^{2t} - 1) + \alpha(e^{6t} - 1)} +  6 \beta \cdot \frac{\beta (e^{2t} - 1)^2}{\prn{1 + 3 \beta (e^{2t} - 1) + \alpha(e^{6t} - 1)}^2} \\
& \qquad \qquad \qquad + \frac{3}{2} \frac{ (1 - \alpha - \beta)^2}{\prn{1 + 3 \beta (e^{2t} - 1) + \alpha(e^{6t} - 1)}^2} \numberthis \label{eq:pr_gf7}
\end{align*} where \(\proman{1}\) follows by plugging in the relations \pref{eq:pr_gf4} and \pref{eq:pr_gf5}, and \(\proman{2}\) holds because \((a + b)^2 \leq 2 a^2 + 2 b^2\) for any \(a, b \geq 0\). In the following, we bound the three terms on the right hand side of \pref{eq:pr_gf7} separately for \(t \geq {1}/{\alpha}\). 
\begin{enumerate} 
\item \textit{Term I:} Ignoring the positive term \(3 \beta (e^{2t} - 1)\) in the denominator, we get that
\begin{align*}
\frac{\beta e^{2t}}{1 + 3 \beta (e^{2t} - 1) + \alpha(e^{6t} - 1)} &\leq \frac{\beta e^{2t}}{1 + \alpha(e^{6t} - 1)} \leq \frac{1}{3} \beta e^{-t + 1/\alpha},
\end{align*} 
where the last inequality follows from using \pref{lem:pr_technical_2} (given below). 
\item \textit{Term II:} For the second term, we note that 
\begin{align*}
\frac{\beta (e^{2t} - 1)^2}{\prn{1 + 3 \beta (e^{2t} - 1) + \alpha(e^{6t} - 1)}^2} &\leq \max_{\beta > 0} ~ \frac{\beta (e^{2t} - 1)^2}{\prn{1 + 3 \beta (e^{2t} - 1) + \alpha(e^{6t} - 1)}^2} \\
&\overeq{\proman{1}} (e^{2t} - 1) \frac{\prn*{1 + \alpha(e^{6t} - 1)}}{3} \cdot \frac{1}{4 \prn*{1 + \alpha(e^{6t} - 1)}^2} \\
&= \frac{1}{12} \cdot \frac{(e^{2t} - 1)}{1 + \alpha(e^{6t} - 1)} \\
&\leq \frac{1}{12} \cdot \frac{e^{2t}}{1 + \alpha(e^{6t} - 1)} \\
&\overleq{\proman{2}} \frac{1}{36} \cdot e^{-t + \frac{1}{\alpha}}, 
\end{align*}
where \(\proman{1}\) holds because the term on the right hand side in the equation above is maximized at \(\beta = (1 + \alpha(e^{6t} - 1))/3(e^{2t} - 1)\), and \(\proman{2}\) follow from an application of \pref{lem:pr_technical_2} (given below). 
\item \textit{Term III:} Since the term on the denominator is larger than $1$, we have that 
\begin{align*}
\frac{ (1 - \alpha - \beta)^2}{\prn{1 + 3 \beta (e^{2t} - 1) + \alpha(e^{6t} - 1)}^2} &\leq \frac{ (1 - \alpha - \beta)^2}{\prn{1 + 3 \beta (e^{2t} - 1) + \alpha(e^{6t} - 1)}} \\
&\leq (1 - \alpha - \beta)^2 \cdot \frac{e^{2t}}{\prn{1 + 3 \beta (e^{2t} - 1) + \alpha(e^{6t} - 1)}} \\
&\leq \frac{1}{3} \cdot (1 - \alpha - \beta)^2 \cdot e^{-t + \frac{1}{\alpha}}, 
\end{align*} where the inequality in the second last line holds for any \(t \geq 0\) and the last line is due to \pref{lem:pr_technical_2} (given below). 
\end{enumerate} 
Plugging the above three bounds in \pref{eq:pr_gf7}, we get that for any \(t \geq \tfrac{1}{\alpha}\), 
\begin{align*}
F(w) &\leq \frac{1}{3} \beta e^{-t + 1/\alpha} + \frac{1}{6} \beta e^{-t + \frac{1}{\alpha}} + \frac{1}{2} \cdot (1 - \alpha - \beta)^2 \cdot e^{-t + \frac{1}{\alpha}} \\
&\leq  \prn[\big]{\beta + \frac{3}{4}  (1 - \alpha - \beta)^2} e^{-t + \frac{1}{\alpha}} \\
&= F(w(0)) e^{-t + \frac{1}{\tri{w(0), w^*}^2}}, 
\end{align*} where in the last line we used the form of \(F\) from \pref{lem:phase_ret_F_dist_relation}-(a) and the fact that \(\alpha = \alpha(0) = \tri{w(0), w^*}^2\) and \(\beta = \beta(0) = \nrm{w(0)}^2 - \tri{w(0), w^*}^2\). Finally, using \pref{lem:gf_decrease}, we note that \(F(w(t)) \leq F(w)\). Combining these two bounds gives us the relation in \pref{eq:pr_gf6} for any \(t \geq 1/\alpha(0)\). 
\end{proof}

\begin{lemma} 
\label{lem:pr_technical_2}
For any \(\alpha > 0\) and \(t \geq 1/\alpha\), 
\begin{align*}
\frac{e^{2t}}{1 + \alpha(e^{6t} - 1)} \leq \frac{e^{-t + \frac{1}{\alpha}}}{3}. 
\end{align*}
\end{lemma}
\begin{proof}[Proof of \pref{lem:pr_technical_2}] We consider two cases when \(\alpha \geq 1\) and when \(\alpha < 1\) separately below:
\begin{enumerate}
\item \textit{Case 1: \(\alpha \geq 1\):} Define \(g(t) = {e^{3t}}/\prn{1 + \alpha(e^{6t} - 1)}\) and note that \(g\) is a non-increasing function of \(t\) for \(\alpha \geq 1\). Thus, for any \(t \geq 1/\alpha\), 
\begin{align*}
	g(t) \leq g(\tfrac{1}{\alpha}) = \frac{e^{\frac{3}{\alpha}}}{1 + \alpha(e^{\frac{6}{\alpha}} - 1)} \leq \frac{1}{3} e^{\frac{1}{\alpha}}, 
\end{align*}
where the last inequality holds because the function \(\zeta(z) = e^z / 3 - e^{3z} / (1 + (e^{6z} - 1)/z)\) is non-negative whenever \(z \geq 0\). Multiplying both the sides by \(e^{-t}\) gives the desired relation. 

\item \textit{Case 2: \(\alpha < 1\):}  In this case, ignoring positive terms in the denominator (since \(1 - \alpha > 0\)), we get 
\begin{align*}
\frac{e^{2t}}{1 + \alpha(e^{6t} - 1)} &= \frac{e^{2t}}{1 - \alpha + \alpha e^{6t}} \leq \frac{e^{2t}}{\alpha e^{6t}} = e^{-t} \cdot \frac{e^{-3t}}{\alpha} \leq \frac{1}{3} e^{-t + {1}/{\alpha}}, 
\end{align*} where the second to last inequality follows from the fact that \(e^{-3t}\) is a decreasing function of \(t\) and thus for \(t \geq {1}/{\alpha}\), we have that \(e^{-3t} \leq e^{-3/\alpha}\). The last inequality holds because \( \tfrac{1}{\alpha} e^{-\frac{3}{\alpha}} \leq \frac{1}{3} e^{1/\alpha}\) for any \(\alpha > 0\).
\end{enumerate}
\end{proof}

The next lemma shows that the rate function in \pref{lem:pr_gf_rate} is admissible.

\begin{lemma} 
\label{lem:phase_retrieval_admissible}
Consider the function \(R\) defined as 
\begin{align*}
R(w, t) = \min \crl{F(w), F(w) e^{-t + \frac{1}{\tri{w, w^*}^2}}}. 
\end{align*}
Then, \(R\) is an admissible rate of convergence for the objective function \(F\). 
\end{lemma}
\begin{proof}[Proof of \pref{lem:phase_retrieval_admissible}] Recall that a sufficient conditions for a rate function \(R\) to be admissible w.r.t.~ the objective \(F\) is that for any point \(w\), 
\begin{align*}
\int_{t=0}^\infty \prn[\Big]{\frac{\partial R(w, t)}{\partial t} + \left\langle \grad_w  R(w, t), \nabla F(w) \right\rangle } \dif t \geq 0. \numberthis \label{eq:generalized_relation_pr}
\end{align*}
Since the function \(R\) is not differentiable at  \(t = {1}/{\tri{w, w^*}^2}\), we use the following definition of the partial derivative 
\begin{align*}
\frac{\partial R(w, t)}{\partial t} &=\begin{cases}
0 & \text{for} \quad t \leq 1/\tri{w, w^*}^2	 \\
- F(w) e^{-t + \frac{1}{\tri{w, w^*}^2}} & \text{for} \quad t > 1/\tri{w, w^*}^2
\end{cases},
\intertext{and}
\grad_w R(w, t) &= \begin{cases}
\grad F(w) & \text{for} \quad t \leq 1/\tri{w, w^*}^2	 \\
\prn*{\grad F(w) - 2 \frac{F(w) w^*}{\tri{w, w^*}^3} } \cdot e^{-t + \frac{1}{\tri{w, w^*}^2}} & \text{for} \quad t > 1/\tri{w, w^*}^2
\end{cases}. 
\end{align*}
Thus, we get that 
\begin{align*}
\int_{t=0}^\infty \frac{\partial R(w, t)}{\partial t} \dif t &= - F(w), 
\end{align*} 
and
\begin{align*}
\int_{t=0}^\infty \tri*{\grad_w R(w, t), \grad F(w)} \dif t &= \int_{t=0}^{\frac{1}{\tri{w, w^*}^2}} \tri*{\grad_w R(w, t), \grad F(w)} \dif t + \int_{\frac{1}{\tri{w, w^*}^2}}^{t=\infty} \tri*{\grad_w R(w, t), \grad F(w)} \dif t \\
&= \int_{t=0}^{\frac{1}{\tri{w, w^*}^2}} \nrm*{\grad F(w)}^2 \dif t \\
&\qquad \qquad \qquad + \int_{\frac{1}{\tri{w, w^*}^2}}^{\infty} \prn*{\nrm{\grad F(w)}^2 - 2 F(w) \frac{ \tri{\grad F(w), w^*}}{\tri{w, w^*}^3} } \cdot e^{-t + \frac{1}{\tri{w, w^*}^2}} \dif t\\ 
&= \int_{t=0}^{\frac{1}{\tri{w, w^*}^2}} \nrm*{\grad F(w)}^2 \dif t + \int_{0}^{\infty} \prn*{\nrm{\grad F(w)}^2 - 2 F(w) \frac{ \tri{\grad F(w), w^*}}{\tri{w, w^*}^3} } \cdot e^{-t} \dif t\\ 
&= \frac{ \nrm*{\grad F(w)}^2}{\tri{w, w^*}^2} + \nrm{\grad F(w)}^2 - 2 F(w) \frac{ \tri{\grad F(w), w^*}}{\tri{w, w^*}^3} \\
&=  \frac{ \nrm*{\grad F(w)}^2}{\tri{w, w^*}^2} + \nrm{\grad F(w)}^2 - 6 F(w) \frac{ (\nrm{w}^2 - 1)}{\tri{w, w^*}^2},  
\end{align*}
where the last line follows from the fact that  \(\grad F(w) = 3(\nrm{w}^2 - 1) w\). Plugging the above in \pref{eq:generalized_relation_pr}, we get that a sufficient condition for \(R\) to be an admissible rate of convergence is that 
\begin{align*}
&\frac{ \nrm*{\grad F(w)}^2}{\tri{w, w^*}^2} + \nrm{\grad F(w)}^2 - F(w) \prn*{\frac{ 6(\nrm{w}^2 - 1)}{\tri{w, w^*}^2} + 1} \geq 0,
\intertext{or equivalently that}
&\nrm*{\grad F(w)}^2 + \tri{w, w^*}^2 \nrm{\grad F(w)}^2 - F(w) \prn*{6\nrm{w}^2 - 6 + \tri{w, w^*}^2} \geq 0.  \numberthis \label{eq:pr_gf8}
\end{align*}

We next observe that \pref{eq:pr_gf8} holds if 
\begin{align*}
0 &\leq \nrm*{\grad F(w)}^2 - F(w) \prn*{6\nrm{w}^2 - 6 + \tri{w, w^*}^2} \\
&\overeq{\proman{1}} 12 \nrm{w}^2 F(w)  - 8 \prn*{\nrm{w}^2 - \tri{w, w^*}^2} - F(w) \prn*{6\nrm{w}^2 - 6 + \tri{w, w^*}^2} \\
&= F(w) \prn*{6 \nrm{w}^2 - \tri{w, w^*}^2 + 6} - 8(\nrm{w}^2 - \tri{w, w^*}^2) \\
&\overeq{\proman{2}} \prn[\Big]{\nrm{w}^2 - \tri{w, w^*}^2 + \frac{3}{4} \prn*{\nrm{w}^2 - 1}^2} \prn*{6 \nrm{w}^2 - \tri{w, w^*}^2 + 6} - 8(\nrm{w}^2 - \tri{w, w^*}^2) \numberthis \label{eq:pr_gf9}, 
\end{align*} where the \(\proman{1}\) and \(\proman{2}\) follow by plugging in the forms of \(\nrm{\grad F(w)}^2\) and \(F(w)\) from \pref{lem:phase_ret_F_dist_relation}. In the following, we argue that the relation \pref{eq:pr_gf9} holds for any \(w\). 

Consider the 2d function 
\begin{align*}
\Lambda (\alpha, \beta) &\ldef{} \prn[\Big]{\beta + \frac{3}{4} \prn*{\alpha + \beta - 1}^2} \prn*{5 \alpha + 6 \beta + 6} - 8 \beta
\end{align*}
and note that \(\Lambda(\alpha, \beta) \geq 0\) whenever \(\alpha \geq 0\) and \(\beta \geq 0\) (this can be easily checked by plotting the two dimensional function \(\Lambda\)). Setting \(\alpha = \tri{w, w^*}^2\) and \(\beta = \nrm{w}^2 - \tri{w, w^*}^2\), we note that both \(\alpha, \beta \geq 0\) and so \pref{eq:pr_gf9} follows immediately, which further implies that the relation in \pref{eq:pr_gf8} holds. Thus, the sufficient conditions for \(R\) to be an admissible rate of convergence hold, and the statement of the lemma follows. 
\end{proof}

\begin{proof}[Proof of \pref{lem:F_gf_pr}] We prove the rate of convergence in \pref{lem:pr_gf_rate} and show its admissibility in \pref{lem:phase_retrieval_admissible} above.	
\end{proof}

\subsubsection{Potential function and self-bounding regularity conditions}
Consider the function 
\begin{align*}
R(w, t) = \min \crl{F(w), F(w) e^{-t + \frac{1}{\tri{w, w^*}^2}}}. 
\end{align*}

\pref{lem:pr_gf_rate} and \pref{lem:phase_retrieval_admissible} imply that \(R\) is an admissible rate of convergence for the objective function \(F\). Thus, using \pref{thm:gf_to_potential} with \(g(z) = z\), we get that the function \(\Phi\)  constructed in the following is an admissible potential function for \(F\), 
\begin{align*}
\Phi(w) &= \int_{t=0}^\infty R(w, t) \dif t \\
&= \int_{t=0}^\infty \min \crl{F(w), F(w) e^{-t + \frac{1}{\tri{w, w^*}^2}}} \dif t \\
&= \int_{t=0}^{t=\tfrac{1}{\tri{w, w^*}^2}} \min \crl{F(w), F(w) e^{-t + \frac{1}{\tri{w, w^*}^2}}} \dif t  + \int_{t=\tfrac{1}{\tri{w, w^*}^2}}^\infty  \min \crl{F(w), F(w) e^{-t + \frac{1}{\tri{w, w^*}^2}}} \dif t \\
&= \int_{t=0}^{t=\tfrac{1}{\tri{w, w^*}^2}} F(w) \dif t  + \int_{t=\tfrac{1}{\tri{w, w^*}^2}}^\infty  F(w) e^{-t + \frac{1}{\tri{w, w^*}^2}} \dif t \\
&= \frac{F(w)}{\tri{w, w^*}^2} + F(w). \numberthis \label{eq:pr_potential}  
\end{align*} 

We first establish the self-bounding regularity conditions for \(F\).
\begin{lemma} 
\label{lem:pr_sb_F_g}
Let \(\nrm{w^*}=1\). For any point \(w\), 
\begin{align*} 
\nrm{\grad F(w)}^2 &\leq 12 F(w)^{3/2} + 12 F(w) 
\intertext{and}
\nrm{\grad^2 F(w)} &\leq 10 + 9 \sqrt{F(w)}. 
\end{align*}
\end{lemma}
\begin{proof}[Proof of \pref{lem:pr_sb_F_g}] We first bound \(\nrm{\grad F(w)}^2\). Using \pref{lem:phase_ret_F_dist_relation}-(c), we have that 
\begin{align*}
\nrm{\grad F(w)}^2 &= 12 \nrm{w}^2 F(w)  - 8 \prn*{\nrm{w}^2 - \tri{w, w^*}^2} \\
&\leq 12 \nrm{w}^2 F(w) \\
&\leq 12 \prn*{\abs*{\nrm{w}^2 - \nrm{w^*}^2} + \nrm{w^*}^2} F(w) \\
&\leq 12 \prn*{\sqrt{F(w)} + \nrm{w^*}^2} F(w) \\
&= 12 F(w)^{3/2} + 12 F(w), 
\end{align*} where the first inequality holds because $\nrm{w}^2 - \tri{w, w^*}^2 \geq 0$ whenever \(\nrm{w^*} \leq 1\), the second inequality is an application of the Triangle inequality and the last inequality follows from \pref{lem:phase_ret_F_dist_relation}-(d). The equality in the last line holds because \(\nrm{w^*} = 1\). Note that the function on the right hand size above is positive and monotonically increasing in \(F(w)\). 

We next bound \(\nrm{\grad^2 F(w)}\). From the form of \(F\) in \pref{lem:phase_ret_F_dist_relation}-(a), we get that 
\begin{align*}
\grad^2 F(w) &= 2\mathrm{I} - 2\prn{w^*} \prn{w^*}^\top + 3 \prn{\nrm{w}^2 - 1} \mathrm{I} + 6 w w^\top. 
\end{align*}

Thus, using Triangle inequality, we have 
\begin{align*}
\nrm{\grad^2 F(w)} &\leq 2 + 2 \nrm{w^*}^2 + 3 \prn{\nrm{w}^2 - 1} + 6 \nrm{w}^2 = 10 + 9  \prn{\nrm{w}^2 - 1} \leq 10 + 9 \sqrt{F(w)}, 
\end{align*} where the equality in the second line holds because \(\nrm{w^*} = 1\) and the last line is due to \pref{lem:phase_ret_F_dist_relation}-(d). 

\end{proof}

We next establish self-bounding regularity conditions for the potential function \(\Phi\). 
\begin{lemma} 
\label{lem:pr_Phi_sb}
Let \(\nrm{w^*}=1\). For any point \(w\), The function \(\Phi\) defined in \pref{eq:pr_potential} satisfies for any point \(w\), 
\begin{align*}
\nrm{\grad \Phi(w)} &\leq 39 \Phi(w)^2 + 17, 
\intertext{and}
\nrm{\grad^2 \Phi(w)} &\leq 54 \Phi(w)^3 + 215 \Phi^2(w) + 23 \Phi(w)  + 79 \leq 300 \Phi(w)^3 + 100. 
\end{align*} 
\end{lemma}
\begin{proof}[Proof of \pref{lem:pr_Phi_sb}] Before delving into self-bounding regularity conditions for \(\Phi\), we first derive an upper bound on \(1/\tri{w, w^*}^2\). Note that 
\begin{align*}
\abs*{1 - \tri{w, w^*}^2} &\leq \abs*{1 - \nrm{w}^2} + \abs*{\nrm{w}^2 - \tri{w, w^*}^2} \\
&\leq \sqrt{F(w)} +  \frac{3}{4} \abs*{  \prn*{\nrm{w}^2 - 1}^2 - F(w)} \\
&\leq \sqrt{F(w)} +  \frac{3}{4} \abs*{\nrm{w}^2 - 1}^2 + F(w) \\
&\leq \sqrt{F(w)} + 2 F(w), 
\end{align*} where the first and the third inequality above follows from Triangle inequality, and the second and the forth inequalities are due to \pref{lem:phase_ret_F_dist_relation}-(a, d). Squaring both the sides, we get that 
\begin{align*}
1 + \tri{w, w^*}^4 - 2 \tri{w, w^*}^2  &\leq 2F(w) + 8 F(w)^2. 
\end{align*}
Ignoring positive terms on the left hand size and dividing both the sides by \(\tri{w, w^*}^2\), we get that 
\begin{align*}
\frac{1}{\tri{w, w^*}^2} &\leq 2 + 2\frac{F(w)}{\tri{w, w^*}^2} + 8 \frac{F(w)^2}{\tri{w, w^*}^2} \\ 
&\leq 2 + 2 \Phi(w) + 8 F(w) \Phi(w) \\
&\leq 2 + 2 \Phi(w) + 8 \Phi^2(w)  \\
&\leq 3 + 9 \Phi^2(w), \numberthis \label{eq:pr_sb_1}
\end{align*} where the inequalities in second and the third line follow from the fact that both \(F(w)/\tri{w, w^*}^2\) and \(F(w)\) are smaller than \(\Phi(w)\) (from the definition in \pref{eq:pr_potential} and because \(F(w) \geq 0\)).  The last line is due to AM-GM inequality. 

We now prove the self-bounding regularity conditions for \(w\). 
\begin{enumerate}[label=\(\bullet\)]
\item \textit{Bound on \(\nrm{\grad \Phi(w)}\).}  Note that 
\begin{align*}
\grad \Phi(w) &= \frac{\grad F(w)}{\tri{w, w^*}^2} - \frac{2 F(w)}{\tri{w, w^*}^3} w^* + \grad F(w). 
\end{align*}
Using Triangle inequality and the fact that \(\nrm{w^*} = 1\), we get 
\begin{align*}
\nrm{\grad \Phi(w)} &\leq \frac{\nrm{\grad F(w)}}{\tri{w, w^*}^2} + 2 \frac{F(w)}{\tri{w, w^*}^2} \cdot \frac{1}{\abs{\tri{w, w^*}}} + \nrm{\grad F(w)} \\
&\overleq{\proman{1}}  \nrm{\grad F(w)} \prn*{\frac{1}{\tri{w, w^*}^2} + 1} + 2 \Phi(w) \cdot \frac{1}{\abs{\tri{w, w^*}}} \\
&\overleq{\proman{2}}  \sqrt{12 F(w)^{3/2} + 12 F(w)} \prn*{\frac{1}{\tri{w, w^*}^2} + 1} + 2 \Phi(w) \cdot \frac{1}{\abs{\tri{w, w^*}}} \\ 
&\overleq{\proman{3}}  \sqrt{15 F^2(w) +  9} \cdot \prn*{\frac{1}{\tri{w, w^*}^2} + 1} +\Phi^2(w) + \frac{1}{\tri{w, w^*}^2 } \\
&\overleq{\proman{4}}  4 F(w) \prn*{\frac{1}{\tri{w, w^*}^2} + 1}  + 3 +\Phi^2(w) + \frac{4}{\tri{w, w^*}^2 },  
\end{align*} where \(\proman{1}\) holds because \(F(w)/  \tri{w, w^*}^2 \leq \Phi(w)\), \(\proman{2}\) is due to \pref{lem:pr_sb_F_g}  and \(\proman{3}\) follows from multiple applications of AM-GM inequality. The inequality $\proman{4}$ is due to subadditivity of square-root and from rearranging the terms. Plugging in the bound in \pref{eq:pr_sb_1} and the definition in \pref{eq:pr_potential} in the above, we get that 
\begin{align*}
\nrm{\grad \Phi(w)} 
&\leq 37 \Phi(w)^2 + 4 \Phi(w) + 15 \\ 
&\leq 39 \Phi(w)^2 + 17,  \numberthis \label{eq:pr_sb_2}  
\end{align*} where the last line holds due to AM-GM inequality. 

\item \textit{Bound on \(\nrm{\grad^2 \Phi(w)}\).} Note that 
\begin{align*}
\grad^2 \Phi(w) &= \grad^2 F(w) \prn*{\frac{1}{\tri{w, w^*}^2} + 1}   - 2 \frac{\grad F(w) (w^*)^\top}{\tri{w, w^*}^3} - 2 \frac{w^* (\grad F(w))^\top}{\tri{w, w^*}^3} + 6 \frac{F(w) \cdot (w^*)(w^*)^\top}{\tri{w, w^*}^4}.  \numberthis \label{eq:pr_sb_4} 
\end{align*}
Using Triangle inequality, Cauchy Schwartz inequality and the fact that \(\nrm{w^*} = 1\), we get 
\begin{align*}
\nrm{\grad^2 \Phi(w)} &\leq \nrm{\grad^2 F(w)} \prn*{\frac{1}{\tri{w, w^*}^2} + 1} + 4 \frac{\nrm{\grad F(w)}}{\abs{\tri{w, w^*}}^3} + 6 \frac{F(w)}{\tri{w, w^*}^4}. 
\end{align*}
We bound each of the terms separately below: 
\begin{enumerate}
\item \textit{Term I:} Using \pref{lem:pr_sb_F_g}, we get that 
\begin{align*}
 \nrm{\grad^2 F(w)} \prn*{\frac{1}{\tri{w, w^*}^2} + 1} &\leq \prn{10 + 9 \sqrt{F(w)}} \prn*{\frac{1}{\tri{w, w^*}^2} + 1} \\
 &\leq 10 + \frac{10}{\tri{w, w^*}^2} + \frac{9 \sqrt{F(w)}}{\tri{w, w^*}^2} \\ 
 &\leq 10 + \frac{10}{\tri{w, w^*}^2} + \frac{9}{2 \tri{w, w^*}^2} + \frac{9}{2} \frac{F(w)}{\tri{w, w^*}^2} \\
 &\leq 55 +  135 \Phi^2(w) + \frac{9}{2} \Phi(w), 
\end{align*} where the second line is due to AM-GM inequality and the last line follows from plugging in \pref{eq:pr_potential} and \pref{eq:pr_sb_1}. 

\item \textit{Term II:} Using the bound from \pref{lem:pr_sb_F_g}, we get 
\begin{align*}
4 \frac{\nrm{\grad F(w)}}{\abs{\tri{w, w^*}}^3} &\leq 4 \sqrt{12 F(w)^{3/2} + 12 F(w)} \cdot \ \frac{1}{\abs{\tri{w, w^*}^3}} \\
&\leq \prn*{16 F(w) + 12} \cdot \frac{1}{\abs{\tri{w, w^*}^3}} \\
&\leq 16 \Phi(w) \cdot \frac{1}{\abs{\tri{w, w^*}}} \\
&\leq 24 + 80 \Phi^2(w), 
\end{align*}
where the line line is due to AM-GM inequality and subadditivity of square-root, the third line is due to \pref{eq:pr_potential}, the forth line again uses AM-GM inequality and the last line follows from plugging in the bound in  \pref{eq:pr_sb_1}. 

\item \textit{Term III:} Using the fact that \(F(w) / \tri{w, w^*}^2 \leq \Phi(w)\) from \pref{eq:pr_potential}, we get that 
\begin{align*}
 \frac{6F(w)}{\tri{w, w^*}^4} &\leq 6 \Phi(w) \cdot \frac{1}{\tri{w, w^*}^2} \\ 
 &\leq 18 \Phi(w) + 54 \Phi(w)^3, 
\end{align*} where the second inequality follows by plugging \pref{eq:pr_sb_1}. 
\end{enumerate}

Plugging the three bounds above in \pref{eq:pr_sb_4}, we get that 
\begin{align*}
\nrm{\grad^2 \Phi(w)} &\leq 54 \Phi(w)^3 + 215 \Phi^2(w) + 23 \Phi(w)  + 79. 
\end{align*}

\end{enumerate}
\end{proof}

\subsubsection{GD for phase retrieval} \label{app:GD_pr}
In the following, we provide the convergence guarantee for GD algorithm. We first define the respective problem dependent quantities and instantiate \pref{thm:GD_guarantee}  to provide an \(O\prn*{{1}/{T}}\) bound for GD. We then provide a refined analysis which improves this bound to \(O(e^{-T})\). 

\paragraph{\(\mb{O(1/T)}\) rate by direct application of \pref{thm:GD_guarantee}.}
\begin{itemize}
\item Setting  \(g(z) = z \) implies the potential function 
\begin{align*}
\Phi(w) = \frac{F(w)}{\tri{w, w^*}^2} + F(w). 
\end{align*}
\item \pref{ass:F_regular} follows from \pref{lem:pr_sb_F_g} which implies that
\begin{align*}
\psi(z) = 12 z^{3/2} + 12 z. 
\end{align*} 
\item \pref{ass:Phi_regular} follows from  \pref{lem:pr_Phi_sb} which implies that
\begin{align*}
\rho(z) &=  300 z^3 + 100. 
\end{align*}
\item The function \(\theta\) is given by
\begin{align*}
\theta(z) =  \int_{y=0}^{z} \frac{1}{\rho(y)} \dif y \leq \frac{z}{100}. 
\end{align*}
\item The monotonically increasing function \(\zeta\) is defined such that  
\begin{align*}
\zeta^{-1}(z) = \int_{y=0}^{z} \frac{\bridge{y}}{\psi(y)} \dif y = \frac{1}{6} \prn*{ \sqrt{z} - \log(1 + \sqrt{z})} \geq \frac{1}{12} \sqrt{z}, 
\end{align*}
which implies that 
\begin{align*}
\zeta(z) \leq 144 z^2. 
\end{align*} 
\end{itemize} 

Note that the function \(\tfrac{\psi(z)}{g(z)} = 12 \sqrt{z} + 12\) is clearly a monotonically increasing function of \(z\). Thus, plugging the above problem-dependent constants in \pref{thm:GD_guarantee}  implies that setting \(\eta\) such that  
\begin{align*}
\eta &\propto \frac{1}{(1 + \Phi(w_0))(1 + \Phi(w_0)^3)}
\end{align*}
implies that GD for any \(T \geq 1\) has the rate 
\begin{align*}
\bridge{F(\wh w_T)} &\lesssim \frac{\Phi(w_0) + \Phi(w_0)^8}{T},  \numberthis \label{eq:pr_slow_rate}
\end{align*}
where recall that \(\Phi(w_0) = \frac{F(w_0)}{\tri{w_0, w^*}^2} + F(w_0) \). 

\paragraph{\(\mb{O(e^{-T})}\) rate via a refined analysis.} 
We can further improve over the rate in \pref{eq:pr_slow_rate} by a refined analysis for GD. In the following, we will show that GD in fact enjoys a \(e^{- O(T - \tau)}\) rate of convergence for GD for all \(T \geq \tau\), where \(\tau\) depends on \(w_0\) and problem dependent parameters specified below. 

Before delving into the proof of the above, we first provide the relevant improved version of problem dependent parameters that hold for any \(w\) for which \(F(w) \leq 1\):  
\begin{itemize}
\item \pref{ass:F_regular} follows from \pref{lem:pr_sb_F_g} which implies that
\begin{align*}
\psi(z) =  24 z. 
\end{align*} 
\item \pref{ass:Phi_regular} follows from  \pref{lem:pr_Phi_sb} which implies that
\begin{align*}
\rho(z) &= 400. 
\end{align*}
\item The function \(\theta\) is given by
\begin{align*}
\theta(z) =  \int_{y=0}^{z} \frac{1}{\rho(y)} \dif y = \frac{z}{400}.  \numberthis \label{eq:pr_new_theta}
\end{align*} 
\end{itemize}

We are now ready to provide the improved convergence rate for GD. Note that using \pref{eq:pr_slow_rate}, there exists some 
\begin{align*}
\tau \leq 20 \prn*{\Phi(w_0) + \Phi(w_0)^8} \numberthis \label{eq:pr_advanced_4} 
\end{align*}
for which \(F(w_{\tau}) \leq 1/20\). Using \pref{lem:phase_ret_F_dist_relation}-(e), we get that such a point \(w_\tau\) must satisfy \(\tri{w_\tau, w^*}^2 \geq 1/4\), which implies that 
\begin{align*}
\Phi(w_{\tau}) &= \frac{F(w_\tau)}{\tri{w_\tau, w^*}^2} + F(w_\tau) \leq 5 F(w_\tau) \leq \frac{1}{4}. 
\end{align*}

In the following, we first show via induction that \(\tri{w_t, w^*} \geq 1/4\) and \(\Phi(w_t) \leq 1/4\) for all \(t \geq \tau\). As shown above, the base case for \(t = \tau\) holds. For the induction step, consider any \(t \geq \tau\) and assume that  \(\tri{w_{t}, w^*}^2 \geq 1/4\) and \(\Phi(w_t) \leq 1/4\); we will show that the same holds for \(w_{t+1}\). Starting from \pref{eq:GDsb_11} in the proof of \pref{thm:GD_guarantee}, we note that 
\begin{align*}
\taylor(\Phi(w_{t +1})) &\leq \taylor(\Phi(w_t))  - \frac{\eta}{2\rho(\Phi(w_0))} \bridge{F(w_t))}. \numberthis \label{eq:pr_advanced_1}
\end{align*} 
However, also note that \(w_t\) satisfies, 
\begin{align*}
F(w_t) \leq \Phi(w_t) = \frac{F(w_t)}{\tri{w_t, w^*}^2} + F(w_t) \leq 5 F(w_t),  \numberthis \label{eq:pr_advanced_2}
\end{align*} where the last inequality holds since \(\tri{w_t, w^*}^2 \geq 1/4\) by induction hypothesis. Plugging the relation \pref{eq:pr_advanced_2} in \pref{eq:pr_advanced_1} and using the fact that \(g(z) = z\), we get that 
\begin{align*}
\taylor(\Phi(w_{t +1})) &\leq \taylor(\Phi(w_t))  - \frac{\eta}{10\rho(\Phi(w_0))} \Phi(w_t), 
\end{align*} 
Plugging in the value of \(\theta\) and \(\rho\) from \pref{eq:pr_new_theta} in the above, we get that 
\begin{align*}
\Phi(w_{t +1}) &\leq \Phi(w_t) - \frac{\eta}{10} \Phi(w_t) \\ 
&= \Phi(w_t)\prn*{1 - \frac{\eta}{10}}. \numberthis \label{eq:pr_advanced_3} 
\end{align*}

The above clearly implies that \(\Phi(w_{t + 1}) \leq \Phi(w_t) \leq 1/4\). Furthermore, from the definition of \(\Phi\), we immediately get that \(F(w_{t + 1}) \leq 1/4\), plugging which in \pref{lem:phase_ret_F_dist_relation}-(e) implies that \(\tri{w_{t + 1}, w^*}^2 \geq 1/4\). This completes the induction step hence showing that \(\tri{w_t, w^*} \geq 1/4\) and \(\Phi(w_t) \leq 1/4\) holds for all \(t \geq \tau\). 

Now, in order to complete the proof of convergence, note that \pref{eq:pr_advanced_3} will hold for all \(t \geq \tau\), recursing which implies that 
\begin{align*}
\Phi(w_{T}) &\leq \Phi(w_\tau) \prn*{1 - \frac{\eta}{10}}^{T - \tau} \leq \Phi(w_\tau) e^{- \eta (T - \tau)/10} \leq \frac{1}{4} e^{- \eta (T - \tau)/10},
\end{align*} where the last inequality holds since \(\Phi(w_\tau) \leq 1/4\). 

Plugging in the value of \(\tau\) from \pref{eq:pr_advanced_3}, we get that for all \(T \geq \tau = 20 \prn*{\Phi(w_0) + \Phi(w_0)^8}\), GD has convergence rate 
\begin{align*}
F(w_T) \leq \Phi(w_{T})  \leq \frac{1}{4} e^{- \frac{\eta (T - \tau)}{10}}. \numberthis \label{eq:pr_advanced_4}  
\end{align*}

\subsubsection{SGD for phase retrieval} 
We build on the problem dependent quantities introduced in \pref{app:GD_pr}. Suppose SGD is run with stochastic gradient estimates that satisfy \pref{ass:gradient_noise} with
\begin{align*}
\chi(z) = \min\crl{\sqrt{z}, c}, 
\end{align*}
where \(c\) is a universal constant. Such a bound is satisfied when the stochastic gradient estimate is  computed by using samples from \(\cS\) where a fresh sample is used for each estimate, i.e. \(\grad f(w; (a, y)) = 4\prn{(a^\top w)^2 - y} (a^\top w)w\) (c.f. \citet[Lemma 7.4, 7.7]{candes2015phase}). Using the above, we define the function $\Lambda$ used in \pref{thm:SGD_guarantee} as 
\begin{align*}
\Lambda(z) = 24 z^{3/2} + 24 z + 2 \min\crl{\sqrt{z}, c}.  
\end{align*}

Fixing any \(\bw\) such that \(F(\bw) \geq F(w_0)\), set \(\kappa = F(\bw)/F(w_0)\), and define \(B = 24 \Phi(\bw)^3 + 24 \Phi(\bw)^2 + 2 \min\crl{\Phi(\bw), c} \lesssim (1 + \Phi(\bw)^3)\). The following guarantee is due to \pref{thm:SGD_guarantee} (in particular the bound in \pref{rem:SGD_bound}). Setting 
\begin{align*}
\eta \leq \frac{1}{2 \log^2(20T) } \cdot \frac{\Phi(\bw) - \Phi(w_0)}{\sqrt{B \Phi(\bw) T}}, 
\end{align*}
the point returned by SGD algorithm after \(T\) iterations satisfies with probability at least \(0.7\), 
\begin{align*} 
\bridge{F(\wh w_T)} &\lesssim \rho(\Phi(\bw))\cdot \frac{\Phi(\bw)}{\Phi(\bw) - \Phi(w_0)} \cdot \sqrt{ B \Phi(\bw)} \cdot \frac{1}{\sqrt{T}}, 
\end{align*} 
where recall that \(\Phi(w) = \frac{F(w)}{\tri{w, w^*}^2} + F(w)\). Since \(g(z) = z\), the above immediately implies a bound on \(F(\wh w_T)\).

%% file: files/appx_applications_utility.tex
\subsection{Proof of \pref{lem:indicator_utility_lemma}} 
The proof of \pref{lem:indicator_utility_lemma} follows by defining a rate function which holds for every initial point. We then get an admissible potential function by using \pref{thm:gf_to_potential}. The desired self-bounding regularity conditions follow by plugging in the given properties of \(\Gamma\) and \(h\) in the lemma statement.

\begin{proof}[Proof of \pref{lem:indicator_utility_lemma}] Note that for any initialization \(w(0) = w\) for which \(h(w) \geq 0\), gradient flow satisfies \(F(w(t))  \leq R(w, t).\) Define the function \(\wt{R}(w, t) = R(w, h(w) t)\). Clearly, for any \(w\), 
\begin{align*} 
	F(w(t)) \leq \wt{R}(w, t) = R(w, h(w) t). 
\end{align*} To see the above, note that when \(h(w) = 0\), the above relation simply reduces to \(F(w(t)) \leq R(w, 0)\) which holds from our assumptions. When \(0 < h(w) \leq 1\), we have that \(F(w(t)) = R(w, t) \leq R(w, h(w)t)\) which again holds because \(R(w, \cdot)\) is monotonically decreasing in \(t\) and because \(h(w) \leq 1\). 

Next, using the premise that \(\wt R\) is admissible rate function w.r.t.~ \(F\), and \pref{thm:gf_to_potential}, we get that the function \(\Phi_g\) defined below is an admissible  potential function w.r.t.~ \(F\) with \(g(z) = z\), 
\begin{align*} 
\Phi_g(w)&= \int_{t=0}^\infty \bR(w, t) \dif t = \int_{t=0}^\infty R(w, h(w) t) \dif t = \frac{\Gamma(w)}{h(w)}. 
\end{align*} 

In the following, we show that \pref{ass:Phi_regular} (self-bounding regularity conditions) hold for the potential function \(\Phi_g\). First note that, for any \(w\), the assumption  \(\prn*{h(w) - h(w^*)}^2 \leq \mu(\Gamma(w))\) implies that 
\begin{align*}
 \mu(\Gamma(w)) &\geq h(w^*)^2 + h(w)^2 - 2 h(w)h(w^*) \\
 &\geq h(w^*)^2  - 2 h(w) h(w^*),
\intertext{which after rearranging the terms implies that}
\frac{1}{h(w)} &\leq \frac{1}{h(w^*)^2} \prn*{ 2  h(w^*) + \frac{ \mu(\Gamma(w))}{h(w)} }  \\
&\leq \frac{1}{h(w^*)^2} \prn*{ 2  h(w^*) +  \mu\prn*{\frac{\Gamma(w)}{h(w)}} }  \\ 
&= \frac{1}{h(w^*)^2} \prn*{ 2  h(w^*) +  \mu\prn*{\Phi_g(w)}}, \numberthis \label{eq:any_initialization1}
\end{align*} where the second inequality holds because \(h(w) \leq 1\) and \(\mu\) satisfies the property that \(k\pi(z) \leq \pi(k z)\) for any \(k \geq 1\).  

We are now ready to establish the self-bounding regularity properties for \(\Phi_g\). 
\begin{enumerate}[label=\((\alph*)\)]  
\item  \textit{$\nrm{\grad \Phi_g(w)}$ satisfies self-bounding regularity.} Using Chain rule and Triangle inequality, we have that  
\begin{align*}
\nrm{\grad \Phi_g(w)} &\leq \frac{\nrm{\grad \Gamma(w)}}{h(w)}  + \frac{\Gamma(w)}{h(w)^2}  \nrm{\grad h(w)} \\ 
&\overleq{\proman{1}} \frac{\lambda(\Gamma(w))}{h(w)}  + \Phi_g(w) \frac{\pi\prn{\Gamma(w)}} {h(w)} \\
&\overleq{\proman{2}}  \frac{1}{h(w)}\lambda\prn*{\frac{\Gamma(w)}{h(w)}}  + \frac{1}{h(w)} \Phi_g(w) \pi\prn*{\frac{\Gamma(w)}{h(w)}} \\
&= \frac{1}{h(w)} \lambda(\Phi_g(w)) + \frac{1}{h(w)}\Phi_g(w) \pi(\Phi_g(w)) \\
&\overleq{\proman{3}} \prn*{ \frac{2}{ h(w^*)} +   \frac{ \mu\prn*{\Phi_g(w)}}{ h(w^*)^2}} \cdot \prn*{\lambda(\Phi_g(w)) + \Phi_g(w) \pi(\Phi_g(w))  }
\end{align*} where \(\proman{1}\) holds due to the assumption that \(\nrm{\grad \Gamma(w)} \leq \lambda(\Gamma(w))\) and \(\nrm{\grad h(w)} \leq \pi(\Gamma(w))\), \(\proman{2}\) holds because \(\lambda\) and \(\pi\) are positive, monotonically increasing functions and \(h(w) \leq 1\). The equality in the next line follows from the definition of \(\Phi_g(w)\), and the inequality \(\proman{3}\) follows from plugging in \pref{eq:any_initialization1}. 

Note that the function 
\begin{align*}
\zeta(z) = \frac{1}{h(w^*)^2} \prn*{ 2  h(w^*) +  \mu\prn*{z}} \cdot \prn*{\lambda(z) + z \pi(z)  }
\end{align*}
appearing on the right side above is positive, monotonically increasing. 

\item \textit{$\nrm{\grad^2 \Phi_g(w)}$ satisfies self-bounding regularity.} Using Chain rule and Triangle inequality, we get that 
\begin{align*} 
\nrm{\grad^2 \Phi_g(w)} &\leq  \frac{\nrm{\grad^2 \Gamma(w)}}{h(w)}  + 2 \frac{\nrm{\grad \Gamma(w)  \grad h(w)^\top}}{h(w)^2}  + \frac{\Gamma(w)}{h(w)^3} \nrm{ \grad h(w)}^2  +  \frac{\Gamma(w)}{h(w)^2}  \nrm{\grad^2 h(w)}.  \numberthis \label{eq:utility_lemma1} 
\end{align*} 
We bound each of the terms in the RHS above separately, as follows: 
\begin{enumerate}[label=\(\bullet\), leftmargin=5mm] 
\item For the first term in \pref{eq:utility_lemma1}, using the relation \(\nrm{\grad^2 \Gamma(w)} \leq \lambda(\Gamma(w))\), we get 
\begin{align*}
\frac{\nrm{\grad ^2 \Gamma(w)}}{h(w)} &\leq \frac{\lambda(\Gamma(w))}{h(w)} \\ 
&\leq \lambda(\Gamma(w)) \cdot \prn*{ \frac{2}{ h(w^*)} +   \frac{ \mu\prn*{\Phi_g(w)}}{ h(w^*)^2}}  \\
&\leq  \lambda(\Phi_g(w)) \cdot \prn*{ \frac{2}{ h(w^*)} +   \frac{ \mu\prn*{\Phi_g(w)}}{ h(w^*)^2}}, 
\end{align*} 
where the second inequality is by plugging in \pref{eq:any_initialization1}, and the last line follows from the fact that  \(h(w) \leq [0, 1]\) and from the definition of \(\Phi_g(w)\). This proves self-bounding regularity conditions for \(\grad \Phi_g(w)\)
\item  For the second term in \pref{eq:utility_lemma1}, using Cauchy-Schwarz inequality, we have 
\begin{align*}
	\frac{2}{h(w)^2}  \nrm{\grad \Gamma(w) \grad h(w)^\top} &\leq \frac{2}{h(w)^2}  \nrm{\grad \Gamma(w)} \nrm{\grad h(w)}  \\
	&\leq 2 \lambda \prn{\Gamma(w)} \cdot \pi\prn{\Gamma(w)} \cdot \prn*{ \frac{2}{ h(w^*)} +   \frac{ \mu\prn*{\Phi_g(w)}}{ h(w^*)^2}}^2 \\ 
&\leq 2 \lambda \prn{\Phi_g(w)} \cdot \pi\prn{\Phi_g(w)} \cdot  \prn*{ \frac{2}{ h(w^*)} +   \frac{ \mu\prn*{\Phi_g(w)}}{ h(w^*)^2}}^2 
\end{align*} 
where the second inequality holds because \(\nrm{\grad \Gamma(w)} \leq \lambda(\Gamma(w))\) and \(\nrm{\grad h(w)} \leq \pi(\Gamma(w))\), and the last inequality follows from the definition of \(\Phi_g(w)\) and the fact that \(h(w) \leq 1\). 
\item For the third term in \pref{eq:utility_lemma1}, using the relation \(\nrm{\grad h(w)} \leq \pi(\Gamma(w))\), we get 
\begin{align*}
\frac{\Gamma(w)}{h(w)^3} \nrm{ \grad h(w)}^2 &= \frac{\Gamma(w)}{h(w)} \cdot \frac{1}{h(w)^2} \cdot \pi^2(\Gamma(w)) \\ 
&\leq \Phi_g(w) \cdot \prn*{ \frac{2}{ h(w^*)} +   \frac{ \mu\prn*{\Phi_g(w)}}{ h(w^*)^2}}^2  \cdot \pi^2(\Phi_g(w)),
\end{align*} where the last line uses the definition of \(\Phi_g\), the fact that \(\pi\) is positive and monotonically increasing, \(h(w) \leq 1\), and the bound in \pref{eq:any_initialization1}. 

\item For the fourth term in \pref{eq:utility_lemma1},  using the relation \(\nrm{\grad^2 h(w)} \leq \pi(\Gamma(w))\), we get 
\begin{align*} 
\frac{\Gamma(w)}{h(w)^2}  \nrm{\grad^2 h(w)} &\leq \frac{\Gamma(w)}{h(w)} \cdot \frac{1}{h(w)} \cdot \pi(\Gamma(W)) \\
&\leq \Phi_g(w) \cdot \prn*{ \frac{2}{ h(w^*)} +   \frac{ \mu\prn*{\Phi_g(w)}}{ h(w^*)^2}} \cdot  \pi(\Phi_g(w)) \end{align*} where the last line uses the definition of \(\Phi_g\), the fact that \(\pi\) is positive and monotonically increasing and the fact that \(h(w) \leq 1\), and the bound in \pref{eq:any_initialization1}. 
\end{enumerate}

Clearly, each of the bounds above consists of a positive, monotonically increasing function on the right hand side, thus proving self-bounding regularity conditions for \(\grad^2 \Phi_g(w)\). 
\end{enumerate} 
\end{proof}

%% file: files/appx_applications_matrix_completion.tex
\subsection{Matrix Square root}
For any symmetric \(W \in \bbR^{d \times d}\), the population loss for matrix square root problem is given by\footnote{Following the convention, we denote matrix valued variables throughout this section using capital Roman aphabet.}
\begin{align*} 
F(W) = \nrm{W^2 - M}^2_F,  \numberthis \label{eq:msqrt_F} 
\end{align*} where \(M\) is a positive-definite matrix. Note that the global minima of the above objective is obtained at \(W = \sqrt{M}\). 

The following technical lemma establishes some useful properties of \(F\). 
\begin{lemma}
\label{lem:msqrt_F_properties} 
 The function \(F\) given in \pref{eq:msqrt_F} satisfies for any \(W\), 
\begin{enumerate}[label=\((\alph*)\)]
\item \(\grad F(W) = 2(2 W^3 - M W - WM)\), 
\item $\nrm*{\grad F(W)}_F^2 \geq 16 \smin(W^2) F(W)$, 
\end{enumerate} where \(\smin(W)\) denotes the minimum singular value of \(W\). 
\end{lemma} 
\begin{proof} 
\begin{enumerate}[label=$(\alph*)$]
\item The relation follows from Chain rule. 
\item The proof is identical to the proof of {\citet[Lemma 4.5]{jain2017global}}. Note that 
\begin{align*}
   \tri{\grad F(W), \grad F(W)} &= 4 \tri*{ (W^2 - M )W + W(W^2 - M), (W^2 - M )W + W(W^2 - M)} \\
   &\geq 16 \smin(W^2) F(W). 
\end{align*}
\end{enumerate} 
\end{proof}

\subsubsection{Rate of convergence for gradient flow}
We first note the following technical lemma whose proof is identical to the proof of \citet[Lemma 4.2]{jain2017global} as \(\eta \rightarrow 0\). 

\begin{lemma}[{\citet[Lemma 4.2]{jain2017global}}] 
\label{lem:msqrt_smin_inc} 
For any initial point \(W_0\) and \(t \geq 0\), the point $W(t)$ on the gradient flow path with \(W(0) = W_0\) satisfies 
\begin{align*}
\smin(W(t)^2) &\geq \min \crl*{ \smin(W_0^2), \frac{\smin(M)}{100}}. 
\end{align*} 
\end{lemma} 

Before providing a rate of convergence for GF for the matrix square root problem, we first define additional notation. Let $\alpha = \smin(M)/1600$, and define the function 
\begin{align*}
\phi(Z) &= \frac{-1}{\gamma} \log(\trace{e^{- \gamma Z}} + e^{- 16 \alpha \gamma }), \numberthis \label{eq:phi_defn_msqrt}
\intertext{and the function}
h(W) &=  \sigma\prn*{\phi(W^2) - \alpha},   \numberthis \label{eq:h_defn_msqrt}
\end{align*} where \(\sigma\) denotes a smoothened version of the indicator function and is given by  
\begin{align*}
\sigma(z) &\ldef{} \begin{cases} 
 0 & \text{if} \quad z \leq 0 \\ 
	\frac{2}{\alpha^2} z^2 & \text{if} \quad 0 \leq z \leq \alpha/ 2  \\
	- \frac{2}{\alpha^2} z^2 + \frac{4}{\alpha} z - 1 & \text{if} \quad \alpha/2 \leq z  \leq \alpha \\ 
	1 & \text{if} \quad \alpha \leq z 
\end{cases}. \numberthis \label{eq:sigma_defn_msqrt}
\end{align*}

The following technical lemma establishes some useful properties of the function \(\phi\) and \(h\).  

\begin{lemma} 
\label{lem:msqrt_h_properties} 
 Let \(\gamma > 0\).  For any point \(W\), we have 
\begin{enumerate}[label=\((\alph*)\)]
\item \(\min \crl*{\smin(W^2), 16 \alpha} - \frac{\log(d + 1)}{\gamma} \leq \phi(W^2) \leq \min \crl*{\smin(W^2), 16 \alpha}\). 
\item \(\grad\subs{W} \phi(W) = \tfrac{e^{- \gamma W}}{\trace{e^{-\gamma W}} + e^{- 16 \gamma \alpha}}\) and \(\grad\subs{W} \phi(W^2) = \tfrac{ 2 e^{- \gamma W^2} W}{\trace{e^{-\gamma W^2}} + e^{- 16 \gamma \alpha}}\). 
\item \(\prn*{h(W) - h(\sqrt{M})}^2 \leq \frac{2}{\alpha} F(W)\). 
\item \(\nrm{\grad h(W)} \leq \frac{4}{\alpha} \prn*{F(W)^{1/4} + \sqrt{\nrm{M}}}\). 
\item \(\nrm{\grad^2 h(W)}  \leq  16 \prn*{\frac{2}{\alpha^2} + \frac{1}{\alpha}} \prn*{1 + \gamma \nrm{M} + \gamma \sqrt{F(W)}}.\)
\item if \(F(W) \leq \sigma_d(M)^2 / 4\), then \(W\) must satisfy \(\sigma_d(W^2) \geq 800 \alpha\). Furthermore, if \(\gamma \geq \tfrac{\log(d + 1)}{\gamma}\), the \(W\) satisfies \(h(W) = 1\). 
\end{enumerate}
where $\alpha = \smin(M)/1600$. 
\end{lemma} 
\begin{proof}[Proof of \pref{lem:msqrt_h_properties}] We prove each part separately below: 
\begin{enumerate}[label=\((\alph*)\)]
\item For the upper bound, note that 
\begin{align*} 
\phi(W^2) &= \frac{-1}{\gamma} \log(\sum_{i=1}^d {e^{- \gamma \sigma_i(W^2)}} + e^{- 16 \alpha \gamma })  \\ 
&\leq \frac{-1}{\gamma} \log(\min \crl*{ {e^{- \gamma \sigma_d(W^2)}}, e^{- 16 \alpha \gamma } }) \\
&= \min \crl*{\smin(W^2), 16 \alpha}, 
\end{align*}
where the inequality in the second line holds because \(- \log(z)\) is a decreasing function of \(z\). 

For the lower bound, again using monotonicity of the function \(-\log(z)\), we get that 
\begin{align*}
\phi(W^2) &= \frac{-1}{\gamma} \log(\sum_{i=1}^d {e^{- \gamma \sigma_i(W^2)}} + e^{- 16 \alpha \gamma })  \\ 
                 &\geq  \frac{-1}{\gamma} \log((d + 1) {e^{- \gamma \min \crl*{\smin(W^2), 16 \alpha}} })  \\
                 &\geq \min\crl*{\smin(W^2), 16 \alpha} - \frac{\log(d + 1)}{\gamma}. 
\end{align*} 

\item The proof is a straightforward application of the Chain rule for matrix derivatives.  
\item Since \(\sigma\) is \(2/\alpha\)-Lipschitz, we have that 
\begin{align*}
\prn{h(W) - h(\sqrt{M})}^2 &= \prn*{\sigma \prn{\phi(W^2) - \alpha} - \sigma \prn{\phi(M) - \alpha} }^2 \\
&\leq \frac{2}{\alpha} \prn*{\phi(W^2) - \phi(M)}^2 \\
&\leq \frac{2}{\alpha} \sup_{\substack{ t \in [0, 1] \\ Z = M t + (1 - t)W^2}}\nrm{\grad\subs{Z} \phi(Z)}_F^2 \cdot \nrm{W^2 - M}_F^2 \\ 
&=\frac{2}{\alpha} \sup_{\substack{ t \in [0, 1] \\ Z = M t + (1 - t)W^2}}\nrm*{\frac{e^{- \gamma Z}}{\trace{e^{-\gamma Z}} + e^{- 16 \gamma \alpha}}}_F^2 \cdot \nrm{W^2 - M}_F^2 \\ 
&\leq \frac{2}{\alpha} \nrm{W^2 - M}^2 = \frac{2}{\alpha} F(W) , 
\end{align*} 
where the inequality in the third line above holds due to Fundamental theorem of calculus and using Cauchy-Schwarz. The inequality is due to the fact that the first term in the product is always smaller than \(1\). 

\item Using Chain rule for matrix derivatives, we get that 
\begin{align*}
\nrm{\grad h(W)} &= \sigma'(\phi(W^2) - \alpha) \nrm{\grad\subs{W} \phi(W^2)} \\ 
&\leq \frac{2}{\alpha} \nrm{\grad\subs{W} \phi(W^2)} \\
&= \frac{2}{\alpha} \cdot \nrm*{\tfrac{ 2 e^{- \gamma W^2} W}{\trace{e^{-\gamma W^2}} + e^{- 16 \gamma \alpha}}} \\
&\leq \frac{4}{\alpha} \cdot \nrm*{\tfrac{ e^{- \gamma W^2}}{\trace{e^{-\gamma W^2}} + e^{- 16 \gamma \alpha}}} \nrm{W}, 
\end{align*}
where the first inequality is due to the fact that \(\sigma'(z) \leq 2 /\alpha\), the equality in the third line is from plugging in the form of \(\grad \subs{W} \phi(W^2)\), and the last inequality is due to Cauchy-Schwarz. Using that fact that $\nrm*{\tfrac{ e^{- \gamma W^2}}{\trace{e^{-\gamma W^2}} + e^{- 16 \gamma \alpha}}} \leq 1$ and that $$\nrm{W} = \sqrt{\nrm{W^2}} \leq \sqrt{\nrm{W^2 - M} + \nrm{M}} \leq \sqrt{\nrm{W^2 - M}_F + \nrm{M}} =  \sqrt{\sqrt{F(W)} + \nrm{M}}$$
in the above, we get that 
\begin{align*}
\nrm{\grad h(W)} \leq \frac{4}{\alpha} \prn*{F(W)^{1/4} + \sqrt{\nrm{M}}}.
\end{align*}

\item Using Chain rule for matrix derivatives and Triangle Inequality, we get that 
\begin{align*}
\nrm{\grad^2 h(W)}  
&\leq 4\gamma \prn*{ \sigma''(\phi(W^2) - \alpha) + \sigma'(\phi(W^2) - \alpha)} \nrm*{\frac{e^{-\gamma W^2} W}{\trace{e^{-\gamma W^2}} + e^{- 16 \gamma \alpha}}}^2 \\
& \qquad \qquad \qquad + 2 \sigma'(\phi(W^2) - \alpha) \prn*{  \nrm*{\frac{e^{-\gamma W^2}}{\trace{e^{-\gamma W^2}} + e^{- 16 \gamma \alpha}}} + 2 \gamma \nrm*{\frac{W^2 e^{-\gamma W^2}}{\trace{e^{-\gamma W^2}} + e^{- 16 \gamma \alpha}}}} \\
&\leq 4\gamma \prn*{ \sigma''(\phi(W^2) - \alpha) + \sigma'(\phi(W^2) - \alpha)} \nrm*{W^2} + 2 \sigma'(\phi(W^2) - \alpha) \prn*{1 + 2 \gamma \nrm{W^2}}  \\
&\leq 16 \prn*{\frac{2}{\alpha^2} + \frac{1}{\alpha}} \prn*{1 + \gamma \nrm{W^2}}, 
\end{align*} where the second inequality above follows from Cauchy-Schwarz inequality, using the fact that $\nrm*{\tfrac{ e^{- \gamma W^2}}{\trace{e^{-\gamma W^2}} + e^{- 16 \gamma \alpha}}} \leq 1$ and from the observation that \(W\) is symmetric PD. Using the fact that 
\begin{align*}
\nrm{W^2} \leq \nrm{W^2 - M} + \nrm{M} \leq  \nrm{W^2 - M}_F + \nrm{M} =  \sqrt{F(W)} + \nrm{M}
\end{align*}
in the above, we get that 
\begin{align*}
\nrm{\grad^2 h(W)}  \leq  16 \prn*{\frac{2}{\alpha^2} + \frac{1}{\alpha}} \prn*{1 + \gamma \nrm{M} + \gamma \sqrt{F(W)}}. 
\end{align*}
\item We note that 
\begin{align*}
\abs*{\sigma_d(W^2) - \sigma_d(M)}^2 &\leq \nrm{W^2 - M}^2 \leq \nrm{W^2 - M}^2_F = F(W).  
\end{align*}
Thus, for any \(W\) for which \(F(W) \leq (\sigma_d(M)/2)^2\), the above implies that 
\begin{align*}
\frac{\sigma_d(M)}{2} \leq \sigma_d(W^2) \leq \frac{3 \sigma_d(M)}{2}. 
\end{align*}
The final bound follows by noting that \(\sigma_d(M) = 1600 \kappa\). Furthermore, if \(\gamma \geq \frac{\log(d + 1)}{\gamma}\), then we have that 
\begin{align*}
\phi(W^2) - \alpha \geq 14 \alpha, 
\end{align*}
which implies that \(h(W) = 1\). 
\end{enumerate} 	
\end{proof}

We next provide a rate of convergence for gradient flow on the matrix square root problem, when the initialization is well behaved.  
\begin{lemma}[\pref{lem:F_gf_matrix_completion} in the main body] 
\label{lem:msqrt_gf_rate_limited}
 Consider the objective function \(F\) given in \pref{eq:msqrt_F}. Then, for any initial point \(W(0) = W_0\) for which \(h(W_0) >  0\), where \(h\) is given in \pref{eq:h_defn_msqrt}, the point \(w(t)\) on its gradient flow path satisfies  
\begin{align*}
F(W(t)) \leq \wt R(W_0, t) \ldef{} F(W_0) \exp\prn*{- 16 \alpha t}. 
\end{align*} 
\end{lemma}

\begin{proof} Due to chain rule, we have that  
\begin{align*}
\frac{\dif F(W(t))}{\dif t} &= \tri*{\grad F(W(t)), \frac{\dif W(t)}{\dif t}} \\
&= - \nrm*{\grad F(W(t))}_F^2  && \text{(since \( \tfrac{\dif W(t)}{\dif t} = - \grad F(W(t))\))} \\
&\leq - 16 \smin(W(t)^2) F(W(t))  && \text{(using \pref{lem:msqrt_F_properties}-(b)} \\ 
&\leq - 16 \min \crl*{ \smin(W_0^2), \frac{\smin(M)}{100}} F(W(t)). && \text{(using \pref{lem:msqrt_smin_inc})}
\end{align*} 
	
Noting that \(F(W(t)) > 0\), rearranging both the sides and integrating with respect to \(t\), we get that 
\begin{align*}
\int_{\tau = 0}^t \frac{1}{F(w(\tau))} \dif F(W(\tau)) &\leq - 16 \min \crl*{ \smin(W_0^2), \frac{\smin(M)}{100}}  \int_{\tau = 0}^t \dif t. 
\end{align*} 

The above implies that 
\begin{align*}
F(W(t)) &\leq F(W(0)) \exp\prn*{- 16 t \min \crl*{ \smin(W_0^2), \frac{\smin(M)}{100}} } \\ 
&\leq F(W_0) \exp\prn*{-16 \alpha t},  
\end{align*}
where the second line above holds since 
\begin{align*}
\min \crl*{ \smin(W_0^2), \frac{\smin(M)}{100}} \geq \phi(W_0^2) \geq \alpha ,
\end{align*} where the first inequality is due to \pref{lem:msqrt_h_properties}-(a) and the second inequality holds because  \(\phi(W_0^2) > \alpha\) since \(h(W_0) > 0\).
\end{proof} 

Note that the rate in \pref{lem:msqrt_gf_rate_limited} holds for any \(W\) for which \(h(W) > 0\). However, we can extend the above to define a rate function that holds for any \(W\). Define \begin{align*}
R(W, t) = \wt R(W, t \cdot h(W)) = F(W) e^{- 16 \alpha h(W)}, 
\end{align*}
and note that for any point \(W_0\), the GF path from \(W_0\) satisfies \(F(W(t)) \leq  R(W_0, t)\). The proof is straightforward: when \(W\) is such that \(h(W) = 0\), the condition reduces to showing that  \(F(w(t)) \leq \wt R(W_0, 0) = F(W_0)\) which holds for any GF path (\pref{lem:gf_decrease}). On the other hand, when \(W_0\) is such that \(0 < h(W_0) \leq 1\), we have that \(F(W(t)) \leq \wt R(W_0, t) \leq \wt R(W_0, t \cdot h(W)) = R(W, t)\) since \(R\) is monotonically decreasing in \(W\). 

In the following lemma, we show that the function \(R\) is in-fact an admissible rate of convergence w.r.t.~\(F\), albeit under mild conditions on \(\gamma\). 

\begin{lemma} 
\label{lem:msqrt_admissible} 
Let \(\gamma \geq {\log(d + 1)}/{\alpha}\). Consider the function \(R\) defined as 
\begin{align*}
R(w, t) = F(W) e^{-  16 \alpha t h(W) },
\end{align*} 
where \(h\) is given in \pref{eq:h_defn_msqrt}. Then, \(R\) is an admissible rate of convergence w.r.t.~\(F\). 
\end{lemma}
\begin{proof}[Proof of \pref{lem:msqrt_admissible}] Recall that a sufficient conditions for a rate function \(R\) to be admissible w.r.t.~\(F\) is that for any point \(W\), 
\begin{align*} 
\int_{t=0}^\infty \prn[\Big]{\frac{\partial R(W, t)}{\partial t} + \left\langle \grad\subs{W}  R(W, t), \nabla F(W) \right\rangle } \dif t \geq 0. \numberthis \label{eq:msqrt_admissible_1}
\end{align*} 

We note that 
\begin{align*}
\int_{t=0}^\infty \frac{\partial R(W, t)}{\partial t} \dif t &= R(W, \infty) - R(W, 0) =  - F(W) \indicator{h(W) > 0}, 
\end{align*} 
and due to Chain rule, 
\begin{align*}
\int_{t=0}^\infty \tri*{\grad_w R(w, t), \grad F(W)} \dif t  &= \frac{\nrm{\grad F(W)}^2}{16 \alpha h(W)} - F(W) \frac{\tri{\grad h(W), \grad F(W)}}{16 \alpha h(W)^2}. 
\end{align*}

Taking the two terms together and rearranging, the condition in \pref{eq:msqrt_admissible_1}
 is equivalent to 
\begin{align*}
\nrm{\grad F(W)}^2 \geq 16 \alpha h(W) F(W) \indicator{h(W) > 0} + \frac{F(W)}{16 \alpha h(W)^2} \tri{\grad h(W), \grad F(W)},   \numberthis \label{eq:msqrt_admissible_2}
\end{align*}
Recall that \(h(W)= \sigma(\phi(W^2) - \alpha)\). In the following, we show that the above relation holds for any PD matrix \(W\), thus showing that  \(R\) is an admissible rate of convergence w.r.t.~\(F\). We divide the proof into the following cases: 
\begin{enumerate}[label=\(\bullet\)]
\item \textit{Case 1: when \(\phi(W^2) \leq \alpha\).} In this case, both \(h(W) = 0\) and \(\grad h(W) / h(W) = 0\) (by definition) and thus the condition in \pref{eq:msqrt_admissible_2} is trivially satisfied.
\item  \textit{Case 2: when \(\phi(W^2) \geq 2\alpha\).} In this case, \(h(W) = 1\) but \(\grad h(W) / h(W) = 0\) (by definition) and thus the condition in \pref{eq:msqrt_admissible_2} reduces to showing that \(\nrm{\grad F(W)^2} \geq 16 \alpha F(W)\), which holds due to \pref{lem:msqrt_F_properties}-(b) and the fact that \(h(W) \geq 2 \alpha\) implies that \(\smin(W) \geq 2 \alpha\) (due to \pref{lem:msqrt_h_properties}-(a)). 

\item \textit{Case 3: when \(\alpha \leq \phi(W^2) \leq 2\alpha\).} We first show that in this case, 
\begin{align*}
\alpha \leq \smin(W^2) \leq 16 \alpha.   \numberthis \label{eq:msqrt_admissible_3}
\end{align*}
The first inequality holds due to \pref{lem:msqrt_h_properties}-(a) which implies that \(\smin(W^2) \geq \phi(W^2) \geq \alpha\). The second inequality can be proved via contradiction. Suppose that \(\smin(W) \geq 16 \alpha\), then again due to \pref{lem:msqrt_h_properties}-(a), we must have that for any \(\gamma \geq \log(d +1) / \alpha\), 
\begin{align*}
\phi(W^2) &\geq \min\crl{\smin(W^2), 16 \alpha} - \frac{\log(d + 1)}{\gamma} \\ 
&\geq 16 \alpha -  \frac{\log(d + 1)}{\gamma}  \geq 15 \gamma, 
\end{align*} which contradicts the fact that  \(\phi(W^2) \leq 2 \alpha\). Thus, \pref{eq:msqrt_admissible_3} holds. We next argue that under \pref{eq:msqrt_admissible_3}, 
\begin{align*}
\tri{\grad\subs{W} h(W), \grad F(W)} \leq 0. \numberthis \label{eq:msqrt_admissible_4}
\end{align*} 
Note that 
\begin{align*}
\tri{\grad\subs{W} h(W), \grad F(W)} &= \sigma'(\phi(W^2) - \alpha)\tri{\grad\subs{W} \phi{W^2}, \grad F(W)} \\ 
&= \frac{\sigma'(\phi(W^2) - \alpha)}{\trace{e^{-\gamma W^2}} + e^{- 4\gamma \alpha}} \tri{e^{- \gamma W^2} W, \grad F(W)}, 
\end{align*}
where there the second equality follows from \pref{lem:msqrt_h_properties}. Next, observe that $\sigma'(\phi(W^2) - \alpha)$ and $\trace{e^{-\gamma W^2}} + e^{- 4\gamma \alpha}$ are both non-negative. Thus, to show \pref{eq:msqrt_admissible_4}, it suffices to show that \(\tri{e^{- \gamma W^2} W, \grad F(W)} \leq 0\). Note that  
\begin{align*}
\tri{e^{- \gamma W^2} W, \grad F(W)} &= 2 \tri{e^{- \gamma W^2} W, 2 W^3 - M W - WM } \\ 
&= 2 \trace{e^{- \gamma W^2} W \prn*{2 W^3 - M W - WM}} \\ 
&\overeq{\proman{1}} 4 \prn*{\trace{e^{- \gamma W^2} W^4} - \trace{e^{- \gamma W^2} W^2 M}} \\ 
&\overleq{\proman{2}} 4 \prn*{ \trace{e^{- \gamma W^2} W^4 } - \smin(M) \trace{e^{- \gamma W^2} W^2} } \\ 
&= 4 \prn*{ \trace{e^{- \gamma W^2} W^4 } - 1600 \alpha  \trace{e^{- \gamma W^2} W^2}}, 
\end{align*} where \(\proman{1}\) holds because \(\trace{AB} = \trace{BA}\) and because the matrices \(e^{- \gamma W^2}\) and \(W\) commute. The inequality \(\proman{2}\) follows from the fact that for PD matrices \(A, B\), we have \(\smin(B) \trace{A} \leq \trace{AB} \leq \sigma_d(B) \trace{A}\) \cite[Inequality-(1)]{fang1994inequalities}. The last line uses the fact that \(\alpha = \smin(M)/1600\). For the ease of notation, let \(\beta_i\) denote the \(i\)-th largest singular value of \(W\). Since \(W\) is symmetric PD, we note that the term in the RHS above can be further simplified as 
\begin{align*}
\trace{e^{- \gamma W^2} W^4 } - 1600 \alpha  \trace{e^{- \gamma W^2} W^2} &= \sum_{i=1}^d \prn*{ e^{- \gamma \beta_i^2} \beta_i^2 \prn{\beta_i^2 - 1600 \alpha}} \\
&\overleq{\proman{3}} \sum_{i \in \cI} \prn*{ e^{- \gamma \beta_i^2} \beta_i^2 \prn{\beta_i^2 - 1600 \alpha}} + e^{- \gamma \beta_d^2} \beta_d^2 \prn{\beta_d^2 - 1600 \alpha} \\
&\overleq{\proman{4}} \sum_{i \in \cI} e^{- \gamma \beta_i^2} \beta_i^4  - 1584 e^{- \gamma \alpha} \alpha^2, 
\end{align*} where in \(\proman{3}\), the set \(\cI \ldef{} \crl{1 \leq i \leq d - 1 \mid \beta_i^2 \geq 1600 \alpha}\) consists of all the indices upto \(d - 1\) for the corresponding term in the sum is positive. \(\proman{4}\) follows by ignoring negative term and using \pref{eq:msqrt_admissible_3}. For the first term in the RHS above, using the fact that for \(\beta_i \geq 1600 \alpha\) and \(\gamma \geq {\log(d)}/{\alpha}\), we have  
\begin{align*}
 e^{- \gamma \beta_i^2} \beta_i^4 &\leq e^{-800 \gamma \alpha} \alpha^2 
 \intertext{which implies that}
 \trace{e^{- \gamma W^2} W^4 } - 1600 \alpha  \trace{e^{- \gamma W^2} W^2} &\leq (d - 1) \alpha^2 e^{-800 \gamma \alpha} - 1584 e^{- \gamma \alpha} \alpha^2 \leq 0, 
\end{align*}
where the last inequality holds for any \(\gamma \geq \log(d)/\alpha\). 

Combining all the above bounds implies that \(\tri{\grad h(W), \grad F(W)} \leq 0\), and thus  \pref{eq:msqrt_admissible_2} reduces to showing that \(\nrm{\grad F(W)^2} \geq 16 \alpha h(W) F(W)\), which holds due to \pref{lem:msqrt_F_properties}-(b) and because \(h(W) \leq 1\). 
\end{enumerate} 
\end{proof} 

\subsubsection{Potential function and self-bounding regularity conditions} \label{app:msqrt_potential} 
We first establish the self-bounding regularity conditions for \(F\).
\begin{lemma}
\label{lem:f_sb_msqrt} 
For any symmetric and positive definite \(W\), the function \(F\) given in \pref{eq:msqrt_F} satisfies 
\begin{align*}
  \nrm*{\grad F(W)} \leq \nrm*{\grad F(W)}_F &\leq 2 F(W)^{3/4} + 2 \sqrt{\nrm{M} F(W)}, 
 \intertext{and}
 \nrm{\grad^2 F(W)} &\leq 6\sqrt{F(W)} + 8\nrm{M}. 
 \end{align*}
\end{lemma} 
\begin{proof}[Proof of \pref{lem:f_sb_msqrt}] Since \(\grad F(W) =  (W^2 - M) W  +  W (W^2 - M)\), we have  
\begin{align*}
\nrm{\grad F(W)}^2_F &\leq 2 \nrm{(W^2 - M) W }^2_F + 2 \nrm{W (W^2 - M)}^2_F \\
&\leq 4 \smax(W)^2 \nrm{W^2 - M}^2_F \\
&\leq 4 \smax(W^2) F(W), 
\end{align*}
where the last line holds because \(W\) is symmetric and positive definite which implies that \(\smax(W)^2 = \smax(W^2)\), and from the definition of \(F(W)\). Using the fact that 
\begin{align*}
 \smax(W^2) &\leq  \smax(W^2 - M) + \smax(M)  \leq \nrm{W^2 - M}_F + \nrm{M} = \sqrt{F(W)} + \nrm{M}, 
\end{align*}
we get  
\begin{align*}
 \nrm*{\grad F(W)}_F^2 &\leq 4 F(W)^{3/2} + 4 \smax(M) F(W), 
 \intertext{which implies that}
  \nrm*{\grad F(W)}_F &\leq 2 F(W)^{3/4} + 2 \sqrt{\nrm{M} F(W)}. 
 \end{align*}

For the bound on \(\nrm{\grad^2 F(W)}\), note that using Chain rule and Triangle inequality, we have 
\begin{align*} 
\nrm{\grad^2 F(W)} &\leq 6\nrm{W^2} + 2\nrm{M}  
\leq 6\nrm{W^2 - M} + 8\nrm{M}  
=  6\sqrt{F(W)} + 8\nrm{M}. 
\end{align*} 
\end{proof}

We define the admissible potential function using \pref{lem:indicator_utility_lemma}. First recall the definition of \(h\) that 
\begin{align*}
h(W) = \sigma\prn*{\phi(W^2) - \alpha}, 
\end{align*} here \(\phi\) is given in \pref{eq:phi_defn_msqrt} and \(\sigma\) is given in \pref{eq:sigma_defn_msqrt}. Next, recall \pref{lem:msqrt_gf_rate_limited} which shows that for any initial point \(W(0) = W_0\) for which \(h(W_0) >  0\), the point \(w(t)\) on its gradient flow path satisfies  
\begin{align*}
F(W(t)) \leq F(W_0) \exp\prn*{- 16 \alpha t} \rdef{} R(W_0, t). 
\end{align*} 
Clearly, as shown in \pref{lem:msqrt_admissible}, the function \(R(W, h(W)t)\) is an admissible rate of convergence w.r.t.~\(F\). We next note that the function \(F\) is minimized at the point \(W^* = \sqrt{M}\) and establish the following properties: 

\begin{enumerate}[label=$(\alph*)$] 
 \item The function \(\Gamma(W) \ldef{} \int_{t=0}^\infty R(W, t) \dif t\) is continuously differentiable, and \(\max\crl{\nrm{\grad \Gamma(W)}, \nrm{\grad^2 \Gamma(W)}} \leq \lambda(\Gamma(W))\) where \(\lambda\) is a positive, monotonically increasing function.  
\item  \(\max\crl{\nrm{\grad h(W)}, \nrm{\grad^2 h(W)}} \leq \pi(\Gamma(W))\) where \(\pi\) is a positive, monotonically increasing function.  
\item  \(\prn{h(W) - h(W^*)}^2 \leq \mu(\Gamma(W))\) where \(\mu\) is a positive, monotonically increasing function with the property that \(k \mu(z) \leq \mu(k z)\) for any \(k \geq 1\). 
 \end{enumerate} 
~ 
\begin{proof}[Proof of properties (a)-(c) above] ~
\begin{enumerate}[label=$(\alph*)$, leftmargin=8mm] 
\item Note that 
\begin{align*}
\Gamma(w) = \int_{t=0}^\infty R(w, t) \dif t = \frac{F(W)}{16 \alpha}. 
\end{align*}
Thus, following the bound in \pref{lem:f_sb_msqrt}, we note that 
\begin{align*}
  \nrm*{\grad \Gamma(W)} &\leq 2 \Gamma(W)^{3/4} + 2 \sqrt{\nrm{M} \Gamma(W)}, 
 \intertext{and} 
 \nrm{\grad^2 \Gamma(W)} &\leq 6\sqrt{\Gamma(W)} + 8\nrm{M}. 
 \end{align*}
Thus, we can define the function \(\lambda\) such that \(\lambda(z) = O(z^{3/4} + \nrm{M} + 1)\), which is clearly positive and monotonically increasing. 

\item From \pref{lem:msqrt_h_properties}-(d) and (e), we note that
\begin{align*}
\nrm{\grad h(W)} &\leq \frac{4}{\alpha} \prn*{F(W)^{1/4} + \sqrt{\nrm{M}}}.  
\intertext{and}
\nrm{\grad^2 h(W)} &\leq  16 \prn*{\frac{2}{\alpha^2} + \frac{1}{\alpha}} \prn*{1 + \gamma \nrm{M} + \gamma \sqrt{F(W)}}.
\end{align*}

Thus, we define the function 
\begin{align*}
\pi(z) &= \frac{4}{\alpha} \prn*{\prn{16 \alpha z}^{1/4} + \sqrt{\nrm{M}}} + 16 \prn*{\frac{2}{\alpha^2} + \frac{1}{\alpha}} \prn*{1 + \gamma \nrm{M} + \gamma \sqrt{16 \alpha z}} \\
&= O\prn*{ \prn*{\frac{1}{\alpha^2} + \frac{1}{\alpha}} \prn*{1 + \gamma \nrm{M} + \gamma \sqrt{16 \alpha z}}} ,
\end{align*} 
where the second line follows from recursive applications of AM-GM inequality. We note that the function \(\pi\) above is positive and monotonically increasing. 

\item From \pref{lem:msqrt_h_properties}-(c), we note that 
\begin{align*}
(h(w) - h(\sqrt{M}))^2 &\leq \frac{2}{\alpha} F(w) = 32 \Gamma(w).  
\end{align*}
Thus, we can define the function \(\mu(z) = 32 z\) which clearly satisfies the desired properties. 
\end{enumerate}
\end{proof}

Thus, all the required conditions in \pref{lem:indicator_utility_lemma} are satisfied which implies that the function
\begin{align*}
\Phi(w) = \frac{\Gamma(w)}{h(w)} =  \frac{F(w)}{16 \alpha \sigma\prn*{\phi(W^2) - \alpha}} \numberthis \label{eq:msqrt_admissible_potential} 
\end{align*} is an admissible potential function w.r.t.~\(F\) with \(g(z) = z\). Furthermore, following the proof of \pref{lem:indicator_utility_lemma}, we note that the function \(\Phi\) satisfies the following self-bounding regularity condition 
\begin{align*}
\nrm{\grad^2 \Phi(w)} &\leq \rho(\Phi(w)), 
\end{align*}
where the function \(\rho\) is given by
\begin{align*}
\rho(z) = \prn*{\lambda(z)  + z \pi(z)} \cdot \prn*{ \frac{2}{ h(W^*)} +   \frac{ \mu\prn*{z}}{ h(W^*)^2}} + \prn*{2 \lambda \prn{z} \cdot \pi\prn{z} + z \pi^2(z)} \cdot  \prn*{ \frac{2}{ h(W^*)} +   \frac{ \mu\prn*{z}}{ h(W^*)^2}}^2.  
\end{align*}
Using the fact that \(\lambda(z) = O(z^{3/4} + \nrm{M} + 1)\), \(\mu(z) = 32z\) and  \(\pi(z) = O\prn*{ \prn*{\frac{1}{\alpha^2} + \frac{1}{\alpha}} \prn*{1 + \gamma \nrm{M} + \gamma \sqrt{16 \alpha z}}}\) in the above, and repeatedly applying AM-GM, we get that 
\begin{align*}
\rho(z) = O \prn*{\prn*{1 + \gamma}^2 \prn*{1 + \nrm{M}}^2 \prn*{ \frac{2}{ h(W^*)} +   \frac{1}{ h(W^*)^2}}^2 \prn*{1 + z^4}}. \numberthis \label{eq:msqrt_sb_potential}
\end{align*}

\subsubsection{GD for matrix square root}  \label{app:GD_msqrt}
In the following, we provide the convergence guarantee for GD algorithm. We first define the respective problem dependent quantities and instantiate \pref{thm:GD_guarantee}  to provide a \(O(1/T)\) convergence bound for GD. We then provide a refined analysis which improves this bound to $O(e^{-T})$. 

\paragraph{\(\mb{O(1/T)}\) rate by direct application of \pref{thm:GD_guarantee}.} 

\begin{itemize}
\item \pref{lem:indicator_utility_lemma} implies the potential function
\begin{align*}
\Phi_g(w) =  \frac{F(w)}{16 \alpha \sigma\prn*{\phi(W^2) - \alpha}} 
\end{align*} with \(g(z) = z\). See \pref{app:msqrt_potential}  for more details. 
\item \pref{ass:F_regular} follows from \pref{lem:f_sb_msqrt} which implies that 
\begin{align*}
\psi(z) = 4 z^{3/2} + 4 \nrm{M} z.  
\end{align*}
\item \pref{ass:Phi_regular} follows from \pref{eq:msqrt_sb_potential} which implies that   
\begin{align*}
\rho(z) &= O \prn*{\prn*{1 + \gamma}^2 \prn*{1 + \nrm{M}}^2 \prn*{ \frac{2}{ h(W^*)} +   \frac{1}{ h(W^*)^2}}^2 \prn*{1 + z^4}} \\
&= L (1 + z^4),  
\end{align*}  where we defined \(L\) to hide the constants and the problem dependent terms. 

\item The function \(\theta\) is given by \(\theta(z) =  \int_{y=0}^{z} \frac{1}{\rho(y)} \dif y\).  
\item The function \(\zeta\) is defined such that  
\begin{align*}
\zeta^{-1}(z) &= \int_{y=0}^{z} \frac{\bridge{y}}{\psi(y)} \dif y = \int_{y=0}^{z} \frac{1}{4 \sqrt{y} + 4 \nrm{m}} \dif y.  
\end{align*}
\end{itemize} 
We note that \(\tfrac{\psi(z)}{g(z)} = 4 \sqrt{z} + 4 \nrm{M}\) is a monotonically increasing function of \(z\). Thus, using \pref{thm:GD_guarantee}, we get that setting \(\eta\) appropriately, GD converges at the rate of 
\begin{align*}
F(\wh w_T) &\leq \frac{2 \taylor(\Phi_g(w_0)) \psi(\zeta(\Phi_g(w_0))) \rho^2(\Phi_g(w_0))}{\bridge{\zeta(\Phi_g(w_0))}} \cdot \frac{1}{T} \\
&= \frac{\nu(w_0)}{T},  \numberthis \label{eq:msqr_slow_rate1}
\end{align*}
where the problem dependent constants can be computed by plugging in the definitions provided above, and the function \(\nu\) is defined to contain all the problem dependent parameters in the right hand side above. 

\paragraph{\(\mb{O(e^{-T})}\) rate via a refined analysis.} 
We can further improve over the rate in \pref{eq:msqr_slow_rate1} by a refined analysis for GD. In the following, we will show that GD in fact enjoys a \(e^{- O(T - t_0)}\) rate of convergence for GD for all \(T \geq t_0\), where \(t_0\) depends on \(w_0\) and problem dependent parameters specified below. 

Before delving into the proof of the above, we first provide the relevant improved version of problem dependent parameters that hold for any \(w\) for which \(F(w) \leq \prn*{\frac{\smin(M)}{2}}^2\):  
\begin{itemize}
\item  We first note that ${\sigma\prn*{\phi(W^2) - \alpha}} = 1$. 
\item Thus, \pref{lem:indicator_utility_lemma} implies the potential function 
\begin{align*}
\Phi_g(w) =  \frac{F(w)}{16 \alpha} 
\end{align*} with \(g(z) = z\). 
\item \pref{ass:F_regular} follows from \pref{lem:f_sb_msqrt} which implies that 
\begin{align*}
\psi(z) = 8\nrm{M} z, 
\end{align*}
since the above bound is only used when \(z \leq \nrm{W}/2\). 
\item \pref{ass:Phi_regular} follows from \pref{eq:msqrt_sb_potential} which implies that   
\begin{align*}
\rho(z) &= O \prn*{\prn*{1 + \gamma}^3 \prn*{1 + \nrm{M}}^6 \prn*{ \frac{2}{ h(W^*)} +   \frac{1}{ h(W^*)^2}}^2} \rdef{} \bar{L},  
\end{align*} 
since the above bound is only used when \(z = \Phi(w) \leq 800^2 \alpha\).

\item The function \(\theta\) is given by 
\begin{align*}
\theta(z) =  \int_{y=0}^{z} \frac{1}{\rho(y)} \dif y = \frac{z}{\bar{L}} \numberthis \label{eq:msqrt_fast1} 
\end{align*}
 \end{itemize}

We are now ready to provide the improved convergence rate for GD. Note that using \pref{eq:msqr_slow_rate1}, we have that there exists some
\begin{align*}
t_0 \leq 8 \nu(w_0) / \smin(M)^2 \numberthis \label{eq:msqrt_fast1} 
\end{align*}
such that \(F(w_{t_0}) \leq \smin(M)^2 / 8\). Using \pref{lem:msqrt_h_properties}-(f),  the above implies that \(h(w_{t_0}) = 1\). In the following, we will show via induction that \(F(w_{t}) \leq \smin(M)^2 / 8\) and \(h(w_{t}) = 1\) for all \(t \geq t_0\). The base case with \(t = t_0\) is shown above. For the induction step, consider any \(t \geq t_0\) and assume that \(F(w_{t}) \leq \smin(M)^2 / 8\) and \(h(w_{t}) = 1\); we will show that the same holds for \(w_{t+1}\). 
Starting from \pref{eq:GDsb_11} in the proof of \pref{thm:GD_guarantee}, we note that 
\begin{align*}
\taylor(\Phi(w_{t +1})) &\leq \taylor(\Phi(w_t))  - \frac{\eta}{2\rho(\Phi(w_0))} \bridge{F(w_t))}. 
\end{align*} 
However, note that \(w_t\) satisfies \(F(w_t) \leq \smin(M)^2 / 8\) and \(h(w_t) = 1\). Since, each update of GD is of magnitude at most \(\eta\), we also have that \(F(w_{t +1}) \leq \smin(M)^2 / 4\) and thus \(h(w_{t+1}) = 1\). Thus, plugging the forms of \(\theta, \rho, \Phi\) and \(g\) from \pref{eq:msqrt_fast1}, we get that 
\begin{align*}
F(w_{t+1}) &\leq F(w_t) \prn*{1 - \frac{8 \alpha \eta }{\bar{L}}}.  \numberthis \label{eq:msqrt_fast2} 
\end{align*}
The above clearly implies that \(F(w_{t+1}) \leq F(w_t) \leq \smin(M)^2 / 8\) and thus \(h(w_{t+1}) = 1\). This completes the induction step. 

Now, in order to complete the proof of convergence, note that \pref{eq:msqrt_fast2} will hold for all \(t \geq t_0\), recursing which implies that 
\begin{align*}
F(w_t) &\leq F(w_{t_0)}) \prn*{1 - \frac{8 \alpha \eta }{\bar{L}}}^{t - t_0} \leq F(w_{t_0)} e^{- \frac{8 \alpha \eta (t - t_0)}{\bar{L}}} \leq \smin(M)^2 / 8 e^{- \frac{8 \alpha \eta (t - t_0)}{\bar{L}}}.
\end{align*}

\subsubsection{SGD for matrix square root} 
We build on the problem dependent quantities introduced in \pref{app:GD_msqrt}. Suppose SGD is run with stochastic gradient estimates that satisfy \pref{ass:gradient_noise} with \(\chi(z) = \sigma^2\). Such a bound is satisfied in the classical stochastic optimization setting in which \(\grad f_\ms(w; z) = 2(W^2 - M) W  + 2W (W^2 - M) + \varepsilon_t\) where \(\varepsilon_t\) is a sub-Gaussian random variable with mean \(0\) and variance \(\sigma^2\). Using the above, we define the function $\Lambda$ used in \pref{thm:SGD_guarantee} as 
\begin{align*}
\Lambda(z) = 12 \sqrt{z} + 8 \nrm{M} + \sigma^2. 
\end{align*}

Fix any \(\bw\) such that \(\Phi(\bw) \geq \Phi(w_0)\) and define \(B = \Lambda(\zeta(\Phi(\bw)))\). Thus, \pref{thm:SGD_guarantee}  (in particular the bound in \pref{rem:SGD_bound}) implies that with probability at least \(0.7\), the point \(\wh w_T\) returned by SGD algorithm satisfies for any \(\kappa > 1\), 
\begin{align*} 
\bridge{F(\wh w_T)} &\lesssim \rho(\Phi(\bw)) \cdot \frac{\Phi(\bw)} {\Phi(\bw) - \Phi(w_0)} \cdot \sqrt{ B \theta(\Phi(\bw))} \cdot \frac{1}{\sqrt{T}}. 
\end{align*} 
Since \(g(z) = z\), the above immediately implies a bound on \(F(\wh w_T)\).  

%% file: files/appx_applications_chaterjee.tex
\subsection{Extending \cite{chatterjee2022convergence}}

Given a function $r:\reals^d \mapsto \reals^+$ and a monotonically increasing, positive function $g$, define 
\begin{align*}
    \alpha_{r,g}(w_0,\kappa) = \inf_{w : \|w - w_0\|_2 \le \kappa, F(w) \ne 0} \frac{\nabla r(w)^\top \nabla F(w)}{g(F(w))}
\end{align*}
Our main assumption on the initial point $w_0$ is that for some $\kappa >0$ and some functions $R$ and $g$, 
\begin{align*}
\int_{0}^\infty \sqrt{g^{-1}\left(\frac{r(w_0)}{t \alpha_{r,g}(w_0,\kappa)}\right) dt}< \frac{\kappa}{\sqrt{2 H}}
\end{align*}

The next lemma shows that for any initial point $w_0$ that satisfies the local condition above, one has a rate of convergence for GF starting from \(w_0\). 
\begin{lemma}
\label{lem:chatterjee}
Suppose $w_0$ satisfies \pref{eq:saurav} for some functions $R$ and $g$, and radius $\kappa = \kappa_0> 0$. Then, gradient flow starting from $w(0) = w_0$ satisfies for any \(t \geq 0\), 
$$
F(w(t)) \le g^{-1}\left(\frac{r(w_0)}{t \alpha(w_0,\kappa_0)} \right). 
$$ 
\end{lemma}
\begin{proof}[Proof of \pref{lem:chatterjee}]
From our assumption, let $R$ $g$ and $\kappa >0$ be given such that 
\begin{align*}
\int_{0}^\infty \sqrt{g^{-1}\left(\frac{r(w_0)}{t \alpha_{r,g}(w_0,\kappa)}\right) dt}< \frac{\kappa}{\sqrt{2 H}}
\end{align*}
First note that by the definition of $\alpha_{r,g}(w_0,\kappa)$, we have that for any point $w$ such that $\|w- w_0\|_2 \le \kappa$, 
$$
g(F(w))  \le  \frac{\nabla r(w)^\top \nabla F(w)}{\alpha_{r,g}(w_0,\kappa)}
$$
This implies that if we take $\Phi(w) = \frac{r(w)}{{\alpha_{r,g}(w_0,\kappa)}}$ as a potential, then for every point $w$ that is within distance $\kappa$ from $w_0$, $\Phi$ satisfies property \pref{eq:linearity_property} w.r.t.~$g$ for any point that is within distance $\kappa$ from $w_0$. Now consider the gradient flow path starting at $w_0$ and let $t_0$ be the first time the gradient flow path reaches a distance of $\kappa$ from $w_0$. Till this time, we can apply \pref{thm:potential_to_gf} and conclude that for any $t < t_0$, 
$$
g(F(w(t))) \le \frac{\Phi(w_0)}{t } = \frac{r(w_0)}{t \alpha(w_0,\kappa_0)}   
$$
Next, we will argue that $t_0 = \infty$. To this end, note that 
\begin{align*}
    \|w(t_0) - w(0)\|_2 &= \left\|\int_{0}^{t_0} \nabla F(w(t)) dt\right\|_2\\
    &\le \int_{0}^{t_0}  \left\|\nabla F(w(t)) \right\|_2 dt\\
    &\le \int_{0}^{t_0}  \sqrt{2 H  F(w(t)) } dt\\
    &\le  \sqrt{2 H} \int_{0}^{t_0}   \sqrt{g^{-1}\left(\frac{r(w_0)}{t \alpha(w_0,\kappa_0)} \right)} dt
\end{align*}
Note note that since $t_0$ is the first time we reach distance $\kappa$ from $w_0$, till that point, we have that the entire GF path is within the $\kappa$ radius from $w_0$ and hence, from our condition, 
$\int_{0}^\infty \sqrt{g^{-1}\left(\frac{r(w_0)}{t \alpha_{r,g}(w_0,\kappa)}\right) dt}< \frac{\kappa}{\sqrt{2 H}}$. USing this above, we conclude that
$$
\|w(t_0) - w(0)\|_2 \le  \sqrt{2 H} \int_{0}^{t_0}   \sqrt{g^{-1}\left(\frac{r(w_0)}{t \alpha(w_0,\kappa_0)} \right)} < \kappa
$$
But this is a contradiction since at $t_0$, the distance to $w_0$ should be $\kappa$ by definition of $t_0$. But we have shown that the distance is strictly smaller than $\kappa$. Hence we can conclude that $t_0 = \infty$. Hence we can conclude that for any $t > 0$ in fact, 
$$
F(w(t)) \le g^{-1}\left(\frac{r(w_0)}{t \alpha(w_0,\kappa_0)} \right)
$$
\end{proof}